\documentclass[runningheads]{llncs}

\usepackage{eccv}

\usepackage{eccvabbrv}

\usepackage{graphicx}
\usepackage{booktabs}
\usepackage{algorithm}
\usepackage[noend]{algpseudocode}
\usepackage{tikz}
\usepackage{wrapfig}
\usepackage{dirtree}

\usepackage[accsupp]{axessibility}  

\usepackage[pagebackref,citecolor=eccvblue]{hyperref}

\usepackage{orcidlink}

\usepackage[dvipsnames]{xcolor}
\usepackage{multirow}
\usepackage{pifont} 
\usepackage{bbm}
\usepackage{verbatim}
\usepackage{makecell}            
\usepackage{dblfloatfix}
\usepackage{enumitem}            
\usepackage{xspace}

\setlist[itemize, 1]{label =\raisebox{-0.05\height}{\scalebox{1.4}{\textbullet}}}
\setlist[itemize]{noitemsep, topsep=0.1cm, leftmargin=0.5cm}

\definecolor{my_blue}{rgb}{0.2, 0.6, 1}

\newcommand{\noIndentHeading}[1]{\noindent\textbf{#1}}

\definecolor{XLcolor}{rgb}{0.858, 0.188, 0.478}

\newcommand{\forExample}{\textit{e.g.}\xspace}
\newcommand{\thatIs}{\textit{i.e.}\xspace}

\newcommand{\cmark}{\checkmark}
\newcommand{\xmark}{\ding{53}}

\newcommand{\scaleFraction}{0.85}

\newcommand{\myTopRule}{\Xhline{2\arrayrulewidth}}

\newcommand{\monoThreeD}{Mono3DOD\xspace}
\newcommand{\monoDE}{MonoDE\xspace}

\newcommand{\twoD}{$2$D\xspace}
\newcommand{\threeD}{$3$D\xspace}

\newcommand{\iouThreeD}{IoU$_{3\text{D}}$\xspace}

\newcommand{\lidar}{LiDAR\xspace}

\newcommand{\kitti}{KITTI\xspace}
\newcommand{\nuscenes}{nuScenes\xspace}
\newcommand{\waymo}{Waymo\xspace}

\newcommand{\val}{Val\xspace}

\newcommand{\frontal}{Frontal\xspace}

\newcommand{\kittiThreeSixty}{KITTI-360\xspace}

\newcommand{\ddad}{DDAD\xspace}

\newcommand{\depthmap}{depthmap\xspace}
\newcommand{\depthmaps}{depthmaps\xspace}
\newcommand{\Depthmap}{Depthmap\xspace}
\newcommand{\Depthmaps}{Depthmaps\xspace}
\newcommand{\stereoLeft}{l}
\newcommand{\stereoRight}{r}

\newcommand{\setOfThreeDPts}{\mathcal{V}}

\newcommand{\pixel}{\mathbf{p}}

\newcommand{\intrinsic}{\mathbf{K}}

\newcommand{\transVec}{\mathbf{t}}

\newcommand{\rotMat}{\mathbf{R}}
\newcommand{\leftPixelSym}{p}
\newcommand{\rightPixelSym}{q}
\newcommand{\leftPixel}{\mathbf{\leftPixelSym}}
\newcommand{\rightPixel}{\mathbf{\rightPixelSym}}
\newcommand{\idntyMat}{\mathbf{I}}
\newcommand{\xcoord}{x}
\newcommand{\ycoord}{y}
\newcommand{\zcoord}{z}
\newcommand{\transpose}{\intercal}

\newcommand{\realDomain}{\mathbb{R}}

\newcommand{\first}[1]{$\textcolor{blue}{\mathbf{#1}}$}
\newcommand{\second}[1]{$\textcolor{orange}{\mathbf{#1}}$}
\newcommand{\firstKey}[1]{\textcolor{blue}{\textbf{#1}}}
\newcommand{\secondKey}[1]{\textcolor{orange}{\textbf{#1}}}
\newcommand{\sota}{SoTA\xspace}

\newcommand{\mathDash}{$-$}

\newcommand{\ap}{AP\xspace}

\newcommand{\apThreeDForty}{\ap$_{3\text{D}|R_{40}}$\xspace}

\newcommand{\bracketPercentage}{[\%]}

\newcommand{\absRel}{ARel\xspace}
\newcommand{\sqRel}{SRel\xspace}
\newcommand{\rmse}{RMS\xspace}
\newcommand{\rmseLog}{RMS\textsubscript{log}\xspace}
\newcommand{\logTen}{log$_{10}$\xspace}
\newcommand{\siLog}{SI\textsubscript{log}\xspace}

\newcommand{\kinematicVideo}{Kinematic\xspace}

\newcommand{\caddn}{CaDDN\xspace}

\newcommand{\monodtr}{MonoDTR\xspace}

\newcommand{\monorun}{MonoRUn\xspace}

\newcommand{\monodle}{MonoDLE\xspace}
\newcommand{\monopair}{MonoPair\xspace}
\newcommand{\mThreeDrpn}{M3D-RPN\xspace}
\newcommand{\dFourLCN}{D4LCN\xspace}
\newcommand{\dfrNet}{DFR-Net\xspace}

\newcommand{\idisc}{iDisc\xspace}
\newcommand{\zoeDepth}{ZoeDepth\xspace}

\definecolor{cvprblue}{rgb}{0.21,0.49,0.74}

\newcommand{\methodName}{RePLAy\xspace}

\title{\methodName: \textcolor{ForestGreen}{Re}move \textcolor{ForestGreen}{P}rojective \textcolor{ForestGreen}{L}iDAR \Depthmap \textcolor{ForestGreen}{A}rtifacts via Exploiting Epipolar Geometr\textcolor{ForestGreen}{y}}

\titlerunning{\methodName}

\author{Shengjie Zhu$^*$ \and Girish Chandar Ganesan$^*$ \and Abhinav Kumar \and Xiaoming Liu }
\authorrunning{S.~Zhu et al.}

\institute{
Department of Computer Science and Engineering,\\
Michigan State University, East Lansing, MI, USA, 48824 \\
\email{[zhusheng,~ganesang,~kumarab6,~liuxm]@msu.edu}\\
\textcolor{magenta}{\url{https://shngjz.github.io/RePLAy/}}
}

\begin{document}

\setcounter{secnumdepth}{4}

\maketitle
\def\thefootnote{*}\footnotetext{Equal Contributions}
\def\thefootnote{\arabic{footnote}}

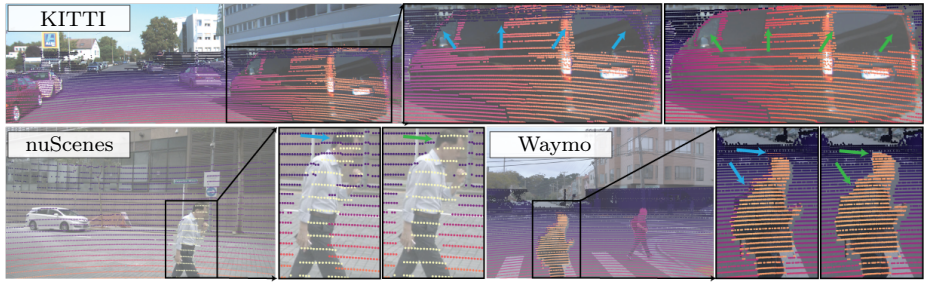
\begin{figure}[H]
  \captionsetup{font=small}
  \centering
  \begin{tikzpicture}
  \draw (0, 0) node[inner sep=0] {\includegraphics[width=\linewidth]{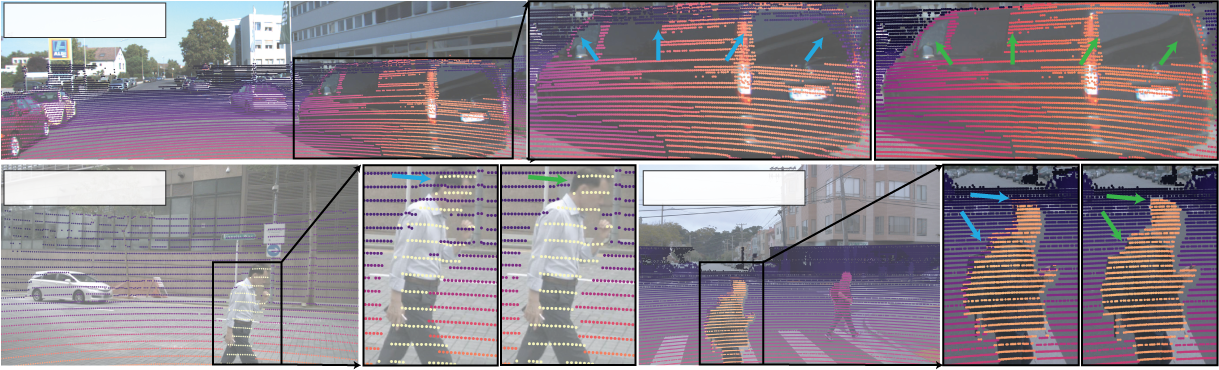}};
  \draw (-5.2, 1.65) node[inner sep=0] {\fontsize{8.0}{10}\selectfont \kitti};
  \draw (-5.25, -0.023) node[inner sep=0] {\fontsize{8.0}{10}\selectfont \nuscenes};
  \draw (1.2, -0.04) node[inner sep=0] {\fontsize{8.0}{10}\selectfont \waymo};
  \end{tikzpicture}
    \vspace{-6mm}
    \caption{\small 
    \textbf{Remove Projective LiDAR \Depthmap Artifacts.} 
    The \lidar projected \depthmap has misalignment with the RGB camera.
    The background scans highlighted by \textcolor{Cyan}{blue arrows} incorrectly overlay the foreground such as cars and pedestrians.
    The artifacts persist unattended in most AV datasets including \kittiThreeSixty~\cite{liao2022kitti360}, \nuscenes~\cite{caesar2020nuscenes}, \waymo~\cite{sun2020scalability}, and \ddad~\cite{packnet}.
    We propose an easy-to-use, parameter-free, and analytical solution to remove the projective artifacts, shown by \textcolor{ForestGreen}{green arrows}.
    \vspace{-3mm}}
    \label{fig:teaser}
\end{figure}

\begin{abstract}
\threeD sensing is a fundamental task for Autonomous Vehicles.
Its deployment often relies on aligned RGB cameras and \lidar.
Despite meticulous synchronization and calibration, systematic misalignment persists in \lidar projected \depthmap.
This is due to the physical baseline distance between the two sensors.
The artifact is often reflected as background \lidar incorrectly projected onto the foreground, such as cars and pedestrians.
The \kitti dataset uses stereo cameras as a heuristic solution to remove artifacts. 
However most AV datasets, including \nuscenes, \waymo, and \ddad, lack stereo images, making the \kitti solution inapplicable.
We propose \textit{\methodName}, a parameter-free analytical solution to remove the projective artifacts.
We construct a binocular vision system between a hypothesized virtual \lidar camera and the RGB camera.
We then remove the projective artifacts by determining the epipolar occlusion with the proposed analytical solution.
We show unanimous improvement in the State-of-The-Art (\sota) monocular depth estimators and \threeD object detectors with the artifacts-free \depthmaps.

\end{abstract}

\section{Introduction}\label{sec:intro}

\lidar device scans the surrounding environment via emitting laser pulses into the \threeD space.
Its applications span across Autonomous Vehicles (AV)~\cite{sun2020scalability, royo2019overview, li2020lidar}, AR/VR~\cite{hillmann2019comparing, milanovic2017fast}, and Robotics\cite{hutabarat2019lidar}.
Taking AV as an example, the \lidar is synchronized and calibrated against the RGB camera.
The correspondence between RGB pixels and \lidar \threeD points is determined by projecting the \lidar \threeD point onto the image plane, as shown with the \cref{fig:teaser} \depthmaps.

Despite ideal synchronization and calibration, there still exists systematic misalignment between the RGB camera and \lidar.
\cref{fig:teaser} highlights the projective artifacts in \lidar \depthmap with \textcolor{Cyan}{{blue arrows}}, where the background scans incorrectly overlay on the foreground objects.


\begin{figure}[!t]
    \centering
    \subfloat[Autonomous Vehicles~\cite{caesar2020nuscenes}]{
    \begin{tikzpicture}
        \draw (0,0) node[inner sep=0] {\includegraphics[height=0.9cm]{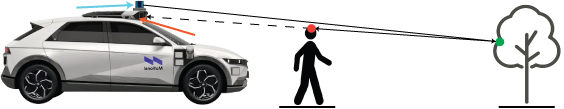}};
        \draw (-0.37,0.15) node[inner sep=0] {\fontsize{3.0}{10}\selectfont Camera};
        \draw (-2.1,0.4) node[inner sep=0] {\fontsize{3.0}{10}\selectfont \lidar};
    \end{tikzpicture}
    \label{fig:sensor_AV}
    }\quad 
    \subfloat[Robotics Machines~\cite{ebadi2022present}]{
    \begin{tikzpicture}
        \draw (0,0) node[inner sep=0] {\includegraphics[height=0.9cm]{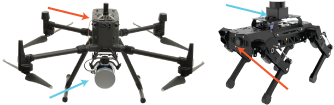}};
        \draw (-1.3,0.5) node[inner sep=0] {\fontsize{3.0}{10}\selectfont Camera};
        \draw (-1.4,-0.35) node[inner sep=0] {\fontsize{3.0}{10}\selectfont \lidar};
        \draw (1.4,-0.35) node[inner sep=0] {\fontsize{3.0}{10}\selectfont Camera};
        \draw (0.5,0.5) node[inner sep=0] {\fontsize{3.0}{10}\selectfont \lidar};
    \end{tikzpicture}
    } \quad
    \subfloat[Headset~\cite{herskovitz2020making}]{
    \begin{tikzpicture}
        \draw (0,0) node[inner sep=0] {\includegraphics[height=0.9cm]{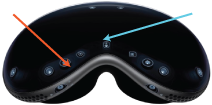}};
        \draw (-0.6,0.5) node[inner sep=0] {\fontsize{3.0}{10}\selectfont Camera};
        \draw (0.9,0.5) node[inner sep=0] {\fontsize{3.0}{10}\selectfont \lidar};
    \end{tikzpicture}
    } 
    \caption{
    \textbf{\lidar-RGB sensor sets} have a physical baseline distance between the two sensors, resulting in \cref{fig:teaser} projective artifacts.
    \textbf{(a)} The \textcolor{ForestGreen}{background} scanned by \lidar incorrectly overlays on the \textcolor{RedOrange}{foreground} pedestrian observed by the RGB camera.
    \vspace{-3mm}}
    \label{fig:sensor}
\end{figure}
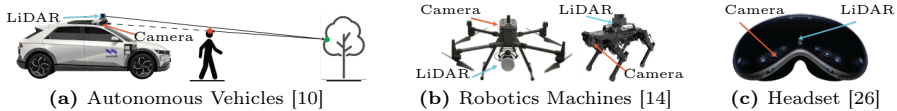

The projective artifacts arise from the physical baseline distance between the RGB camera and \lidar. 
This physical limitation is ubiquitous among different LiDAR-RGB sensor sets used in Robotics, AR/VR, and AV tasks (\cref{fig:sensor}). 
The baseline distance (\cref{fig:sensor_AV}) renders certain \threeD scans exclusively visible to \lidar, rather than the RGB camera.
For those scans, their correspondence with image pixels is absent, causing \cref{fig:teaser} projective artifacts.

\begin{figure}[!t]
    \centering
    \subfloat[Raw Depthmap]{
    \includegraphics[width=0.31\linewidth,trim={1cm 0 1.5cm 0},clip]{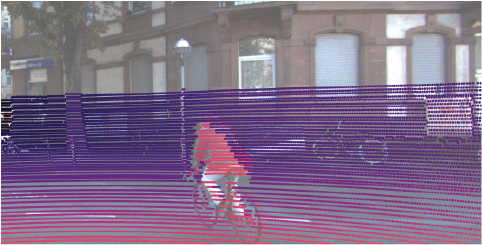}  } \;
    \subfloat[\methodName]{\includegraphics[width=0.31\linewidth,trim={1cm 0 1.5cm 0},clip]{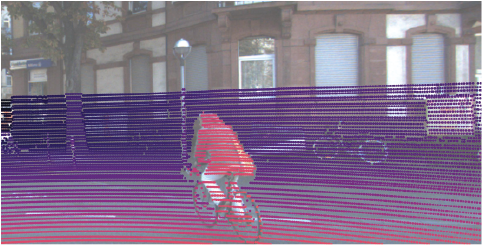}} \; 
    \subfloat[KITTI~\cite{uhrig2017sparsity} with Stereo]{\includegraphics[width=0.31\linewidth,trim={1cm 0 1.5cm 0},clip]{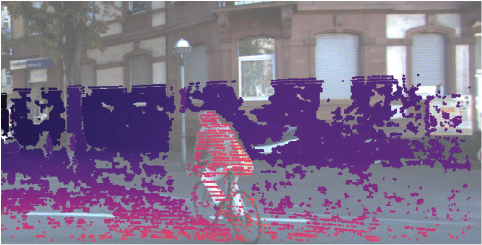}}
    \vspace{-2mm}
    \caption{
    \textbf{Qualitative comparison with KITTI semi-dense depthmap.}
    \methodName only requires the \lidar calibration matrix.    
    KITTI~\cite{uhrig2017sparsity} uses multi-frame fusion with the stereo camera to remove projective artifacts.
    The higher density of \kitti depthmap is from the additional sensor and temporal fusion.
    \vspace{-3mm}}
    \label{fig:compare_kitti}
\end{figure}

The \kitti~dataset \cite{kitti2012benchmark} provides a heuristic solution to remove the projective artifacts, which we show in \cref{fig:compare_kitti}.
However, the method requires stereo sensors, making it inapplicable to other AV datasets.
Thus, the projective artifacts persist \textbf{unattended} in most AV datasets. \cref{tab:data_survey} summarizes these datasets.

We propose \methodName: a parameter-free, non-learning, and analytical solution to resolve the projective artifacts.
\methodName only requires the relative pose between the \lidar scanner and the RGB camera, \textit{i.e.}, their extrinsic matrix or calibration matrix.
It applies to all AV datasets surveyed in \cref{tab:data_survey} and diverse sensor settings included in \cref{fig:sensor}.
\methodName does not even require RGB images, hence being intact at the asynchronization noise between the RGB camera and \lidar.
\cref{fig:compare_kitti} provides a qualitative comparison with \kitti solution.

\methodName first adopts an auto-calibration algorithm to calibrate a hypothesized virtual camera at the origin of the \lidar \threeD point cloud.
The virtual \lidar camera formulates a stereo system with the RGB camera.
We attribute the projective artifact to the epipolar occlusion in stereo vision. 
We propose an analytical solution for detecting epipolar occlusion using epipolar geometry. 

\lidar projected \depthmap is essential for diverse downstream applications.
We select two to verify our effectiveness: monocular depth estimation (\monoDE)~\cite{bhat2023zoedepth, wei2023surrounddepth, zhu2020edge, piccinelli2023idisc} and monocular \threeD object detection (\monoThreeD) that uses auxiliary \depthmap ~\cite{wu2023virtual, wu2023transformation, yang2022deepinteraction, reading2021categorical}.
We benchmark their performance trained using \lidar \depthmap cleaned with multiple algorithms and \methodName on AV datasets.
We demonstrate \textbf{unanimous} improvement in State-of-The-Art (SoTA) methods~\cite{reading2021categorical, huang2022monodtr, bhat2023zoedepth} on both tasks.
We additionally release the processed \depthmap of five major AV datasets to benefit the community.
Our contributions are:
\begin{itemize}
    \item[\ding{51}] We utilize epipolar geometry to remove the projective artifacts in the \lidar \depthmap.
    \item[\ding{51}] We demonstrate unanimous improvement in the SoTA monocular depth and monocular \threeD object detection methods.
    \item[\ding{51}] We provide processed \depthmap in \kitti~\cite{kitti2012benchmark}, \kittiThreeSixty~\cite{liao2022kitti360}, \nuscenes~\cite{caesar2020nuscenes}, \waymo~\cite{sun2020scalability}, and \ddad~\cite{packnet} datasets.
\end{itemize}

\begin{table*}[!t]
    \centering
    \caption{
    \textbf{\lidar \Depthmap Projective Artifacts} exist in most Autonomous Vehicle Datasets.
    \kitti uses stereo images as a heuristic solution. 
    }
    \resizebox{\linewidth}{!}{%
    \begin{tabular}[width=\linewidth]{l | c c c c c c c c|}
    \hline
     & \kitti~\cite{kitti2012benchmark} & \kittiThreeSixty~\cite{liao2022kitti360} & \nuscenes~\cite{caesar2020nuscenes} & \waymo~\cite{sun2020scalability} & \ddad~\cite{packnet} & ApolloScape~\cite{huang2018apolloscape} & Argoverse~\cite{wilson2023argoverse}
     \\
    \hline
    Artifacts & \textcolor{ForestGreen}{\textbf{No}} & \textcolor{RedOrange}{\textbf{Yes}} & \textcolor{RedOrange}{\textbf{Yes}} & \textcolor{RedOrange}{\textbf{Yes}} & \textcolor{RedOrange}{\textbf{Yes}} & \textcolor{RedOrange}{\textbf{Yes}} & \textcolor{RedOrange}{\textbf{Yes}} 
    \\
    \hline
    \end{tabular}
    }
    \label{tab:data_survey}
\end{table*}

\section{Related Work}\label{rel_work}

\noIndentHeading{\lidar and Camera Misalignment.}
Multiple sources of misalignment exist between \lidar and RGB camera. 
Foggy, rainy, and snowy weather negatively impacts both \lidar~\cite{michaud2015towards, filgueira2017quantifying, kutila2018automotive} and RGB cameras~\cite{garg2005does, garg2007vision, spinneker2014fast, garvelmann2013observation}.
The slight drift of \lidar and camera accumulates during vehicle operation.
\cite{yu2023benchmarking, bai2022transfusion, li2022deepfusion} investigate the robustness to the spatial misalignment. 
\cite{yu2023benchmarking} benchmarks the temporal asynchronization between \lidar and RGB camera.
The projective artifact is another misalignment that universally exists in multiple AV datasets, as shown in \cref{tab:data_survey}.
Our solution complements prior study~\cite{yu2023benchmarking} by addressing it.

\noIndentHeading{Heuristic Solution for Projective Artifacts.}
\kitti~\cite{kitti2012benchmark, uhrig2017sparsity} dataset uses the stereo image as a heuristic solution.
\cite{uhrig2017sparsity} produces a stereo reconstruction through semi-global matching~\cite{hirschmuller2007stereo}.
They additionally augment the stereo results with $11$ neighboring temporal frames.
The projective artifacts are removed via consistency verification between the stereo \depthmap and \lidar \depthmap.
Compared to ours, \depthmap from \cite{uhrig2017sparsity} has improved density.
However, the requirement of stereo sensors limits its application, as summarized in \cref{tab:data_survey}.
Additionally, \cite{uhrig2017sparsity} has a complicated pipeline, and is therefore, not used even in the \kittiThreeSixty~\cite{liao2022kitti360} dataset, which shares a similar setup.

\noIndentHeading{Stereo Occlusion Detection.}
Stereo co-planar cameras are classically formulated as the half-occlusion problem in stereo vision, \textit{i.e.}, the area occluded in one camera of a stereo vision system.
There has been extensive research on detecting the half-occlusion region~\cite{wang2019local, arbelaez2010contour, martin2004learning, xie2015holistically}.
Several works~\cite{ishikawa1998occlusions, bobick1999large, birchfield1999depth, wei2005asymmetrical, zitnick2000cooperative} jointly optimize occlusion and disparity.
Others~\cite{weng1988two, fua1993parallel, sun2005symmetric} apply left-right consistency check.
The area receives different estimations from left and right views are excluded.
\methodName focusses on setups without stereo but where \lidar provides rough 3D structures.
Compared to the half-occlusion baseline which proposes a heuristic solution, \methodName is an analytical solution.

\begin{algorithm}[!t]
    \caption{\small \methodName: Remove Projective Artifacts}
    \label{algo:dense_algo}
    \begin{algorithmic}[1]
        \Require \lidar Point Cloud $\mathcal{V}$, Calibration Matrix $\mathbf{P} = \begin{bmatrix}
            \mathbf{R} & \mathbf{t}
        \end{bmatrix}$, Camera Intrinsic $\mathbf{K}_r$
        \Ensure Occluded Pixels $\mathcal{O}$
        \State Auto-Calibration 
        \Comment{\cref{sec:lidar_rgb}}
        \State Set Virtual \lidar Camera $\mathbf{K}_l$
        \State Produce Virtual \lidar Depthmap $\mathbf{D}_l$
        \State Densify Virtual \lidar Depthmap $\overline{\mathbf{D}}_l$
        \For {$\mathbf{v} \in \mathcal{V}$}
            \State $\mathbf{p}_1, \; d_{\mathbf{p}_1} \longleftarrow \pi(\mathbf{R}\mathbf{v} \mid \mathbf{K}_l)$
            \For {$\mathbf{p}_2 \in \mathcal{L}_{\mathbf{p}_1}$}
                \Comment{\cref{eqn:occ_set} and \cref{fig:pipeline}}
                \State $d_{\mathbf{p}_2} \longleftarrow $ bilinear interpolate $\overline{\mathbf{D}}_l$ at pixel location $\mathbf{p}_2$
                \Comment{\cref{fig:pipeline}}
                \If {$ g(\mathbf{p}_1,\mathbf{p}_2, d_{\mathbf{p}_1}, d_{\mathbf{p}_2} \mid \mathbf{K}_l, \mathbf{K}_r, \begin{bmatrix}
                    \mathbf{I} & \mathbf{t}
                \end{bmatrix}) \leq 0$} \Comment{\cref{eqn:suffi}}
                    \State $\{ \mathbf{p}_1 \} \cup \mathcal{O}$
                    \State break
                \EndIf
            \EndFor
        \EndFor
    \end{algorithmic}
\end{algorithm}

\noIndentHeading{Downstream Tasks using \lidar \Depthmap.}
Multiple downstream tasks use \lidar projected \depthmaps. 
\textit{E.g.}, 
\monoDE methods~\cite{yuan2022neural,lee2019big, bhat2021adabins, zhu2023lighteddepth, bhat2023zoedepth} use \depthmap as the GT supervision.
\monoThreeD methods~\cite{wu2023virtual, wu2023transformation, yang2022deepinteraction, kumar2024seabird,liu2021voxel} use \depthmap for supervising auxiliary task along with \threeD object detection.
When deployed to AV datasets other than KITTI, the aforementioned methods~\cite{lin2020depth, zheng2022depth} have diminished performance due to the projective artifacts.
\methodName improves their performance by resolving the projective artifacts in \lidar \depthmap.

\section{\methodName}

We now describe \methodName in \cref{algo:dense_algo}.
\cref{sec:lidar_rgb} attributes the projective artifacts to the epipolar occlusion induced by the binocular vision system between the virtual \lidar camera and RGB camera.
\cref{sec:solution} outlines our analytical solution.
\cref{sec:discuss} hold further discussions.

We execute \cref{sec:lidar_rgb} auto-calibration over the entire dataset.
We initialize \lidar intrinsic $\mathbf{K}_l$ the same as the RGB camera $\mathbf{K}_r$ with slight augmentation to ensure a sufficient Field-of-View.

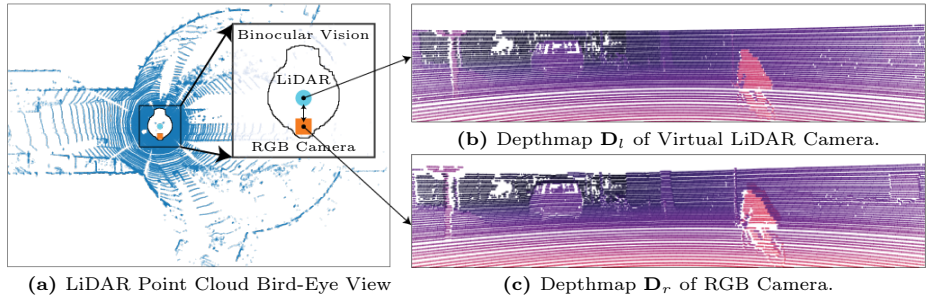
\begin{figure}[!t]
    \begin{minipage}{0.44\textwidth}%
        \begin{subfigure}{\linewidth}%
        \begin{tikzpicture}
            \draw (0, 0) node[inner sep=0]{\includegraphics[width=\textwidth]{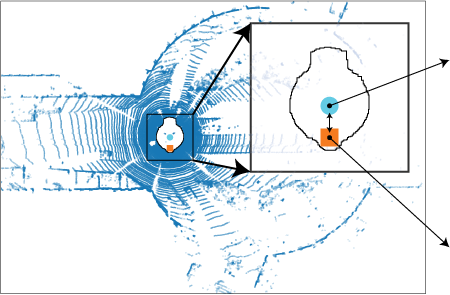}};
            \draw(1.25,0.75) node[inner sep=0]{\fontsize{3.0}{10}\selectfont \lidar };
            \draw(1.25,-0.15) node[inner sep=0]{\fontsize{3.0}{10}\selectfont RGB Camera};
            \draw(1.25,1.35) node[inner sep=0]{\fontsize{3.0}{10}\selectfont Binocular Vision};
        \end{tikzpicture}
            
            \caption{\lidar Point Cloud Bird-Eye View}
            \label{fig:bev}
        \end{subfigure}
    \end{minipage}%
    \begin{minipage}{0.56\textwidth}%
        \begin{subfigure}{\linewidth}%
            \includegraphics[width=\linewidth]{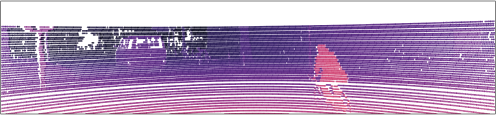}
            \caption{ Depthmap $\mathbf{D}_l$ of Virtual \lidar Camera.}
            \label{fig:depth_wo_art}
        \end{subfigure}
        \begin{subfigure}{\linewidth}%
            \includegraphics[width=\linewidth]{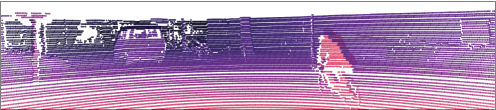}
            \caption{Depthmap $\mathbf{D}_r$ of RGB Camera.}
            \label{fig:depth_wt_art}
        \end{subfigure}
    \end{minipage}%
    \vspace{-1mm}
    \caption{
    \small 
    \textbf{Virtual \lidar camera and RGB Binocular System.} 
    In (a), we plot the bird-eye view of the point cloud $\mathcal{V}$ from \lidar.
    A binocular system is set between virtual \lidar (epipolar left) and RGB camera (epipolar right).
    Subfigure (b) plots the depthmap $\mathbf{D}_l$ from \lidar view.
    As stated in \cref{collary_lidar}, depthmap $\mathbf{D}_l$ does not have projective artifacts.
    Compare (b) and (c), the artifacts arise when transfer from \lidar to RGB camera, due to the epipolar occlusion among the binocular system.
    We formally define the projective artifacts (or epipolar occlusion) in \cref{eqn:def_occ}.
    \vspace{-3mm}
    }
    \label{fig:stereo_set}
\end{figure}

\subsection{\lidar \Depthmap Projective Artifacts as Epipolar Occlusion}\label{sec:lidar_rgb}

\cref{fig:sensor} demonstrates the baseline distance between various \lidar and RGB camera sensor sets.
\cref{fig:sensor_AV} shows that the background \textcolor{ForestGreen}{tree} is solely visible to \lidar.
Intuitively, the background \textcolor{ForestGreen}{tree} is occluded by the foreground \textcolor{RedOrange}{pedestrian} in a binocular vision system of the virtual \lidar camera and RGB camera.

\noIndentHeading{Virtual \lidar Camera.}
\lidar produces a point cloud $\mathcal{V}$ by emitting laser pulses into \threeD space.
Mathematically, we assume the \lidar point cloud $\mathcal{V}$ as a set of \threeD rays originating from the coordinate origin:
\begin{equation}
    \mathcal{V} = \{\mathbf{v}_i\} = \{{\mathbf{0} + \overline{\mathbf{r}}_i \cdot t_i} \mid i \in \mathbb{Z},  \; t > 0, \;\overline{\mathbf{r}}_i \in \mathbb{R}^3  \},
    \label{eqn:ray}
\end{equation}
where $\mathbf{v}_i$ is a \threeD point. 
$\overline{\mathbf{r}}_i$ is a normalized direction vector, and $t_i$ is the euclidean distance between the \threeD point and \lidar.
If we position a virtual \lidar camera with arbitrary resolution and intrinsic $\mathbf{K}_l$ at the \lidar's location, then:
\begin{lemma}
\label{collary_lidar}
The \depthmap of the virtual \lidar camera is free of projective artifacts, \thatIs, no two scanned points $\mathbf{v}_1$ and $\mathbf{v}_2$ overlap after projection.
\begin{equation}
\forall \mathbf{v}_1, \mathbf{v}_2 \in \mathcal{V}, \;\Vert\pi(\mathbf{v}_1 \mid \mathbf{K}_l)-\pi(\mathbf{v}_2 \mid \mathbf{K}_l)\Vert_2 \neq 0,
\end{equation}
\end{lemma}
where $\pi (\cdot)$ stands for the projection process. 
See proof in Supp. (Appendix A).
We visualize the artifacts-free depthmap $\mathbf{D}_l$ from the virtual \lidar camera in \cref{fig:depth_wo_art}.

\begin{figure}[!t]
    \centering
    \subfloat{
    \begin{tikzpicture}
        \draw (0,0) node[inner sep=0] {\includegraphics[width=\linewidth]{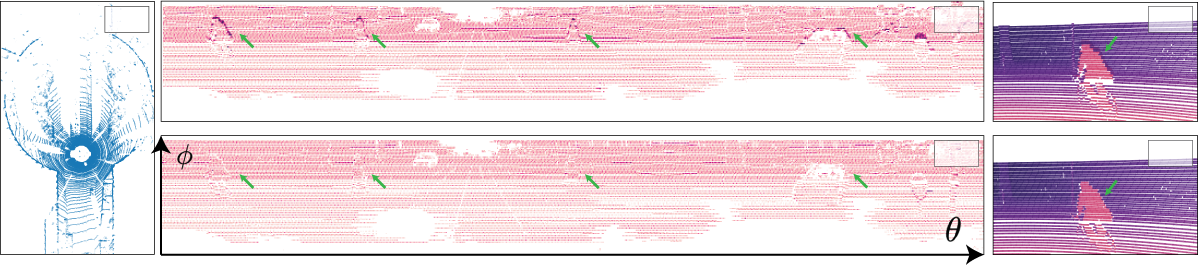}};
        \draw (-4.9,1.15) node[inner sep=0, anchor=west] {\fontsize{3.0}{10}\selectfont \textbf{a}};
        \draw (3.55,1.15) node[inner sep=0, anchor=west] {\fontsize{3.0}{10}\selectfont \textbf{b}};
        \draw (3.55,-0.22) node[inner sep=0, anchor=west] {\fontsize{3.0}{10}\selectfont \textbf{c}};
        \draw (5.75,1.15) node[inner sep=0, anchor=west] {\fontsize{3.0}{10}\selectfont \textbf{d}};
        \draw (5.75,-0.22) node[inner sep=0, anchor=west] {\fontsize{3.0}{10}\selectfont \textbf{e}};
    \end{tikzpicture}
    }
     \vspace{-1mm}
    \caption{
    \textbf{Auto-Calibration.} The \lidar point cloud $\mathcal{V}$ in (a) is mapped to spherical coordinate $\theta - \phi$ with \cref{eqn:spherecal}.
    We use a binary matrix $\mathbf{S}$ to record the unique pixel locations $\mathcal{S}$ after rasterization.  
    Plots (b) and (c) are matrix $\mathbf{S}$ before and after auto-calibration where duplication (marked by arrows) is minimized with \cref{eqn:auto_loss} loss $L$.
    Plots (d) and (e) contrast the virtual \lidar depthmap $\mathbf{D}_l$ before and after auto-calibration where \cref{collary_lidar} is fulfilled.
    \vspace{-3mm}}
    \label{fig:auto_calib}
\end{figure}
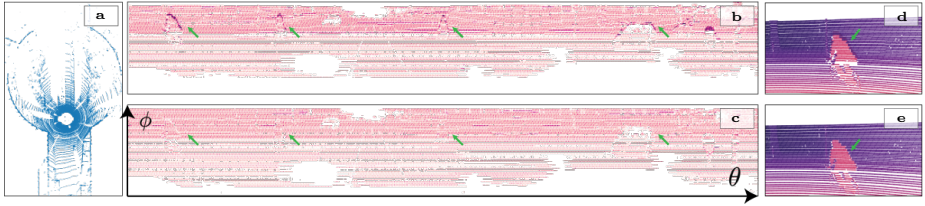

\noIndentHeading{Auto-Calibration.} 
In practice, \cref{collary_lidar} may not hold in a realistic \lidar point cloud due to possible bias in the point cloud coordinate origin $\mathbf{0}$.
\cref{fig:auto_calib}d, an example from the \kitti dataset shows persistent projective artifacts in the depthmap $\mathbf{D}_l$ when without any treatment.
We perform auto-calibration to fulfill \cref{collary_lidar} by re-adjusting the point cloud origin.
\cref{fig:stereo_set}b  and \cref{fig:auto_calib}e show artifacts-free depthmap after auto-calibration.

Auto-calibration aims to resolve the artifacts in the virtual \lidar camera depthmap $\mathbf{D}_l$.
We convert each \lidar point cloud to the spherical coordinate system.
This hypothesizes an omnidirectional image $\mathbf{S}$ of the \lidar point cloud.
Denote the origin to estimate as $\mathbf{o}$.
Set the new \threeD point $\mathbf{u} = \mathbf{v} + \mathbf{o}$, we have the polar angle $\theta$ and azimuthal angle $\phi$ after applying a rasterization operation $[\cdot]$:
\begin{equation}
\small
\theta = [\arccos(u_x / (u_x^2 + u_z^2))], \, \phi = [\arccos(u_y / (u_x^2 + u_y^2 + u_z^2))], \, \begin{bmatrix}
    \theta & \phi
\end{bmatrix} = f(\mathbf{v} + \mathbf{o}).
\label{eqn:spherecal}
\end{equation}
Intuitively, function $f(\cdot)$ converts a 3D point $\mathbf{u} \in \mathcal{V}$ to a discrete 2D coordinate in $\mathbf{S}$.
We denote the set of unique coordinates as $\mathcal{S}$:
\begin{equation}
\mathcal{S} = \{ f(\mathbf{v} + \mathbf{o}) \mid \mathbf{v} \in \mathcal{V}  \}.
\label{eqn:unique_coord}
\end{equation}
We implement \cref{eqn:unique_coord} with a binary matrix $\mathbf{S}$.
Its elements are  $1$ at each discrete coordinate and  $0$ otherwise.
The size of $\mathcal{S}$ equals to summarizing the matrix $\mathbf{S}$.
Artifacts become duplication in set $\mathcal{S}$.
Removing artifacts is to minimize duplication.
The loss $L$ is then defined as:
\begin{equation}
L = \| \mathcal{V} \| - \| \mathcal{S} \|.
\label{eqn:auto_loss}
\end{equation}
The loss $L$ is not differentiable. 
In each iteration, we approximate its gradient by ablating origin $\mathbf{o}$ with a small deviation and perform coordinate descent.

\noIndentHeading{Virtual \lidar Camera and RGB Binocular System.}
In \cref{fig:stereo_set}, we set up a binocular vision system, denoting the virtual \lidar camera and RGB camera as the epipolar left and right camera respectively.
Correspondingly, set their intrinsic as $\mathbf{K}_l$ and $\mathbf{K}_r$, and  projected \depthmap as $\mathbf{D}_l$ and $\mathbf{D}_r$.
Thus the problem of removing the projective artifacts is translated to detecting the epipolar occlusion between the virtual \lidar camera and RGB camera.

\noIndentHeading{Projective Artifacts are Epipolar Occlusion.}
We ensure that our virtual \lidar has a larger field of view than the RGB camera by adjusting the resolution of the virtual \lidar camera.
Denote the relative movement between the \lidar and RGB camera as $
\mathbf{P} = 
\begin{bmatrix}
    \mathbf{R} & \mathbf{t}
\end{bmatrix},
$
where the auto-calibrated center $\mathbf{o}$ is integrated into $\mathbf{t}.$
Denote the projected pixels on epipolar left (\lidar) and epipolar right (RGB) as $\mathbf{p}$ and $\mathbf{q}$, and their pixel depths as $d$ and $d'$.
We have:
\begin{equation}
\mathbf{p} = \pi (\mathbf{v} \mid \mathbf{K}_l), \quad \mathbf{q} = \pi(\mathbf{R} \mathbf{v} + \mathbf{t} \mid \mathbf{K}_r).
\label{eqn:p_q}
\end{equation}
We then formally define the epipolar occlusion $\mathcal{O}$ on RGB view as:
\begin{equation}
\mathcal{O} =
\{
\mathbf{q}_1 \mid 
\forall \; \mathbf{q}_1, \mathbf{q}_2 \in {\mathcal{Q}}, \;
     d_{\mathbf{q}_1}' > d_{\mathbf{q}_2}', \; \Vert \mathbf{q}_1-\mathbf{q}_2 \Vert_2 = 0
\}.
\label{eqn:def_occ}
\end{equation}
\cref{eqn:def_occ} simply suggests if two pixels in RGB camera overlap with each other, then the one $\mathbf{q}_1$ with larger depth is occluded by the one with smaller depth $\mathbf{q}_2$.
Occlusion detection finds all occluded pixels $\mathbf{q}_1$, \textit{i.e.}, the set $\mathcal{O}$.



\subsection{Epipolar Occlusion under Rotation and Translation}
\label{sec:solution}
\begin{figure}[t!]
  \captionsetup{font=small}
  \centering
  \begin{tikzpicture}
      \draw (0,0) node[inner sep=0] {\includegraphics[width=\linewidth]{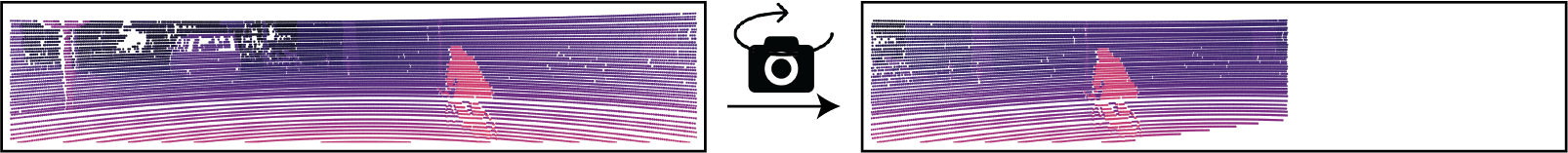}};
      \draw (0,-0.4) node[inner sep] {\fontsize{3.0}{10}\selectfont Rotation};
  \end{tikzpicture}
   \vspace{-4mm}
  \caption{
  \small 
  \textbf{Epipolar Occlusion with Pure Rotation Movement.} 
  We apply a rotation transformation along the $x$-$z$ axis in the camera coordinate system.
  As indicated by \cref{lemma:rot}, a pure rotation movement does not lead to projective artifacts.
  \vspace{-4mm} }
  \label{fig:rot}
\end{figure}

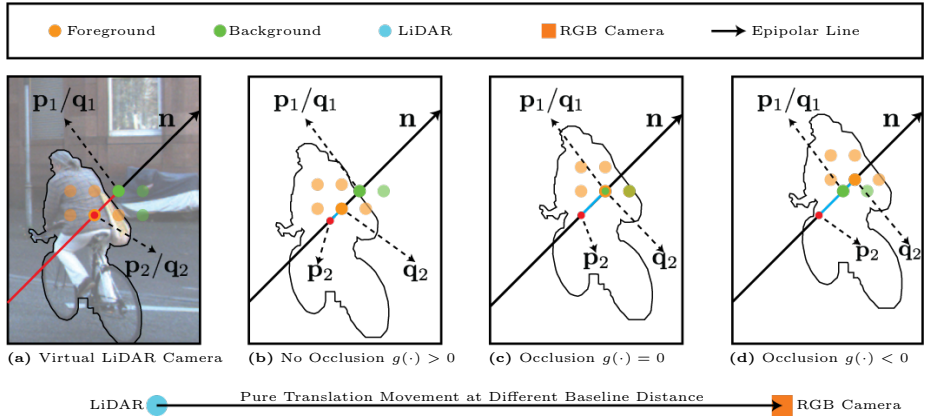
\begin{figure}[t!]
  \captionsetup{font=small}
  \centering
  \begin{tikzpicture}
      \draw (0,0) node[inner sep=0] {\includegraphics[width=\linewidth]{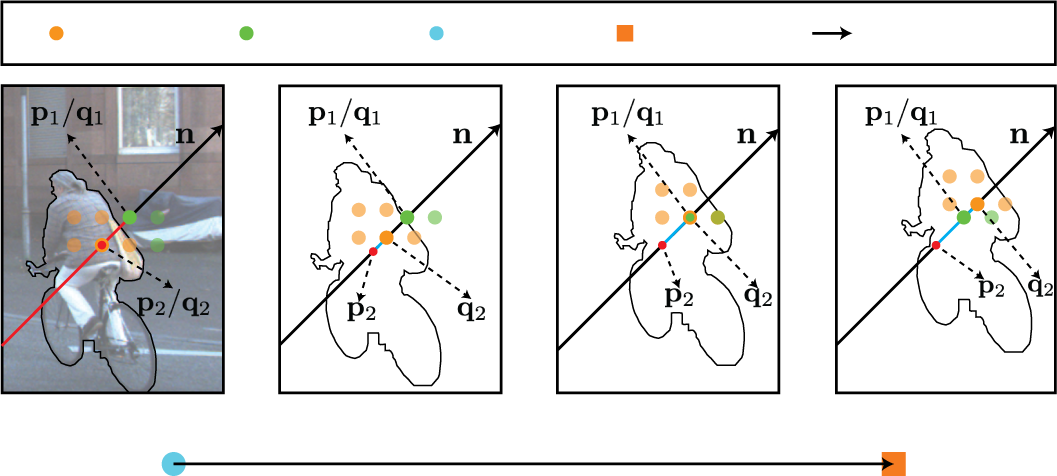}};
      \draw (-5.3,2.35) node[inner sep=0, anchor=west] {\fontsize{3.0}{10}\selectfont Foreground};
      \draw (-3.15,2.35) node[inner sep=0, anchor=west] {\fontsize{3.0}{10}\selectfont Background};
      \draw (-0.9,2.37) node[inner sep=0, anchor=west] {\fontsize{3.0}{10}\selectfont \lidar};
      \draw (1.25,2.37) node[inner sep=0, anchor=west] {\fontsize{3.0}{10}\selectfont RGB Camera};
      \draw (3.8,2.35) node[inner sep=0, anchor=west] {\fontsize{3.0}{10}\selectfont Epipolar Line};
      \draw (-6.1,-2) node[inner sep=0, anchor=west] {\fontsize{3.0}{10}\selectfont \textbf{(a)} Virtual \lidar Camera};
      \draw (-2.9,-2) node[inner sep=0, anchor=west] {\fontsize{3.0}{10}\selectfont \textbf{(b)} No Occlusion $g(\cdot) > 0$};
      \draw (0.3,-2) node[inner sep=0, anchor=west] {\fontsize{3.0}{10}\selectfont \textbf{(c)} Occlusion $g(\cdot) = 0$};
      \draw (3.5,-2) node[inner sep=0, anchor=west] {\fontsize{3.0}{10}\selectfont \textbf{(d)} Occlusion $g(\cdot) < 0$};
      \draw (-3.0,-2.5) node[inner sep=0, anchor=west] {\fontsize{3.0}{10}\selectfont Pure Translation Movement at Different Baseline Distance };
      \draw (-5.0,-2.4-0.2) node[inner sep=0, anchor=west] {\fontsize{3.0}{10}\selectfont \lidar };
      \draw (4.4,-2.4-0.2) node[inner sep=0, anchor=west] {\fontsize{3.0}{10}\selectfont RGB Camera };
  \end{tikzpicture}
  \vspace{-4mm}
  \caption{
  \small 
  \textbf{Epipoplar Occlusion with Pure Translation Movement.} 
  A background pixel $\mathbf{p}_1$ is gradually occluded by a foreground pixel $\mathbf{p}_2$ under pure translation.
  We formulate three binocular systems at different baseline distances.
  For simplicity, we set the depth of $\mathbf{p}_1$ to infinity, making its 2D location static.
  Direction vector $\mathbf{n}$ is set to point from $\mathbf{p}$ to $\mathbf{q}$.
  The scalar $t$ is marked by \textcolor{cyan}{blue line segment}.
  From \cref{lemma:trans}, 
  \textbf{(a)} The pixel $\mathbf{p}_1$ is only occluded by pixel $\mathbf{p}_2$ at the reverse direction of epipolar line (\textcolor{red}{red} line segment).
  \textbf{(b)}  $\mathbf{p}_1$ is not occluded.
  \textbf{(c)}  $\mathbf{p}_1$ is occluded following \cref{eqn:def_occ}.
  \textbf{(d)}  $\mathbf{p}_1$ is considered occluded following \cref{eqn:judge_formal}.
  Intuitively, the occluded pixel $\mathbf{p}_1$ (in a) is ``caught up'' (in c) and ``surpassed'' (in d) by $\mathbf{p}_2$ while moving along the epipolar line.
 \vspace{-3mm} }
  \label{fig:trans}
  
\end{figure}

Suppose the relative camera motion among the \cref{fig:bev} binocular system is an arbitrary camera pose $\mathbf{P} = 
\begin{bmatrix}
    \mathbf{R} & \mathbf{t}
\end{bmatrix}
$.
Set $\idntyMat$ as the identity matrix. 
\methodName decomposes $\mathbf{P}$ into a sequential \textbf{pure rotation} and \textbf{pure translation} movement:
\begin{equation}
    \mathbf{P}_1 = \begin{bmatrix}
    \mathbf{R} & \mathbf{0}
\end{bmatrix}, \;
    \mathbf{P}_2 = \begin{bmatrix}
    \idntyMat & \mathbf{t}
\end{bmatrix}, \;
\mathbf{P} = \mathbf{P}_2 \mathbf{P}_1.
\end{equation}
We outline pure rotation movement in \cref{lemma:rot} and \cref{fig:rot}.
The pure translation movement is illustrated in \cref{lemma:trans} and \cref{fig:trans}.
Without loss of generality, we assume the epipoplar left and right cameras share the same intrinsic, \textit{i.e.}, $\mathbf{K}_l = \mathbf{K}_r$.
When with different intrinsics, we apply a resize and crop operation defined by $\mathbf{A} = \mathbf{K}_l\mathbf{K}_r^{-1}$ to unify the intrinsic.

\begin{lemma}
\textbf{Pure rotation movement}
 does not cause epipolar occlusion.
\label{lemma:rot}
\end{lemma}
We know from epipolar geometry \cite{hartley2003multiple} that pure rotation induces homography between two views.
In other words, there is a one-to-one mapping between pixels before and after rotation. See Supp for more information.
\cref{fig:rot} illustrates \cref{lemma:rot} with a visual example.

\begin{lemma}
\label{lemma:trans}
\textbf{Pure translation movement} causes epipolar occlusion along the direction of the epipolar line.
\end{lemma}

\begin{proof} 
As pure rotation does not cause epipoplar occlusion, we apply the translation movement over the artifacts-free depthmap after the pure rotation.
With slight abuse of the notation, we set the pixels after pure rotation movement as $\mathbf{p}$ and $\mathbf{q}$.
The epipolar line $\mathbf{l}_{\mathbf{p}}$ on the right image from the left image pixel $\mathbf{p}$ is: 
\begin{equation}
    \mathbf{l}_{\mathbf{p}} = \mathbf{K}^{\text{-}\intercal} [\mathbf{t}_{\times}]\mathbf{R} \mathbf{K}^{\text{-}1} \mathbf{p} = \mathbf{K}^{\text{-}\intercal} [\mathbf{t}_{\times}] \mathbf{K}^{\text{-}1} \mathbf{p},
    \label{eqn:overlap_epp}
\end{equation}
where $[\mathbf{t}_{\times}]$ is the cross product in matrix form~\cite{zhu2023lighteddepth}.
Due to a pure translation movement, the rotation $\mathbf{R}$ is removed as being an identity matrix.
Further, the left and right epipolar lines overlap with each other in pure translation:
\begin{equation}
    \mathbf{l}_{\mathbf{q}}  = -\mathbf{K}^{\text{-}\intercal} [\mathbf{t}_{\times}] \mathbf{K}^{\text{-}1} \mathbf{q}, \; {\mathbf{p}}^\intercal \mathbf{l}_{\mathbf{q}}=0, \; {\mathbf{p}}^\intercal \mathbf{l}_{\mathbf{p}} = 0.
    \label{eqn:pure_trans_overlap}
\end{equation}
Also shown in \cref{fig:trans}, pixel $\mathbf{q}$ is hence a function of pixel $\mathbf{p}$:
\begin{equation}
    \mathbf{q} = \mathbf{p} + t \cdot \mathbf{n}, \; t = \|\mathbf{q} - \mathbf{p}\|_2 = \mathbf{n}^\intercal(\mathbf{q} - \mathbf{p}), \; \mathbf{n} = (\mathbf{q} - \mathbf{p}) / \sqrt{\|\mathbf{q} - \mathbf{p}\|_2^2},
    \label{eqn:move_along_epp}
\end{equation}
where $t$ is the moving distance, similar to the disparity in stereo vision.
We define the vector $\mathbf{n}$ pointed from $\mathbf{p}$ to $\mathbf{q}$ as the direction of the epipoplar line $\mathbf{l}$.
$\mathbf{p}$ and $\mathbf{q}$ are related by the equation $d'\mathbf{q} = d \mathbf{K}_r \mathbf{R} \mathbf{K}_l^{\text{-}1} \mathbf{p}+ \mathbf{K}_r \mathbf{t}$. Expanding under pure translation movement gives us:
\begin{equation}
\small
    t^2 = (p_x - q_x)^2 + (p_y - q_y)^2  = \frac{(x + z \cdot q_x)^2 + (y + z \cdot q_y)^2}{(d' - z)^2} = \frac{(x + z \cdot q_x)^2 + (y + z \cdot q_y)^2}{d^2}.
    \label{eqn:start_point}
\end{equation}
Please see the derivation in Supp. (Appendix B). 
Note, we assume pixels $\mathbf{p}$ and $\mathbf{q}$ are respectively visible to the left and right camera.
This suggests $d > 0$ and $d = d' -z > 0$.
\cref{eqn:start_point} suggests moving distance $t^2$ \textbf{monotonously} decreases as the projected depth $d'$ increases.
It fits the intuition where a foreground object has a larger movement than a background object.
From \cref{eqn:def_occ}, when with occlusion, we have $\mathbf{q}_1 = \mathbf{q}_2, \; \text{and} \; d_{\mathbf{q}_1} < d_{\mathbf{q}_2}$.
Combine \cref{eqn:move_along_epp} and \cref{eqn:start_point}:
\begin{equation}
    \mathbf{q}_1 = \mathbf{q}_2, \; \text{and} \; d_{\mathbf{q}_1} < d_{\mathbf{q}_2} \longrightarrow t_1^2 > t_2^2 \longrightarrow \mathbf{n}^\intercal(\mathbf{p}_1 - \mathbf{p}_2) < 0.
    \label{eqn:search_start}
\end{equation}
From \cref{eqn:search_start}, if pixel $\mathbf{q}_1$ is occluded by $\mathbf{q}_2$ in the right view, pixel $\mathbf{q}_2$ must move a longer distance along the epipolar line than $\mathbf{q}_1$.
This suggests, in the left view, $\mathbf{p}_2$ resides at the ``back'' of $\mathbf{p}_1$ \textit{w.r.t.} the moving direction.
In other words, $\mathbf{p}_1$ is only occluded by other pixels along the reverse direction of epipolar line.
\end{proof}

\begin{figure}[!t]
    \centering
    \subfloat[Virtual \lidar Depthmap $\mathbf{D}_l$]{
    \includegraphics[width=0.48\linewidth]{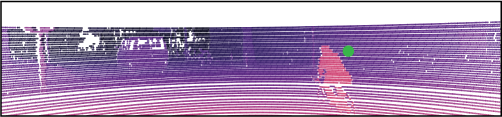}
    } \;
    \subfloat[Densified Virtual \lidar Depthmap $\overline{\mathbf{D}}_l$]{\includegraphics[width=0.48\linewidth]{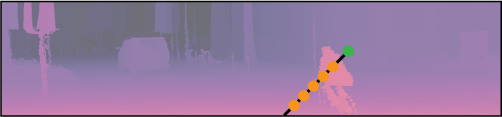}}
     \vspace{-1mm}
    \caption{
    \textbf{Per-pixel Occlusion Examination.}
    To overcome \lidar sparsity, for a \textcolor{ForestGreen}{query pixel}, we verify \textcolor{RedOrange}{inspection pixels} along epipolar line $\mathcal{L}_{\mathbf{p}_1}$ following \cref{eqn:judge_formal} to determine occlusion. 
    Depth $\overline{\mathbf{D}}_l$ is densified with nearest-neighbor interpolation.
    \vspace{-3mm}
    }
    \label{fig:pipeline}
\end{figure}

\noIndentHeading{Epipolar Occlusion Judgement.}
Combine \cref{lemma:rot} and \ref{lemma:trans}, a pixel $\mathbf{p}_1$ in \lidar view is occluded by other pixels $\mathbf{p}_2$ along the reverse of epipolar line direction.
For pixel $\mathbf{p}_1$, the set of inspection pixels is:
\begin{equation}
    \mathcal{L}_{\mathbf{p}_1} = \{ \mathbf{p}_2 \mid \mathbf{p}_2 = \mathbf{p}_1 - \mathbf{n} \cdot k \cdot \Delta t, \; k \in \mathbb{Z}  \}.
    \label{eqn:occ_set}
\end{equation}
$k$ is an integer.
At occlusion, we have $\mathbf{q}_1 = \mathbf{q}_2$, 
combined with \cref{eqn:move_along_epp}:
\begin{equation}
 \mathbf{q}_1 = \mathbf{q}_2 \longrightarrow \mathbf{p}_1 + t_1 \mathbf{n} =  \mathbf{p}_2 + t_2 \mathbf{n} \longrightarrow
\mathbf{n}^\intercal (\mathbf{p}_1 - \mathbf{p}_2) + t_1 - t_2 = 0.
\label{eqn:suffi}
\end{equation}
We use \cref{fig:trans} to explain \cref{eqn:suffi}.
In \cref{fig:trans}a, the pixel $\mathbf{p}_1$ and $\mathbf{p}_2$ has an initial distance $\mathbf{n}^\intercal (\mathbf{p}_1 - \mathbf{p}_2)$.
In \cref{fig:trans}c, after translation movement, each pixel moves a distance $t_1$ and $t_2$ along the epipolar line.
Occlusion happens when pixel $\mathbf{p}_2$ overlaps pixel $\mathbf{p}_1$.
We summarize \cref{eqn:suffi} into a function $g(\cdot)$ as:
\begin{equation}
 \lambda \leq g(\mathbf{p}_1, \mathbf{p}_2, d_{\mathbf{p}_1}, d_{\mathbf{p}_2} \mid \mathbf{K}, \mathbf{P}) \leq 0 ,
\label{eqn:judge}
\end{equation}
where the distance $t_1$ and $t_2$ are computed using intrinsic, camera motion, and depth values.
In \cref{fig:trans}b, when $g(\cdot) > 0$, there is no occlusion.
In \cref{fig:trans}c, when $g(\cdot) = 0$, two pixels overlap and occlusion happens.
However, this is hard due to \lidar sparsity, rasterization, and other noise sources.
Hence, we relax with a threshold $\lambda$.
In practice, we set $\lambda$ to $-\infty$.
This suggests the pixel $\mathbf{p}_1$ is occluded if any pixel $\mathbf{p}_2$ in the epipolar line ``surpasses'' it in the RGB view, as shown in \cref{fig:trans}d. 
Formally, we consider pixel $\mathbf{p}_1$ occluded as:
\begin{equation}
\mathcal{O} = \{\mathbf{p}_1 \mid \exists \; \mathbf{p_2} \! \in \!\mathcal{L}_{\mathbf{p}_1},  g(\mathbf{p}_1, \mathbf{p}_2, d_{\mathbf{p}_1}, d_{\mathbf{p}_2} \mid \mathbf{K}, \mathbf{P}) < 0 \}.
\label{eqn:judge_formal}
\end{equation}
\cref{eqn:judge_formal} is also a necessary condition of epipolar occlusion.
Due to the \lidar sparsity, we apply a nearest-neighbor interpolation over virtual \lidar depthmap $\mathbf{D}_l$ to acquire $d_{\mathbf{p}_2}$ whenever required, as shown in \cref{fig:pipeline}.
\subsection{Discussions}
\label{sec:discuss}




\noIndentHeading{Half-Occlusion~(Depth Buffering) Baseline.}
\cref{rel_work} shows that half-occlusion algorithms solve projective artifacts in stereo vision, \textit{i.e.}, stereo/epipolar occlusion, without knowing \lidar point cloud.
Despite a different setting, they provide a baseline solution.
When preparing RGB depthmap $\mathbf{D}_r$, only pixels with smaller depth are preserved if multiple 3D points $\mathbf{v}$ have duplicate 2D locations after rasterization.
The raw \lidar depthmap uses half-occlusion.
So, we refer Half-Occlusion as ``Raw'' following \cite{godard2019digging}.

\noIndentHeading{Does \methodName simply densify the sparse depthmap to remove occlusion?}
\textbf{No.} Densification alone is insufficient to remove occlusion. \methodName enables robust densification by formulating a stereo system with an \textit{auto-calibration} procedure to remove occlusion.

\noIndentHeading{Universal Existence of Projective Artifacts.}
\cref{fig:sensor} pictorially depicts that projective artifacts exist under all \lidar and camera sensor settings, while \cref{fig:trans} shows that these artifacts scale up with a larger baseline distance between sensors.
We now show that this indeed is the case.
The artifact width distribution plot on the outdoor \kitti and the indoor NYUv2 datasets are shown in Fig. 10 in the Supp.
The \lidar and RGB camera in \kitti have a baseline distance of $0.4m$ \cite{geiger2012we}.
We set the baseline distance to $0.1m$ following the AR headset configuration of \cref{fig:sensor}.
Despite the indoor image having a lesser resolution than outdoor ($480 \times 640$ compared to $375 \times 1242$), we observe a similar distribution. 
Detailed definitions and evaluation protocols are in Supp.
The analysis confirms that projective artifacts exist in both indoor and outdoor datasets.


\section{Experiments}\label{sec:exp}
We benchmark \methodName on \depthmap quality, \monoDE and \monoThreeD tasks.
\cref{sec:qual_assess} assesses depthmap quality, while \cref{sec:mono_depth_exp} and \cref{sec:mono_obj_det}. 
report \monoDE and \monoThreeD results respectively.

\noIndentHeading{Datasets.}
We experiment on \kitti \cite{geiger2013vision}, \nuscenes \cite{caesar2020nuscenes} and \waymo \cite{sun2020scalability} datasets. 
\kitti provides stereo ground truth while the others do not.

\noIndentHeading{Data Splits.}
We use the following splits of the above datasets:





\begin{itemize}
    \item \textit{\kitti Eigen}\cite{eigen2014depth}: $23,\!158$ training and $652$ validation samples of raw \kitti.
    \item \textit{\kitti Stereo}\cite{Menze2015CVPR}: $200$ training and $200$ testing samples out of which $145$ training scenes are mapped to the raw \kitti dataset.
    \item \textit{\nuscenes \frontal \val}: $28,\!130$ training and $2,\!436$ random validation samples (from $6,\!019$ samples) from front camera and top \lidar.
    \item \textit{\waymo \frontal \val}: $52,\!386$ training and $40,\!839$ validation samples from the front camera and top \lidar. We construct its training set by sampling every third frame from the training sequences as in \cite{kumar2022deviant}.
    \item \textit{\kitti Detection \val}: $3,\!748$ training and $3,\!762$ validation images from the left camera for the \monoThreeD task.
\end{itemize}

\noIndentHeading{Ground Truth (GT) \Depthmaps.} We use the following as GT \depthmaps:
\begin{itemize}
    \item \textit{\kitti Eigen}: We use the semi-dense \depthmap available in \kitti.
    \item \textit{\kitti Stereo}: We generate \depthmaps from GT disparity maps, which are denser than semi-dense \depthmaps.
    \item \textit{\nuscenes} and \textit{\waymo \frontal \val}: We generate GT \depthmaps by projecting only those \lidar points present inside GT \threeD bounding boxes. To prepare a reference depthmap, we follow~\cite{park2021pseudo} to filter \lidar point cloud with GT \threeD bounding boxes labels. Correspondingly, the evaluation is confined to the bounding box area, as only depth labels in that region are retained.
\end{itemize}

\subsection{\Depthmap Quality Assessment}\label{sec:qual_assess}

\noIndentHeading{Baselines.}
We compare \methodName to raw \depthmap, which is the half-occlusion (depth buffering) method and a modified half-occlusion method. 
The modified half-occlusion modifies the half-occlusion method (\cref{sec:discuss}) by back-projecting the estimated dense virtual \depthmap($\overline{\mathbf{D}}_l$) onto the RGB camera and only preserving pixels in $\mathbf{D}_r$ that are less than their back-projected values.

\begin{table*}[!t]
    \small
    \caption{\textbf{\Depthmap Quality Assessment} with baselines and \methodName. 
    We adopt KITTI semi-dense depthmap and KITTI Stereo disparity map as the reference GT.
    [Key: \firstKey{Best}, \secondKey{Second Best}, Mod. = Modified]
    }
    \label{tab:kitti_depth_quality}
    \centering
    \setlength\tabcolsep{0.05cm}
    \resizebox{\linewidth}{!}{
    \begin{tabular}{l|l|l|c c c c c c c c c}
        GT & Eval Region & Depthmap  & $\delta_{0.5}$$\uparrow$ & \textbf{$\delta_1$}$\uparrow$ & \textbf{$\delta_2$}$\uparrow$ & \rmse$\downarrow$ & \rmseLog$\downarrow$ & \absRel$\downarrow$ & \sqRel$\downarrow$ & \logTen$\downarrow$ & \siLog$\downarrow$\\
        \myTopRule
        \multirow{3}{*}{Semi-Dense} 
        & \multirow{3}{*}{All} & Raw & $0.973$ & $0.983$ & $0.990$ & $1.935$ & $0.087$ & $0.029$ & $0.332$ & $0.010$ & $8.660$\\
        & & Mod. Half-Occ & \second{0.983} & \second{0.991} & \second{0.996} & \second{1.347} & \second{0.057} & \second{0.019} & \second{0.113} & \second{0.007} & \second{5.712} \\
        & & \methodName & \first{0.985} & \first{0.994} & \first{0.998} & \first{1.241} & \first{0.046} & \first{0.016} & \first{0.063} & \first{0.007} & \first{4.592}\\
        \hline
        \multirow{9}{*}{Stereo} 
        & \multirow{3}{*}{All} & Raw & $0.950$ & $0.962$ & $0.970$ & $4.198$ & $0.169$ & $0.083$ & $1.604$ & $0.024$ & $16.691$\\
        & & Mod. Half-Occ & \second{0.971} & \second{0.982} & \second{0.988} & \second{2.541} & \second{0.105} & \second{0.047} & \second{0.546} & \second{0.016} & \second{10.340}\\
        & & \methodName & \first{0.975} & \first{0.986} & \first{0.991} & \first{2.224} & \first{0.086} & \first{0.034} & \first{0.354} & \first{0.014} & \first{8.362}\\
        \cline{2-12}
        & \multirow{3}*{Foreground} & Raw & $0.859$ & $0.871$ & $0.892$ & $8.726$ & $0.317$ & $0.228$ & $5.974$ & $0.058$ & $29.772$\\
        & & Mod. Half-Occ & \second{0.919} & \second{0.931} & \second{0.949} & \second{5.285} & \second{0.199} & \second{0.154} & \second{2.366} & \second{0.033} & \second{19.250}\\
        & & \methodName & \first{0.925} & \first{0.937} & \first{0.956} & \first{4.837} & \first{0.179} & \first{0.100} & \first{1.929} & \first{0.030} & \first{17.319}\\
        \cline{2-12}
        & \multirow{3}*{Background} & Raw & $0.967$ & $0.980$ & $0.986$ & $2.708$ & $0.113$ & $0.052$ & $0.774$ & $0.017$ & $11.037$\\
        & & Mod. Half-Occ & \second{0.980} & \second{0.991} & \second{0.995} & \second{1.701} & \second{0.070} & \second{0.032} & \second{0.210} & \second{0.013} & \second{6.638}\\
        & & \methodName & \first{0.982} & \first{0.993} & \first{0.998} & \first{1.439} & \first{0.052} & \first{0.027} & \first{0.085} & \first{0.012} & \first{4.722}\\
    \end{tabular}
    }
\end{table*}
\begin{table*}[!t]
    \small
    \caption{\textbf{\Depthmap Quality Assessment on Removed Artifacts}.
    \methodName outperforms the Mod. half-occlusion baseline.
    For removed artifacts, a worse performance suggests a better algorithm.
    [Key: \firstKey{Best}]
    }
    \label{tab:kitti_depth_quality_inv_main}
    \centering
    \setlength\tabcolsep{0.05cm}
    \resizebox{\linewidth}{!}{
    \begin{tabular}{l|l|l|c c c c c c c c}
        GT & Eval Region & Depthmap  & $\delta_{0.5}$$\downarrow$ & \textbf{$\delta_1$}$\downarrow$ & \textbf{$\delta_2$}$\downarrow$ & \rmse$\uparrow$ & \rmseLog$\uparrow$ & \absRel$\uparrow$ & \sqRel$\uparrow$ & \logTen$\uparrow$\\
        \myTopRule
        \multirow{2}{*}{Semi-Dense} 
        & \multirow{2}{*}{All} & Mod. Half-Occ & $0.814$ & $0.846$ & $0.893$ & $5.850$ & $0.275$ & $0.196$ & $3.762$ & $0.053$ \\
        & & \methodName & \first{0.274} & \first{0.363} & \first{0.548} & \first{12.084} & \first{0.587} & \first{0.778} & \first{15.851} & \first{0.202}\\
    \end{tabular}
    }
    \vspace{-0.5cm}
\end{table*}

\noIndentHeading{Results.}
\cref{tab:kitti_depth_quality} compares the \methodName depthmap with these baselines processed \depthmaps. 
\cref{tab:kitti_depth_quality} verifies that \methodName \depthmap outperforms the raw and modified half-occlusion baselines.
\methodName \depthmap agrees more with semi-dense or disparity depthmap after removing the projective artifacts.
\cref{fig:compare_kitti} shows qualitative comparisons.
We believe projective artifacts persist in the half-occlusion method because of the \lidar sparsity, leading to inferior \depthmaps.

\noIndentHeading{Foreground Improvement.}
\cref{tab:kitti_depth_quality} rows $5$ and $6$ shows a significant improvement in foreground depthmap quality.
On metric $\delta_{0.5}$, we reduce the errors by $46.8\% = 1 - \frac{1-0.925}{1-0.859}$.
This aligns with our observation that projective artifacts mostly occur over foreground objects, confirming \cref{fig:teaser}.

\noIndentHeading{Comparison on Removed Artifacts.}
We next compare \methodName on removed artifacts in \cref{tab:kitti_depth_quality_inv_main}.
\cref{tab:kitti_depth_quality_inv_main} shows that that in-addition to removing some artifacts, the modified half-occlusion also removes some non-artifacts (good points). 
Ideally, the removed points should be artifacts; therefore, the best algorithm should perform the worst on the quality assessment of removed points and hence the reversal in the directional indicator for \cref{tab:kitti_depth_quality_inv_main}. 
Qualitative comparison between raw, modified half-occlusion and \methodName is provided in Supp. (Fig. 11).

\noIndentHeading{Auto-Calibration Ablation.}
We next conduct ablations of the Auto-Calibration module (\cref{sec:lidar_rgb}) in 
\cref{tab:ablation_autocalib}. \cref{tab:ablation_autocalib} confirms the effectiveness of Auto-Calibration module in \methodName, since  Auto-Calibration provides $15\%$ and $19\%$ relative improvement in \absRel and \siLog metrics respectively.

\begin{table*}[!t]
    \small
    \caption{\textbf{Auto-Calibration Ablation} on \Depthmap Quality Assessment with semi-dense GT \depthmap.
    [Key: \firstKey{Best}]
    }
    \label{tab:ablation_autocalib}
    \centering
    \setlength\tabcolsep{0.05cm}
    \resizebox{0.9\linewidth}{!}{
    \begin{tabular}{c|c|c c c c c c c c c}
        \Depthmap & AutoCalib & $\delta_{0.5}$$\uparrow$ & \textbf{$\delta_1$}$\uparrow$ & \textbf{$\delta_2$}$\uparrow$ & \rmse$\downarrow$ & \rmseLog$\downarrow$ & \absRel$\downarrow$ & \sqRel$\downarrow$ & \logTen$\downarrow$ & \siLog$\downarrow$\\
        \myTopRule
        \multirow{2}*{\methodName} & \xmark & $0.981$ & $0.990$ & $0.995$ & $1.473$ & $0.057$ & $0.019$ & $0.111$ & $0.007$ & $5.680$\\
         & \cmark & \first{0.985} & \first{0.994} & \first{0.998} & \first{1.241} & \first{0.046} & \first{0.016} & \first{0.063} & \first{0.007} & \first{4.592}\\
    \end{tabular}
    }
\end{table*}

\subsection{\monoDE: Monocular Depth Estimation }\label{sec:mono_depth_exp}

\noIndentHeading{Baselines.}
We compare \methodName to raw and Semi-Dense methods.
We use ZoeDepth~\cite{bhat2023zoedepth} as the depth model due to its superior generalization. 

\begin{table*}[!t]
     \small
    \caption{\textbf{\monoDE Results on KITTI Stereo.} ZoeDepth~\cite{bhat2023zoedepth} trained with \methodName \depthmaps outperforms raw and semi-dense \cite{uhrig2017sparsity} \depthmaps on foreground.
    [Key: \firstKey{Best}, \secondKey{Second Best}].
    }
    \label{tab:kitti_zoe_stereo_eval}
    \centering
    \resizebox{0.9\linewidth}{!}{
    \begin{tabular}{l|l|c c c c c c c c c}
        Eval Region & Depthmap & \textbf{$\delta_1$}$\uparrow$ & \textbf{$\delta_2$}$\uparrow$ & \textbf{$\delta_3$}$\uparrow$ & \rmse$\downarrow$ & \rmseLog$\downarrow$ & \absRel$\downarrow$ & \sqRel$\downarrow$ & \logTen$\downarrow$ & \siLog$\downarrow$\\
        \myTopRule
        \multirow{3}{*}{All} & Raw & $0.955$ & $0.982$ & $0.992$ & $2.647$ & $0.116$ & $0.067$ & $0.420$ & $0.026$ & $11.324$\\
         & Semi-Dense & \first{0.977} & \first{0.994} &\first{ 0.997} & \first{2.274} & \first{0.082} & \first{0.043} & \first{0.201} & \first{0.018} & \first{7.971}\\
         & \methodName & \second{0.974} & \second{0.992} & \second{0.996} & \second{2.288} & \second{0.089} & \second{0.049} & \second{0.248} & \second{0.020} & \second{8.588}\\
         \hline
        \multirow{3}{*}{Foreground} & Raw & $0.886$ & $0.947$ & $0.980$ & $3.692$ & $0.180$ & $0.127$ & $1.112$ & $0.022$ & $16.518$\\
         & Semi-Dense & \first{0.976} & \first{0.990} & \first{0.994} & \first{2.093} & \first{0.091} & \first{0.056} & \first{0.344} & \first{0.022} & \first{8.342}\\
         & \methodName & \second{0.945} & \second{0.978} & \second{0.989} & \second{2.939} & \second{0.128} & \second{0.081} & \second{0.642} & \second{0.029} & \second{11.798}\\
         \hline
        \multirow{3}{*}{Background} & Raw & $0.973$ & $0.993$ & $0.998$ & $2.294$ & $0.084$ & $0.048$ & $0.188$ & $0.020$ & $7.981$\\
         & Semi-Dense & \second{0.978} & \second{0.996} & \second{0.998} & \second{2.283} & \second{0.076} & \second{0.040} & \second{0.167} & \first{0.017} & \second{7.239}\\
         & \methodName & \first{0.981} & \first{0.997} & \first{0.999} & \first{2.084} & \first{0.070} & \first{0.039} & \first{0.138} & \first{0.017} & \first{6.681}\\
    \end{tabular}
    }
\end{table*}

\begin{figure*}[!t]
    \centering
    \begin{subfigure}[]{0.33\textwidth}
        \centering
        \includegraphics[width=\linewidth]{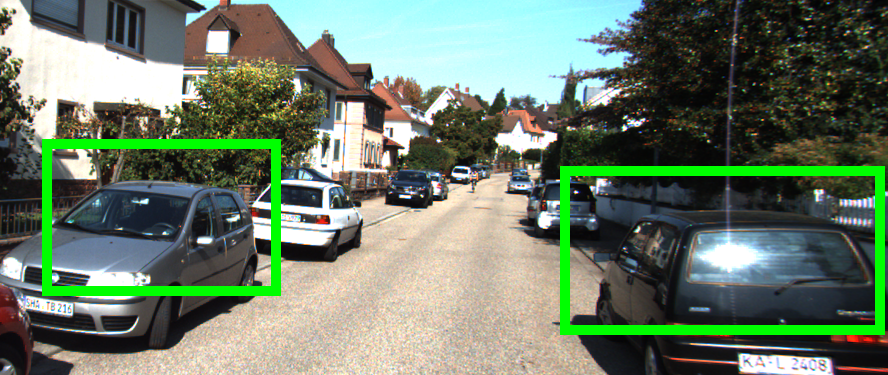}
        \caption{RGB Input}
    \end{subfigure}%
    \hfill
    \begin{subfigure}[]{0.33\textwidth}
        \centering
        \includegraphics[width=\linewidth]{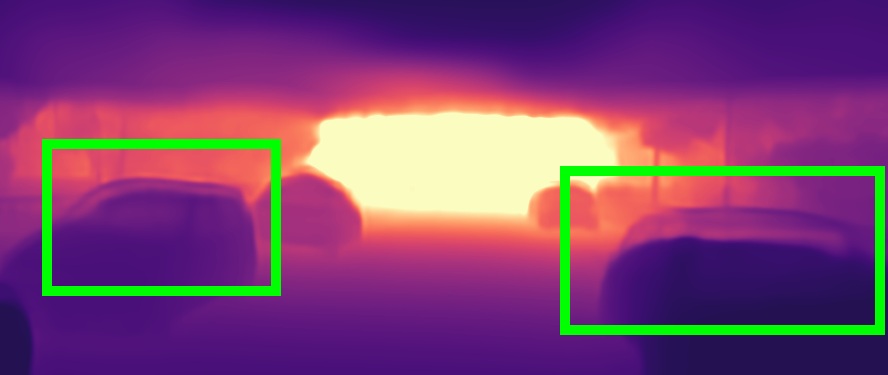}
        \caption{Raw \depthmaps}
    \end{subfigure}%
    \hfill
    \begin{subfigure}[]{0.33\textwidth}
        \centering
        \includegraphics[width=\linewidth]{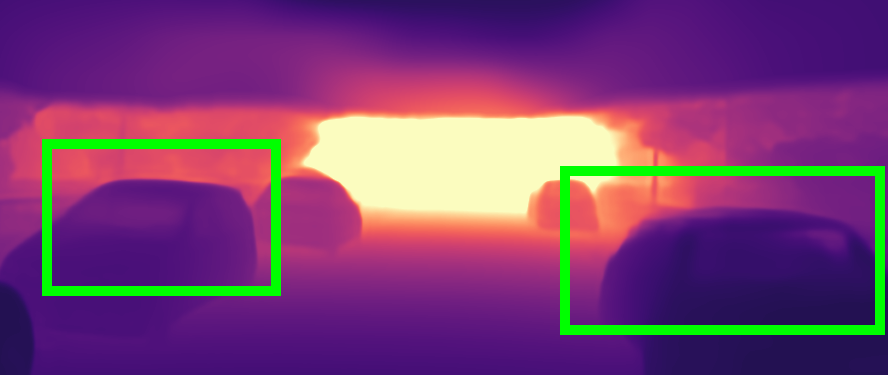}
        \caption{\methodName \depthmaps}
    \end{subfigure}%
    \caption{\textbf{\monoDE Qualitative Results} on \kitti Eigen dataset.
    ZoeDepth~\cite{bhat2023zoedepth} trained with \methodName \depthmaps (c) outperforms the one with raw \depthmaps (b), especially on foreground.}
    \label{fig:monode_vis}
\end{figure*}

\noIndentHeading{KITTI Stereo and Eigen Results.}
We train \zoeDepth\cite{bhat2023zoedepth} from scratch using depthmap with and without applying \methodName on \kitti Eigen split. 
We evaluate on \kitti Stereo and Eigen split in \cref{tab:kitti_zoe_stereo_eval,tab:zoe_kitti_waymo_nuscenes} respectively.
\cref{tab:kitti_zoe_stereo_eval} shows that ZoeDepth trained with \methodName depthmap significantly outperforms the baselines. 
We observe despite semi-dense GT \cite{uhrig2017sparsity} using additional stereo cameras, \methodName outperforms Semi-Dense especially in the background. This may be due to \cite{uhrig2017sparsity} aggressively removing the background pixels around the foreground object boundaries.
\cref{fig:monode_vis} shows this result visually.
ZoeDepth trained with the raw depthmap learns strip-shaped artifacts (\textcolor{ForestGreen}{green boxes}).

\noIndentHeading{Foreground Improvement.}
We additionally ablate depth performance on foreground, background, and entire image using the GT segmentation mask in \cref{tab:kitti_zoe_stereo_eval}.
\cref{tab:kitti_zoe_stereo_eval} shows that ZoeDepth trained with \methodName depthmap outperforms in the foreground area, which agrees with the conclusion of \cref{sec:qual_assess}.

\noIndentHeading{\nuscenes and \waymo \frontal \val Results.}
We benchmark \monoDE on two datasets without stereo sensors: \nuscenes~\cite{caesar2020nuscenes} and \waymo~\cite{sun2020scalability} in \cref{tab:zoe_kitti_waymo_nuscenes}. \cref{tab:zoe_kitti_waymo_nuscenes} confirms that \methodName achieves significant improvements on these datasets.

\noIndentHeading{Effect of Environmental Conditions.}
We next analyze the effect of environmental conditions on the \nuscenes \val split in \cref{tab:zoe_nuscenes_weather}.
We categorize the \nuscenes\cite{caesar2020nuscenes} val set into rainy, night, and sunny following their description tag (\nuscenes-devkit GitHub \#124).
\cref{tab:zoe_nuscenes_weather} shows that \methodName is robust across multiple conditions.


\begin{table}[!tb]
     \small
    \caption{\textbf{\monoDE Results}. 
    ZoeDepth trained with \methodName outperforms the baselines in foreground regions of all datasets. 
    [Key: \firstKey{Best}]
    }
    \label{tab:zoe_kitti_waymo_nuscenes}
    \centering
    \setlength\tabcolsep{0.05cm}
    \resizebox{\linewidth}{!}{
    \begin{tabular}{l|l|l|c c c c c c c c c}
        Data Split & Eval Region & Depthmap & $\delta_1$$\uparrow$ & $\delta_2$$\uparrow$ & $\delta_{3}$$\uparrow$ & \rmse$\downarrow$ & \rmseLog$\downarrow$ & \absRel$\downarrow$ & \sqRel$\downarrow$ & \logTen$\downarrow$ & \siLog$\downarrow$\\
        \myTopRule
        \multirow{2}*{\kitti Eigen} & \multirow{2}*{All} & Raw & $0.869$ & $0.959$ & $0.987$ & $2.911$ & $0.163$ & $0.124$ & $0.606$ & $0.046$ & $13.501$\\
        && \methodName & \first{0.927} & \first{0.985} & \first{0.997} & \first{2.185} & \first{0.110} & \first{0.082} & \first{0.287} & \first{0.033} & \first{8.674}\\
        \hline
        \multirow{2}*{\nuscenes \val} & \multirow{2}*{Foreground} & Raw & $0.971$ & $0.987$ & $0.993$ & $1.347$ & $0.075$ & $0.049$ & $0.245$ & $0.018$ & $6.339$\\
         && \methodName & \first{0.981} & \first{0.993} & \first{0.996} & \first{1.150} & \first{0.059} & \first{0.039} & \first{0.137} & \first{0.016} & \first{4.905}\\
        \hline
        \multirow{2}*{\waymo \val} & \multirow{2}*{Foreground} & Raw & 0.941 & 0.988 & 0.996 & 3.468 & 0.111 & 0.079 & 0.487 & 0.032 & 9.116\\
        && \methodName & \first{0.962} & \first{0.993} & \first{0.998} & \first{3.057} & \first{0.091} & \first{0.063} & \first{0.367} & \first{0.026} & \first{7.322}\\
    \end{tabular}
    }
\end{table}

\begin{table}[!t]
    \small
    \caption{\textbf{\monoDE Analysis to Environmental Conditions} on \nuscenes Val split. ZoeDepth trained with \methodName \depthmaps shows consistent improvements on foreground under all environmental conditions. 
    [Key: \firstKey{Best}]
    }
    \label{tab:zoe_nuscenes_weather}
    \centering
    \setlength\tabcolsep{0.1cm}
    \resizebox{\linewidth}{!}{
    \begin{tabular}{l|c|l|c c c c c c c c c}
        Condition & Samples & Depthmap & $\delta_1$$\uparrow$ & $\delta_2$$\uparrow$ & $\delta_{3}$$\uparrow$ & \rmse$\downarrow$ & \rmseLog$\downarrow$ & \absRel$\downarrow$ & \sqRel$\downarrow$ & \logTen$\downarrow$ & \siLog$\downarrow$\\
        \myTopRule
        \multirow{2}*{Rain} & \multirow{2}{*}{864} & Raw & $0.984$ & $0.991$ & $0.995$ & $1.338$ & $0.058$ & $0.040$ & $0.163$ & $0.016$ & $4.859$\\
        & & \methodName & \first{0.989} & \first{0.993} & \first{0.995} & \first{1.222} & \first{0.050} & \first{0.034} & \first{0.127} & \first{0.015} & \first{4.110}\\
        \hline
        \multirow{2}*{Night} & \multirow{2}{*}{533} & Raw & $0.965$ & $0.982$ & $0.990$ & $1.347$ & $0.089$ & $0.059$ & $0.335$ & $0.021$ & $7.481$\\
        & & \methodName & \first{0.978} & \first{0.991} & \first{0.995} & \first{1.063} & \first{0.067} & \first{0.045} & \first{0.172} & \first{0.017} & \first{5.374}\\
        \hline
        \multirow{2}*{Sunny} & \multirow{2}{*}{1039} & Raw & $0.964$ & $0.987$ & $0.993$ & $1.352$ & $0.081$ & $0.053$ & $0.267$ & $0.019$ & $6.985$\\
        & & \methodName & \first{0.976} & \first{0.993} & \first{0.996} & \first{1.134} & \first{0.062} & \first{0.040} & \first{0.128} & \first{0.016} & \first{5.324}\\
        \hline
        \multirow{2}*{All \cref{tab:zoe_kitti_waymo_nuscenes}} & \multirow{2}{*}{2436} & Raw & $0.971$ & $0.987$ & $0.993$ & $1.347$ & $0.075$ & $0.049$ & $0.245$ & $0.018$ & $6.339$\\
        & & \methodName & \first{0.981} & \first{0.993} & \first{0.996} & \first{1.150} & \first{0.059} & \first{0.039} & \first{0.137} & \first{0.016} & \first{4.905}\\
    \end{tabular}
    }
    \vspace{-0.4cm}
\end{table}

\subsection{\monoThreeD: Monocular 3D Object Detection}
\label{sec:mono_obj_det}
We finally verify the impact of \methodName on the \monoThreeD task.

\noIndentHeading{Baseline.}
We compare \methodName to raw method.
We take \sota \threeD detectors, \caddn \cite{reading2021categorical} and \monodtr \cite{huang2022monodtr}, to showcase the impact of cleaning projected \lidar \depthmap with \methodName, since these detectors use depthmap as an auxiliary training task. 
The two methods cover both convolutional and transformer architectures, making our benchmarking comprehensive.
However, these two \sota detectors use depth supervision in training differently.
\begin{wraptable}{R}{0.55\linewidth}
    \vspace{-5mm}
    \caption{
    \textbf{\kitti \val \monoThreeD Results} on Car category.
    Using \methodName \depthmaps in training boosts the \monoThreeD performance.
    [Key: \firstKey{Best}, \secondKey{Second Best}, $^*$ = Reported, $\dagger$= Retrained, Mod= Moderate.]
    }
    \label{tab:kitti_val_detection}
    \centering
    \scalebox{\scaleFraction}{
    \setlength\tabcolsep{0.1cm}
    \begin{tabular}{l| c |c c c}
    \multirow{3}{*}{Method} & \multirow{3}{*}{Depthmap} & \multicolumn{3}{c}{\iouThreeD $> 0.7$}\\
    & & \multicolumn{3}{c}{\apThreeDForty\bracketPercentage$\uparrow$}\\
    && Easy & Mod & Hard \\
    \myTopRule
    \mThreeDrpn \cite{brazil2019m3d} & \mathDash & $14.53$ & $11.07$ & $8.65$\\
    \monopair \cite{chen2020monopair} & \mathDash & $16.28$ & $12.30$ & $10.42$\\
    \monodle \cite{ma2021delving} & \mathDash & $17.45$ & $13.66$ & $11.68$\\
    \kinematicVideo \cite{brazil2020kinematic} & \mathDash & $19.76$ & $14.10$ & $10.47$\\
    GrooMeD\cite{kumar2021groomed} & \mathDash & $19.67$ & $14.32$ & $11.27$\\
    \monorun \cite{chen2021monorun} & \mathDash & $20.02$ & $14.65$ & $12.61$\\
    \dFourLCN \cite{ding2020learning} & Raw$^*$ & $22.32$ & $16.20$ & $12.30$\\
    \dfrNet \cite{zou2021devil} & Raw$^*$ & $24.81$ & $17.78$ & $14.41$\\
    \hline
    \multirow{2}{*}{\caddn \cite{reading2021categorical}} & Raw$^\dagger$ & \second{21.60} & \second{15.67} & \first{13.24}\\ 
    &\methodName & \first{22.88} & \first{15.70} & \second{13.18}\\
    \hline
    \multirow{2}{*}{\monodtr \cite{huang2022monodtr}} & Raw$^\dagger$ & 24.50 & \second{18.66} & \first{15.69}\\
    &\methodName &  \first{26.15} & \first{18.77} & \first{15.69}\\
    \end{tabular}     
    }
\end{wraptable}

\noIndentHeading{\caddn .}
\caddn uses the densified raw \depthmaps of~\cite{ku2018defense}.
For \caddn experiments, we apply \methodName to the projected \depthmaps and then densify those maps using the same depth completion algorithm~\cite{ku2018defense} and parameters of \cite{reading2021categorical}. 

\noIndentHeading{\monodtr .}
\monodtr~\cite{huang2022monodtr} pre-computes their GT depthmaps by $4\times$ downsampling raw projected \depthmaps.
For \monodtr experiments, we clean the projected \depthmaps and then $4\times$ downsampling.

\noIndentHeading{Results.}
\cref{tab:kitti_val_detection} shows results of retraining both detectors with our \depthmaps on \kitti \val split. 
Cleaned \depthmaps improves the performance of \caddn and \monodtr on most of the metrics.
The biggest gains appear in the Easy category. 
This is expected as Easy cars are closer to the ego camera and receive the maximum number of \lidar points. 
Hence, improving the projected \depthmap by \methodName benefits the Easy category more.

\section{Conclusion}
We propose an analytical solution to remove projective artifacts in \lidar depthmap.
Our parameter-free method, \methodName, applies to diverse \lidar and RGB camera sensor settings.
\methodName only requires the relative position between \lidar and RGB camera.
We verify its benefits with \textbf{unanimous} improvements on SoTA 
\monoDE and \monoThreeD methods.
We release processed depthmaps of five major AV datasets to benefit the community.


\noIndentHeading{Limitation.}
We use nearest-neighbor interpolation to densify virtual \lidar \depthmaps.  
Advanced strategies, \forExample, depth completion, may improve accuracy.

\clearpage
{\small
\bibliographystyle{splncs04}
\bibliography{main}
}

\clearpage
\appendix
\section*{\centering \methodName: \textcolor{ForestGreen}{Re}move \textcolor{ForestGreen}{P}rojective \textcolor{ForestGreen}{L}iDAR \Depthmap \textcolor{ForestGreen}{A}rtifacts via Exploiting Epipolar Geometr\textcolor{ForestGreen}{y}}\
\begin{center}
    \large{Supplementary Material}
\end{center}

\setcounter{page}{19}

\section{Proof of \cref{collary_lidar}}
\label{sec:lemma1_proof}
\begin{proof}
Let $\mathbf{v}_i \in \setOfThreeDPts$ be a 3D point from the set of 3D points($\setOfThreeDPts)$ scanned by the \lidar and $\leftPixel_i = \begin{bmatrix}
    u & v & 1
\end{bmatrix}^\transpose$ be its projection then,
\begin{equation}
    \begin{aligned}
        \pi (\mathbf{v}_i \mid \intrinsic)  = \intrinsic\rotMat\mathbf{v}_i + \intrinsic\transVec = d_i\leftPixel_i
    \end{aligned}
\end{equation}
where $\intrinsic$ is the intrinsic, $\begin{bmatrix}
    \rotMat & \transVec
\end{bmatrix}$ is the extrinsic and $d$ is the depth of the pixel $\leftPixel$.  Since the virtual camera is situated at the \lidar's location, $\transVec = 0$ in the extrinsic matrix, and thus we get
\begin{equation}
    \begin{aligned}
        d_i\leftPixel_i = \intrinsic\rotMat\mathbf{v}_i
    \end{aligned}
\end{equation}
We know that $\intrinsic$ and $\rotMat$ are full rank matrices and since $d_i$ is scalar, we can say that there is a one-to-one mapping between $\mathbf{v}_i$ and $\leftPixel_i$ in the virtual camera setup. In other words, for the sensor pair of \lidar and virtual camera we have:

\begin{equation}
    \begin{aligned}
        \forall \mathbf{v}_1, \mathbf{v}_2 \in \mathcal{V}, \;\Vert\pi(\mathbf{v}_1 \mid \intrinsic)-\pi(\mathbf{v}_2 \mid \intrinsic)\Vert_2 = \Vert\leftPixel_1 - \leftPixel_2\Vert_2 = 0 \\ 
        \text{if} \; \Vert\mathbf{v}_1 - \mathbf{v_2}\Vert_2 = 0
    \end{aligned}
\end{equation}

\end{proof}

\section{Proof of \cref{lemma:trans}}
\label{sec:trans_proof}
\begin{proof}
    Let $\leftPixel' = \begin{bmatrix}
    \leftPixelSym_\xcoord' & \leftPixelSym_\ycoord' & 1
\end{bmatrix}^\transpose$ and $\rightPixel = \begin{bmatrix}
    \rightPixelSym_\xcoord & \rightPixelSym_\ycoord & 1
\end{bmatrix}^\transpose$. Substituting this in \cref{eqn:move_along_epp} we get
\begin{equation}
    \begin{aligned}
        t &= \|\leftPixel' - \rightPixel\|_2 \\
                 &= \|\begin{bmatrix}
        \leftPixelSym_\xcoord' & \leftPixelSym_\ycoord' & 1
        \end{bmatrix}^\transpose - \begin{bmatrix}
        \rightPixelSym_\xcoord & \rightPixelSym_\ycoord & 1
        \end{bmatrix}^\transpose\|_2 \\
        t &= \sqrt{(\leftPixelSym_\xcoord' - \rightPixelSym_\xcoord)^2 + (\leftPixelSym_\ycoord' - \rightPixelSym_\ycoord)^2}
    \end{aligned}
    \label{eqn:mag_eqn}
\end{equation}

\noindent Let us set the depth of $\leftPixel'$ and $\rightPixel$ as $d'$ and $d$ respectively, then:
\begin{equation}
    \small
    \begin{aligned}
    d \rightPixel &= d' \intrinsic_\stereoRight \rotMat \intrinsic_\stereoLeft^{\text{-}1} \leftPixel' + \intrinsic_\stereoRight \transVec \\
    d \begin{bmatrix}
        \rightPixelSym_\xcoord & \rightPixelSym_\ycoord & 1
    \end{bmatrix}^\transpose &= d' \intrinsic_\stereoRight \rotMat \intrinsic_\stereoLeft^{\text{-}1}\begin{bmatrix}
        \leftPixelSym'_\xcoord & \leftPixelSym'_\ycoord & 1
    \end{bmatrix}^\transpose + \intrinsic_\stereoRight \transVec
    \end{aligned}
\end{equation}

\noindent Applying pure translation conditions, \thatIs, $\rotMat=\idntyMat$, we get:
\begin{align}
    \rightPixelSym_\xcoord = \frac{d'\leftPixelSym_\xcoord' + \xcoord}{d' + \zcoord}, \;  \rightPixelSym_\ycoord = \frac{d'\leftPixelSym_\ycoord' + \ycoord}{d' + \zcoord}
\end{align}

\noindent We then calculate $\rightPixelSym_\xcoord - \leftPixelSym_\xcoord'$ as follows:
\begin{equation}
    \begin{aligned}
        \rightPixelSym_\xcoord - \leftPixelSym_\xcoord' &= \frac{d'\leftPixelSym_\xcoord' + \xcoord}{d' + \zcoord} - \leftPixelSym_\xcoord'\\
        &= \frac{d'\leftPixelSym_\xcoord' + \xcoord - d'\leftPixelSym_\xcoord' - z\leftPixelSym_
        \xcoord'}{d' + \zcoord}\\
        &= \frac{\xcoord - z\leftPixelSym_
        \xcoord'}{d' + \zcoord}\\
    \end{aligned}
    \label{eqn:diff1}
\end{equation}

\noindent Similarly, we get:
\begin{equation}
    \rightPixelSym_\ycoord - \leftPixelSym_\ycoord' = \frac{\ycoord - z\leftPixelSym_
        \ycoord'}{d' + \zcoord}
    \label{eqn:diff2}
\end{equation}

\noindent Combining \cref{eqn:mag_eqn}, \cref{eqn:diff1} and \cref{eqn:diff2}, we get:
\begin{equation}
    t^2 = \frac{(\xcoord - z\leftPixelSym_
        \xcoord')^2 + (\ycoord - z\leftPixelSym_
        \ycoord')^2}{(d' + \zcoord)^2}
\end{equation}

\end{proof}

\section{Existence of Projective Artifacts}
\begin{figure}[H]
    \centering
    \subfloat[KITTI~\cite{kitti2012benchmark} at baseline 0.4 meter.]{
    \begin{tikzpicture}
        \draw (0,0) node[inner sep=0]{\includegraphics[width=0.45\linewidth]{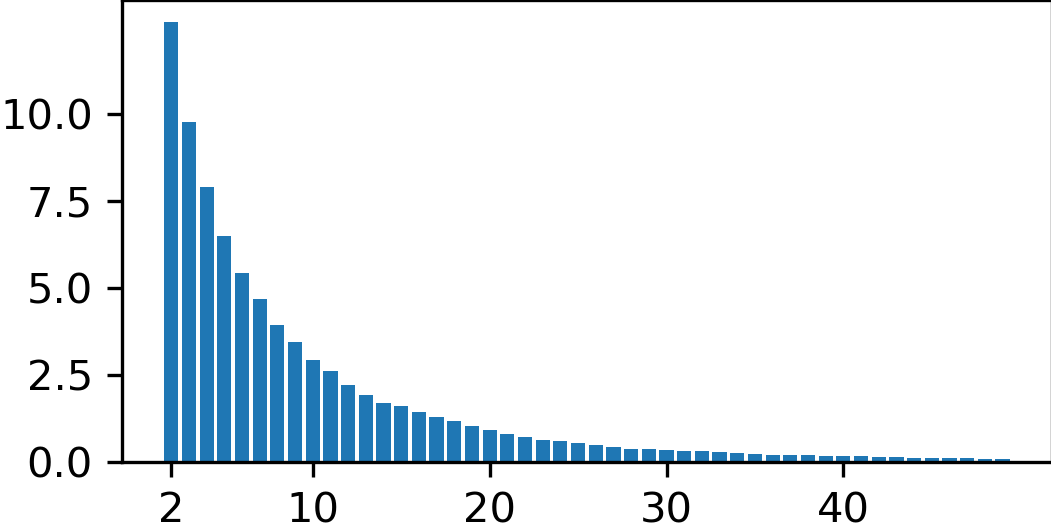}};
        \draw (-1.5,-1.6) node[inner sep=0, anchor=west] {\fontsize{3.0}{10}\selectfont Projective Artifacts Width in Pixels};
        \draw (-2.8,-0.5) node[inner sep=0, anchor=west, rotate=90] {\fontsize{3.0}{10}\selectfont Percentile};
        \label{fig:artifact_exist_kitti}
    \end{tikzpicture}
    } 
    \subfloat[NYUv2~\cite{silberman2012indoor} at baseline 0.1 meter.]{
    \begin{tikzpicture}
        \draw (0,0) node[inner sep=0]{\includegraphics[width=0.45\linewidth]{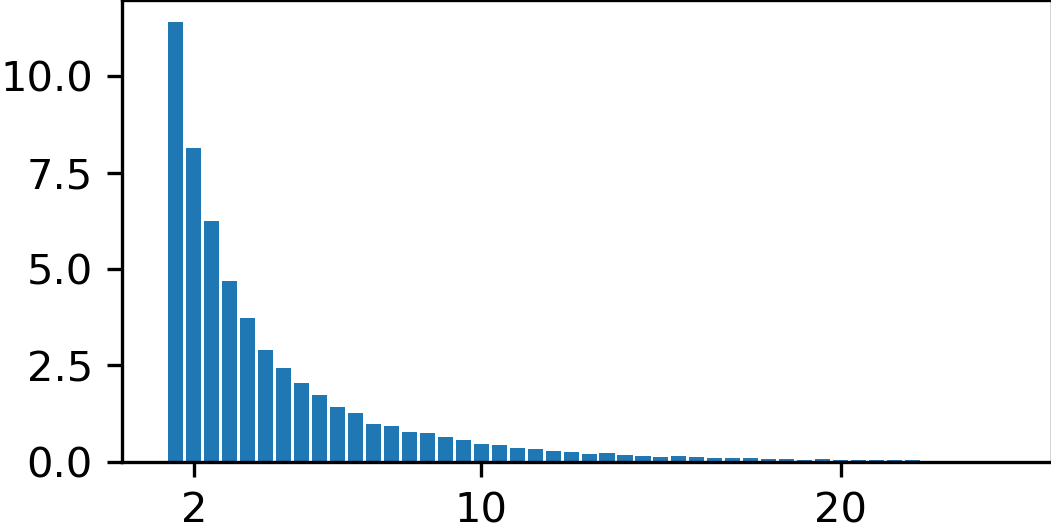}};
        \draw (-1.5,-1.6) node[inner sep=0, anchor=west] {\fontsize{3.0}{10}\selectfont Projective Artifacts Width in Pixels};
        \draw (-2.8,-0.5) node[inner sep=0, anchor=west, rotate=90] {\fontsize{3.0}{10}\selectfont Percentile};
        \label{fig:artifact_exist_scannet}
    \end{tikzpicture}
    }
    \vspace{-1mm}
    \caption{
    \textbf{Universal Existence of Projective Artifacts} in outdoor and indoor scenes.
    We score the projective artifact by its width along epipolar line.
    A larger score suggests more pronounced artifacts.
    We follow popular \lidar and camera sensor settings in analysis.
    For outdoor, we follow \kitti. 
    For indoor, we follow \cref{fig:sensor}b AR headset.
     \vspace{-3mm}}
    \label{fig:artifact_exist}
\end{figure}
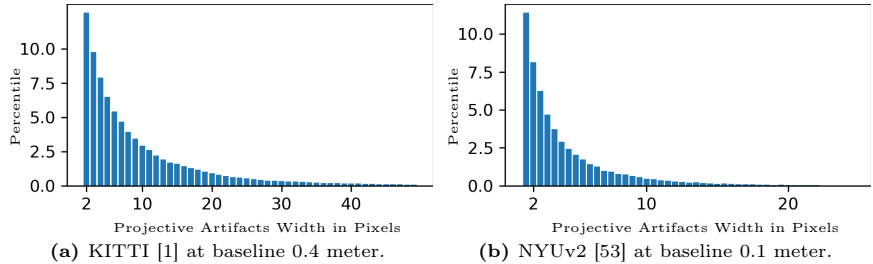

\newpage
\section{Additional Experiments and Results}


\subsection{Modified Half-Occlusion Baseline} \label{sec:halfocc_mod}

\begin{figure}[!t]
    \centering
    \begin{subfigure}[]{0.32\linewidth}
        \centering
        \includegraphics[width=\linewidth]{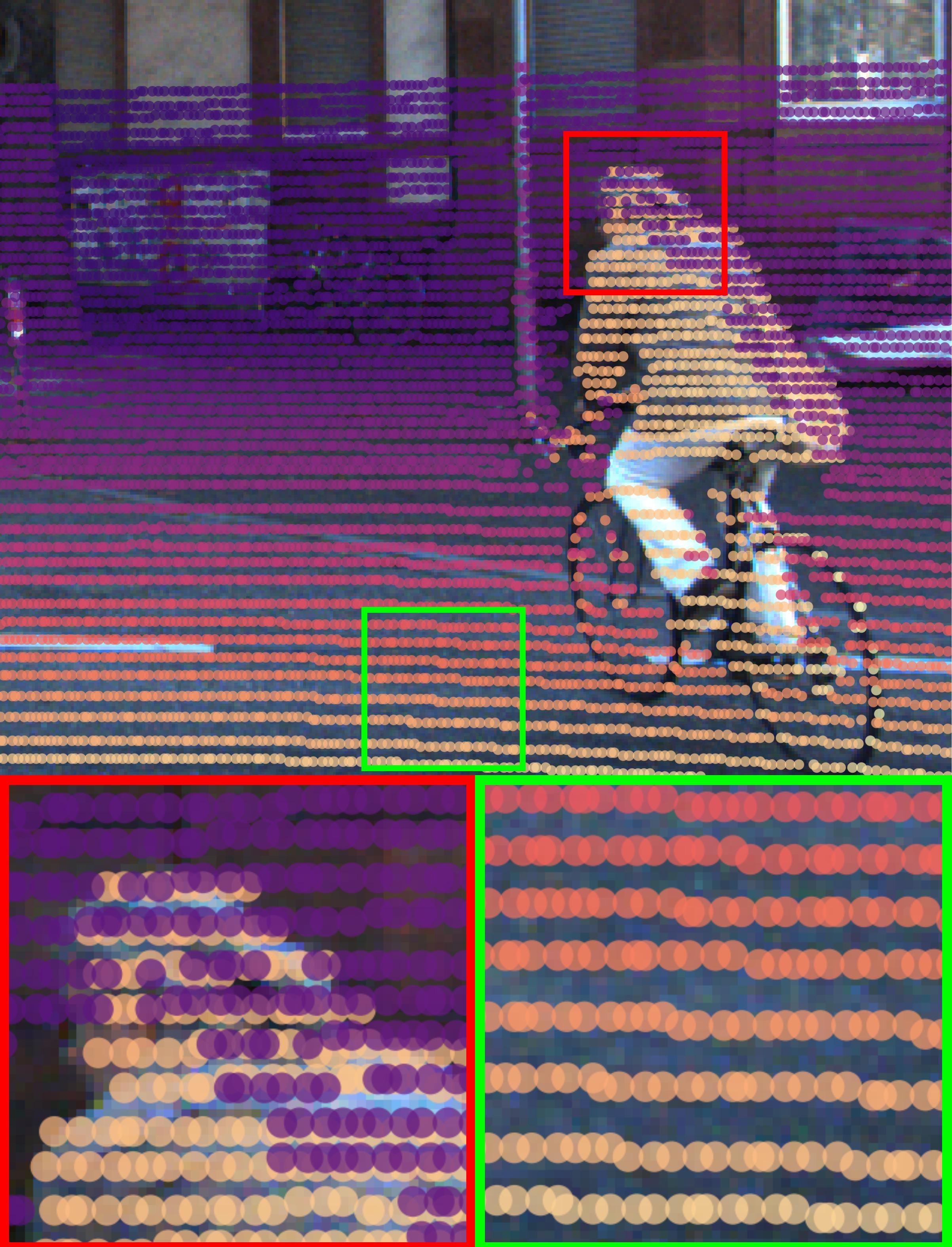}
        \caption{Raw}
    \end{subfigure}
    \begin{subfigure}[]{0.32\linewidth}
        \centering
        \includegraphics[width=\linewidth]{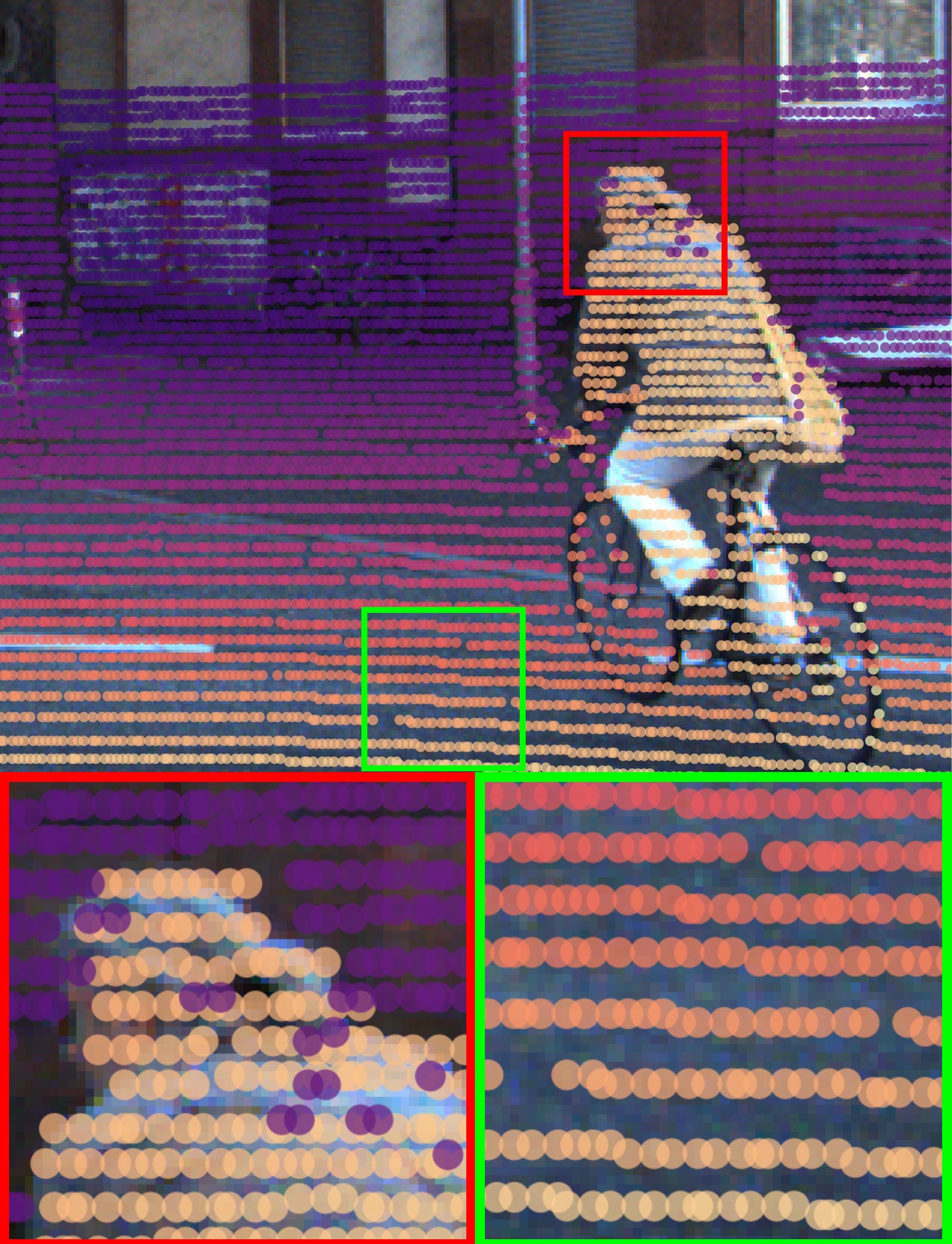}
        \caption{Half-Occlusion-clean}
    \end{subfigure}
    \begin{subfigure}[]{0.32\linewidth}
        \centering
        \includegraphics[width=\linewidth]{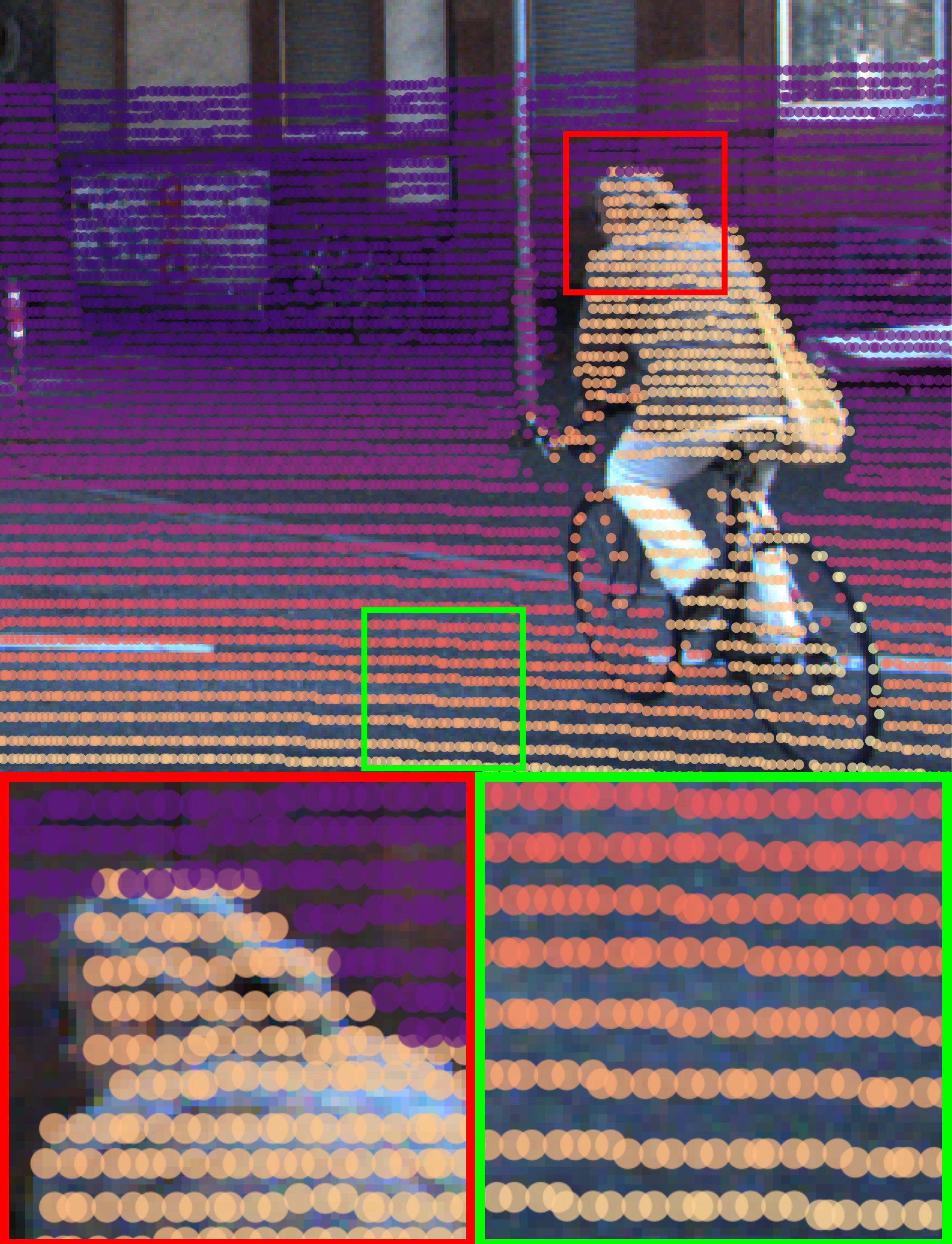}
        \caption{\methodName-clean}
    \end{subfigure}
    \vspace{-10pt}
    \caption{\textbf{Visualization of \depthmaps}. We observe that the half-occlusion algorithm removes some but not all of the projective artifacts from the \depthmaps. Moreover, we also observe that the half-occlusion method removes non-artifact points from the \depthmap which is highly undesirable. \methodName only removes artifacts while preserving non-artifact points.}
    \label{fig:half-occ}
\end{figure}

\cref{fig:half-occ} provides qualitative results to support the quantitative results in \cref{tab:kitti_depth_quality}. As discussed in \cref{sec:qual_assess}, we observe in \cref{fig:half-occ} that the modified half-occlusion only removes some but not all of artifacts. We also observe that in addition to removing some artifacts the modified half-occlusion method also removes some non-artifacts(\thatIs good points) which is highly undesirable.



\subsection{\monoDE}
\begin{figure*}[!t]
    \centering
    \begin{subfigure}[!h]{0.5\textwidth}
        \centering
        \includegraphics[width=\linewidth]{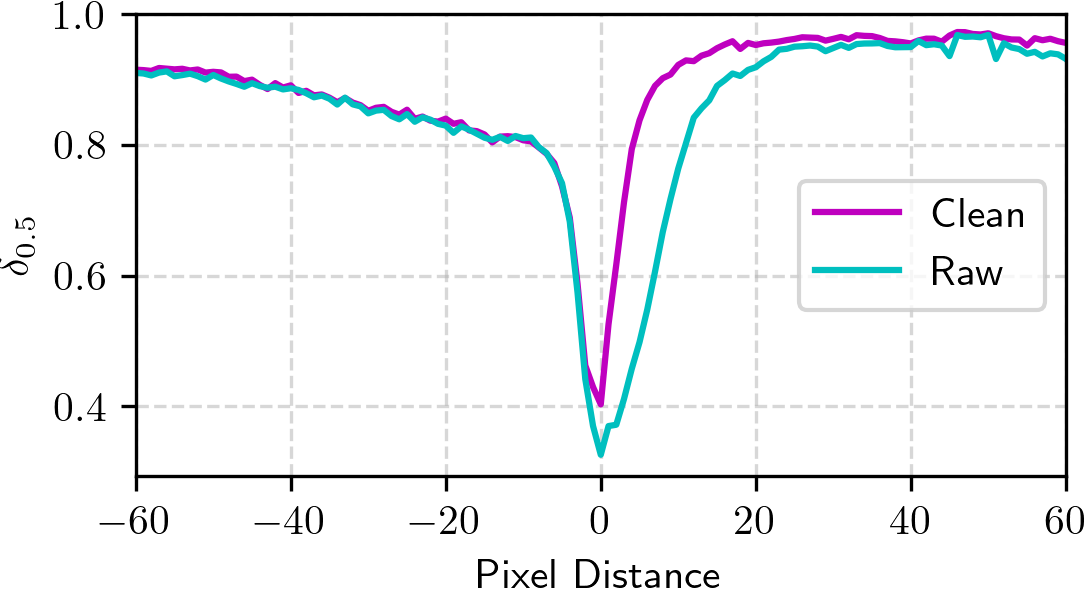}
        \caption{}
    \end{subfigure}%
    \hfill
    \begin{subfigure}[!h]{0.5\textwidth}
        \centering
        \includegraphics[width=\linewidth]{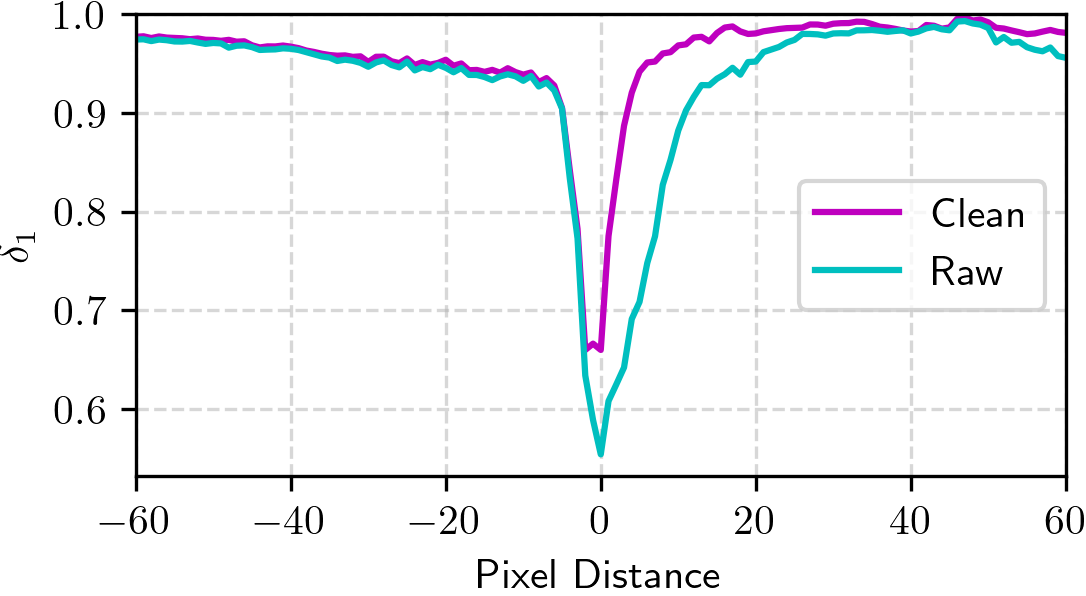}
        \caption{}
    \end{subfigure}
    \begin{subfigure}[!h]{0.5\textwidth}
        \centering
        \includegraphics[width=\linewidth]{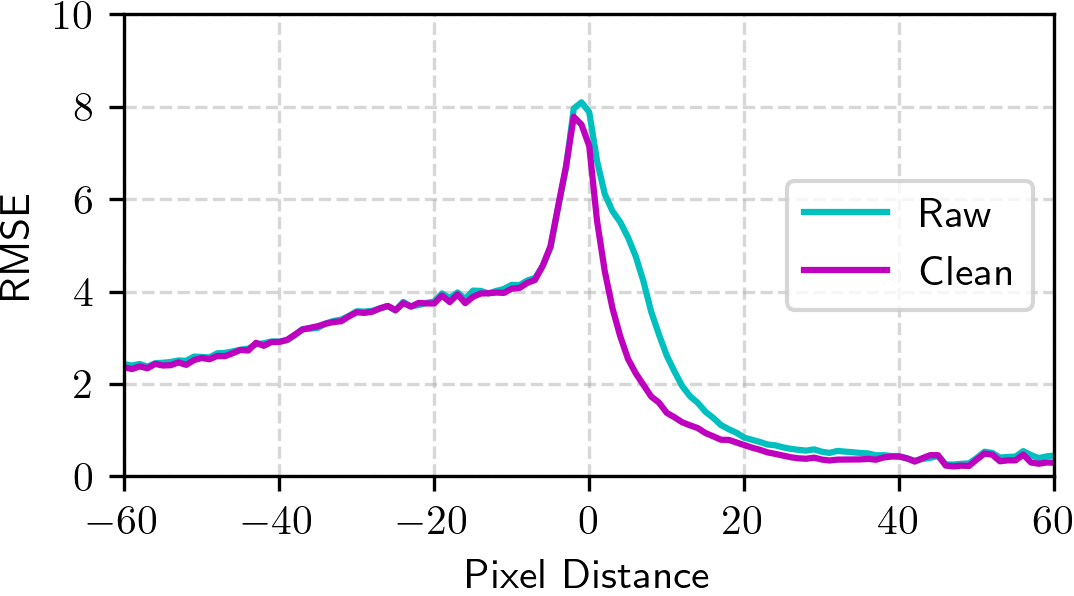}
        \caption{}
    \end{subfigure}%
    \hfill
    \begin{subfigure}[!h]{0.5\textwidth}
        \centering
        \includegraphics[width=\linewidth]{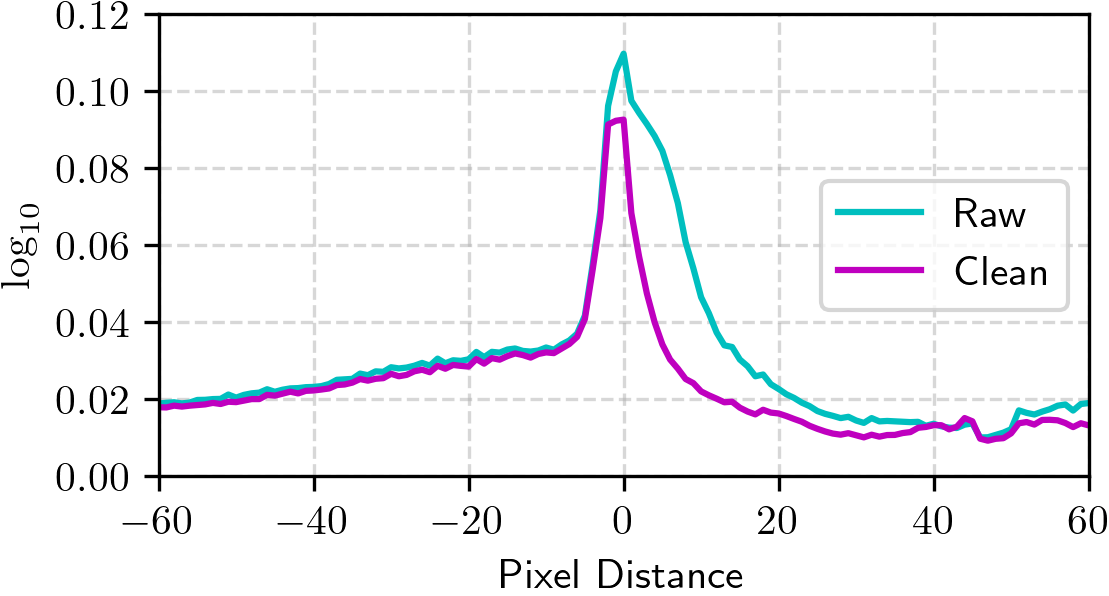}
        \caption{}
    \end{subfigure}
    \caption{\textbf{\monoDE Boundary Performance Analysis} We show the performance of the \zoeDepth trained on raw and \methodName \depthmaps evaluated on \textbf{(a)} $\delta_{0.5}$, \textbf{(b)} $\delta_1$, \textbf{(c)} \rmse and \textbf{(d)} \logTen as a function of signed pixel distance from the foreground/background boundary. [Key: Pixel Distance $\rightarrow$ Foreground ($+x$), Background ($-x$)]}
    \label{fig:monoDE_fore_analysis}
\end{figure*}

\cref{tab:kitti_zoe_stereo_eval} shows that the performance gain in the foreground is greater than that of the background. However, we cannot determine from \cref{tab:kitti_zoe_stereo_eval} whether the performance gain is due to the improvement in the entire foreground region or just specific parts of the foreground.
Therefore, to get a spatial performance analysis of \zoeDepth\cite{bhat2023zoedepth}, we evaluate the performance of the predicted \depthmaps as a function of distance from the foreground/background boundary as shown in \cref{fig:monoDE_fore_analysis}.

\begin{figure*}[!t]
    \begin{subfigure}[]{0.33\textwidth}
        \includegraphics[width=\linewidth]{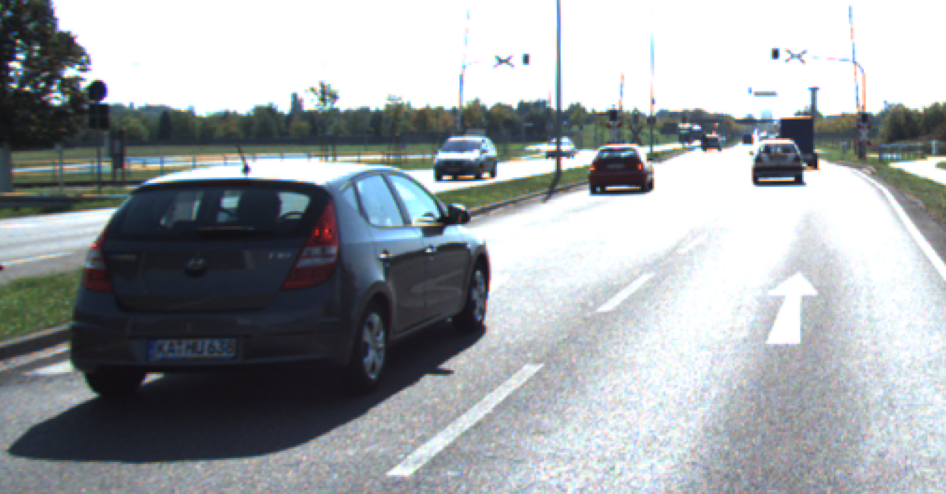}
        \caption{}
    \end{subfigure}%
    \begin{subfigure}[]{0.33\textwidth}
        \includegraphics[width=\linewidth]{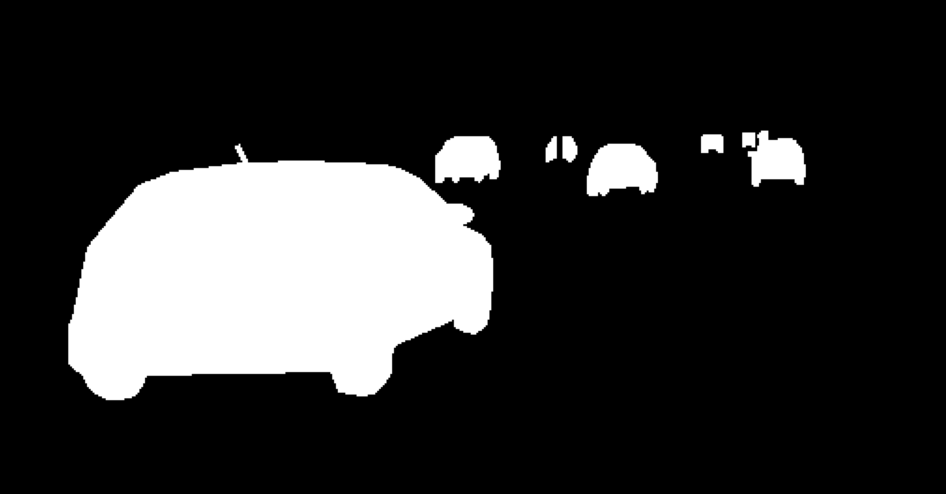}
        \caption{}
    \end{subfigure}%
    \begin{subfigure}[]{0.33\textwidth}
        \includegraphics[width=\linewidth]{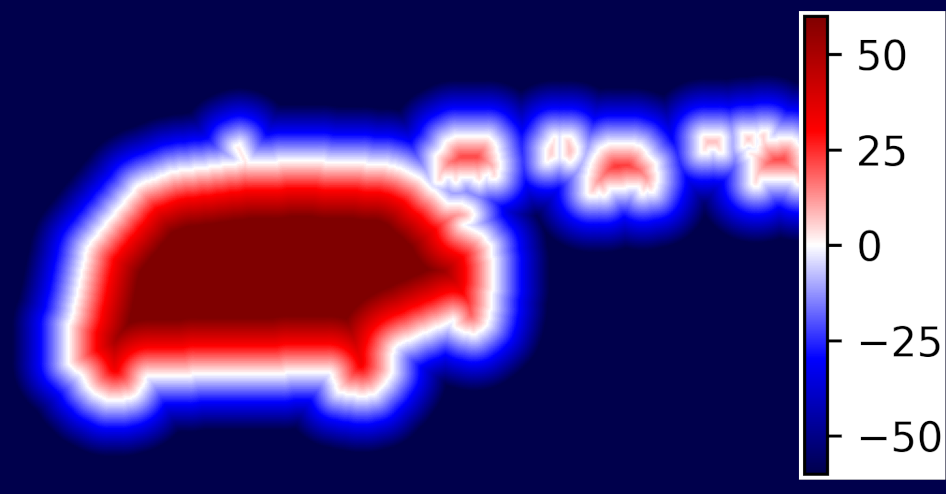}
        \caption{}
    \end{subfigure}
    \caption{\textbf{Visualization of Signed Distance:} We show the \textbf{(b)} signed distance of \textbf{(a)} a \kitti Stereo sample calculated using \textbf{(c)} provided segmentation map.}
    \label{fig:sdf}
\end{figure*}

We perform the analysis on the \kitti Stereo split. We first use the segmentation map provided in the \kitti Stereo split to determine the signed distance function ($SDF(\cdot)$) from the foreground/background boundary as shown in \cref{fig:sdf}. Given a pixel $\pixel$ the signed distance function outputs a scalar $x \in \realDomain$ that implicitly determines foreground/background as follows:
\begin{align}
    SDF(\pixel) = \begin{cases}
        x > 0 & \pixel \in \text{Foreground} \\
        x < 0 & \pixel \in \text{Background} \\
        0 & \pixel \in \text{Boundary} \\
    \end{cases}
\end{align}

We plot the predicted depth performance against the signed distance from the boundary.
We evaluate the performance using  $\delta_{0.5}$, $\delta_1$, \rmse and \logTen. 
We observe from \cref{fig:monoDE_fore_analysis} that
the performance gain gradually increases from around -3-pixel distance, peaks at a 5-pixel distance, and dies down at a 25-pixel distance from the boundary. 
Put differently, although there is a noticeable improvement in performance near the boundary, it's worth mentioning that the performance boost is more significant in the foreground area ($+x$) that is closer to the boundary.
We note that the performance gain is not just limited to the region closer to the boundary.
The results from the analysis agrees with \cref{tab:kitti_zoe_stereo_eval} results.

\begin{figure}[!t]
    \vspace{-0.0cm}
    \centering
    \begin{subfigure}{0.5\linewidth}
        \includegraphics[width=\linewidth]{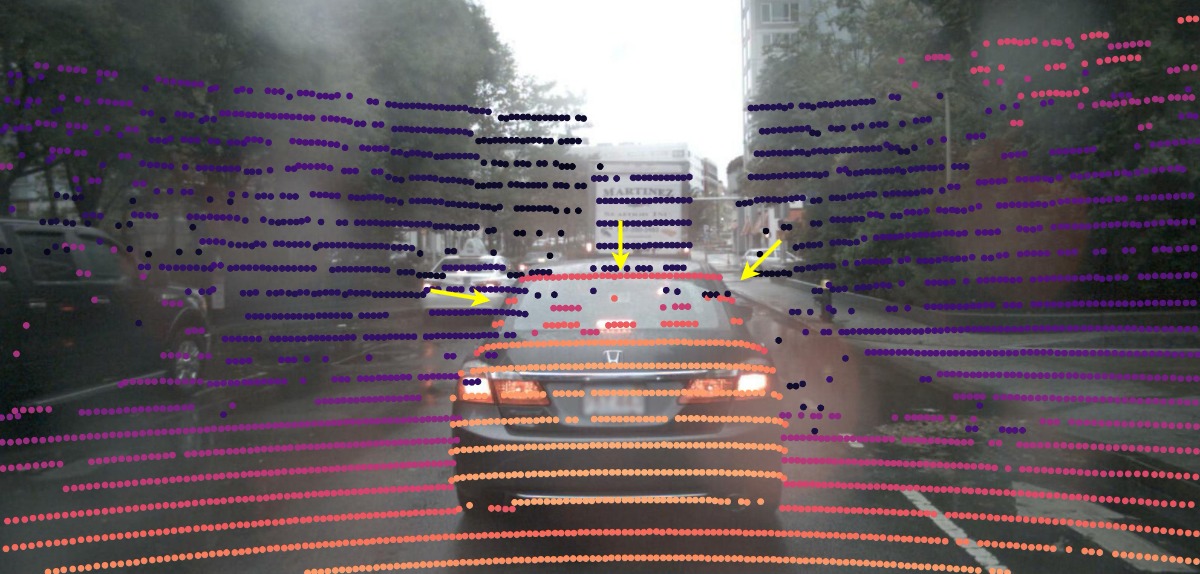}
    \end{subfigure}%
    \begin{subfigure}{0.5\linewidth}
        \includegraphics[width=\linewidth]{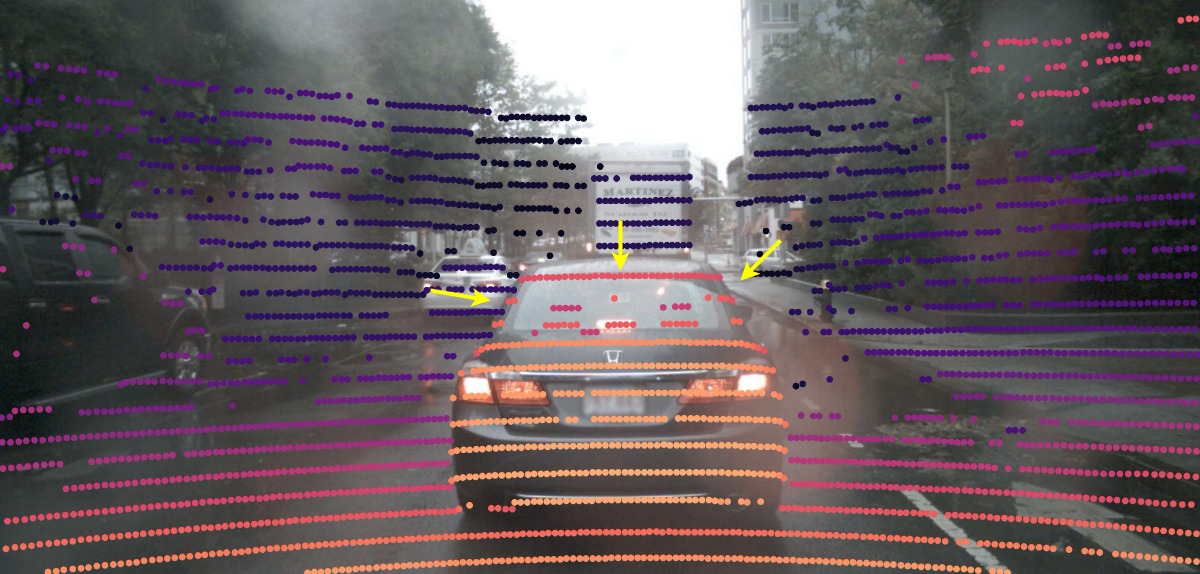}
    \end{subfigure}
    \begin{subfigure}{0.5\linewidth}
        \includegraphics[width=\linewidth]{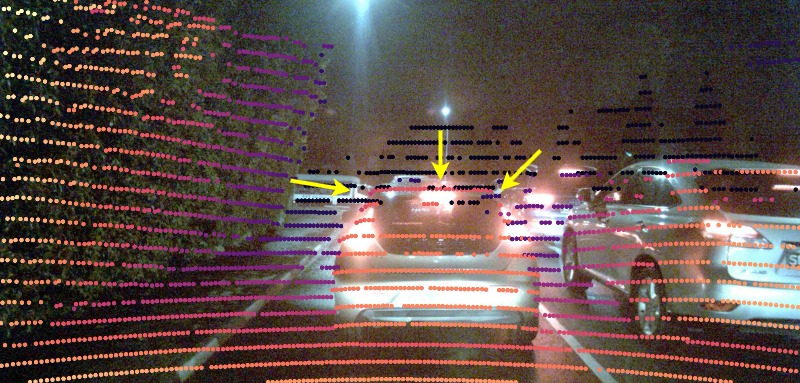}
    \end{subfigure}%
    \begin{subfigure}{0.5\linewidth}
        \includegraphics[width=\linewidth]{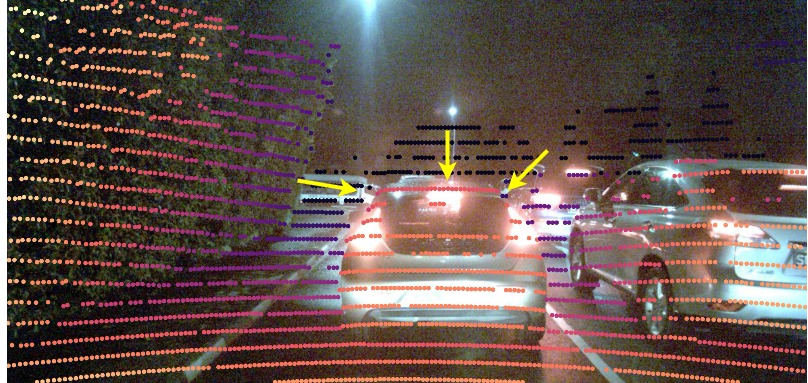}
    \end{subfigure}
    \caption{\textbf{Qualitative comparison} between raw (left) and \methodName (right) \depthmaps in rain (top) and night (bottom). 
    }
    \label{fig:qualitive_weather_comparison}
    \vspace{-0.4cm}
\end{figure}


\clearpage
\subsection{\monoThreeD}
\begin{wrapfigure}{R}{0.55\linewidth}
    \centering
    \includegraphics[width=\linewidth]{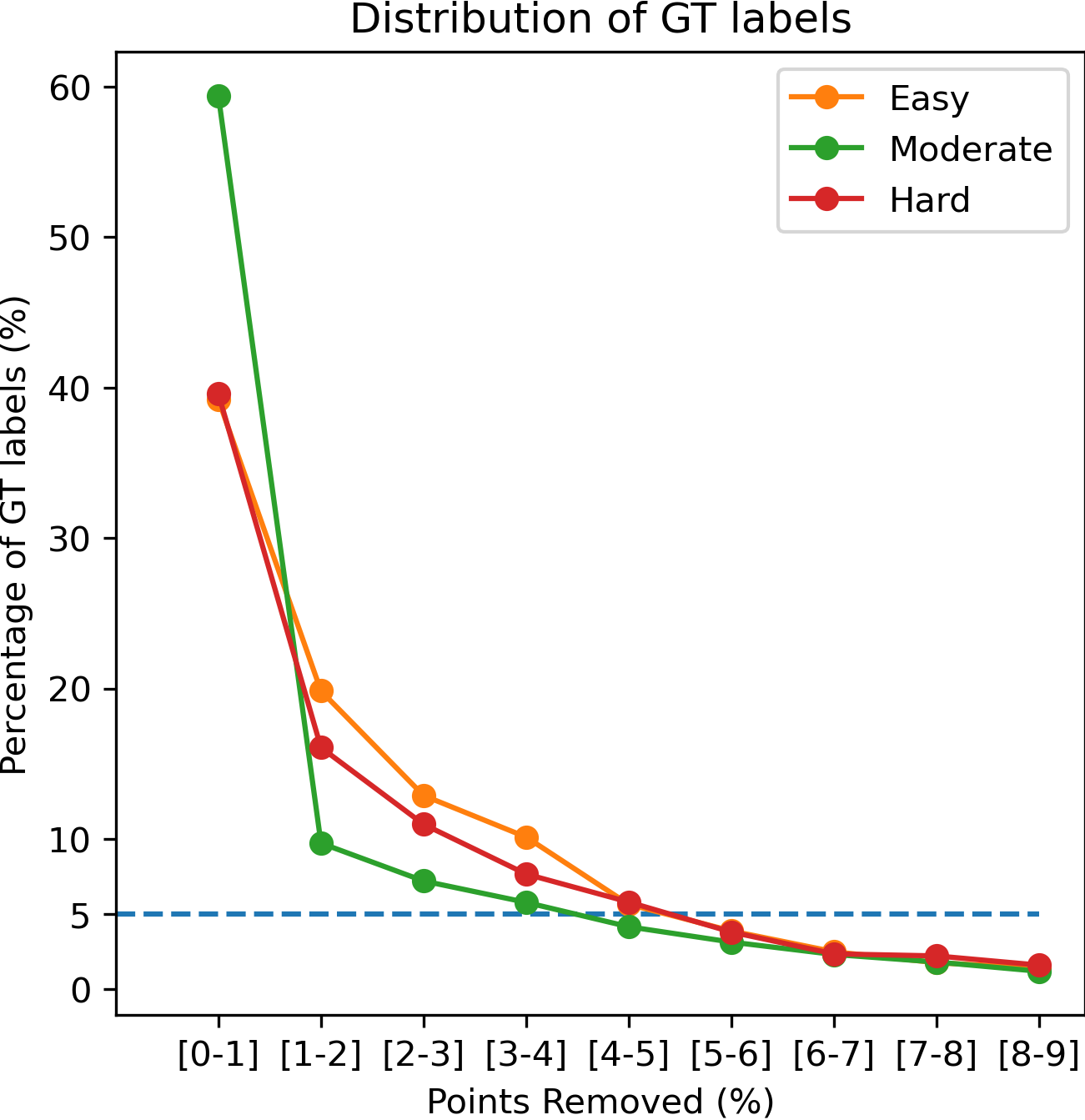}
    \caption{\textbf{Distribution of GT labels:} Left axis: Percentage of total GT labels present in a bin; Right axis: Percentage of points removed grouped into bins of width $1\%$}
    \label{fig:label_dist}
\end{wrapfigure}
\cref{tab:kitti_val_detection} shows performance improvement on the Easy category, but negligible improvement in the Moderate and Hard category.
The performance gain in the Easy category is due to the artifacts removed from the pre-computed depth maps of \cite{huang2022monodtr}.
Hence, we investigate whether the artifacts present within the \twoD bounding boxes of GT labels vary for the object categories of Easy, Moderate, and Hard. Furthermore, we examine the predicted labels to determine whether the correlation between artifacts and performance is strong or weak.
Therefore we perform analysis on the percentage of artifacts detected and removed for each GT label in \kitti Detection Val split and their effect on the performance of \cite{huang2022monodtr} to support the results reported in \cref{tab:kitti_val_detection}. 

As mentioned in \cref{sec:mono_obj_det}, \cite{huang2022monodtr} pre-computes $4\times$ downsampled raw projected \depthmaps to be used as depth supervision. 
For each of the GT labels, we estimate the percentage of points removed between raw and clean pre-computed  \depthmaps within GT \twoD bounding boxes.
We categorize the removal percentages into 1\% width bins and plot the distribution of GT labels in each bin in \cref{fig:label_dist}.

\begin{figure*}[!t]
    \centering
    \begin{subfigure}[!h]{0.32\textwidth}
        \centering
        \includegraphics[width=\linewidth]{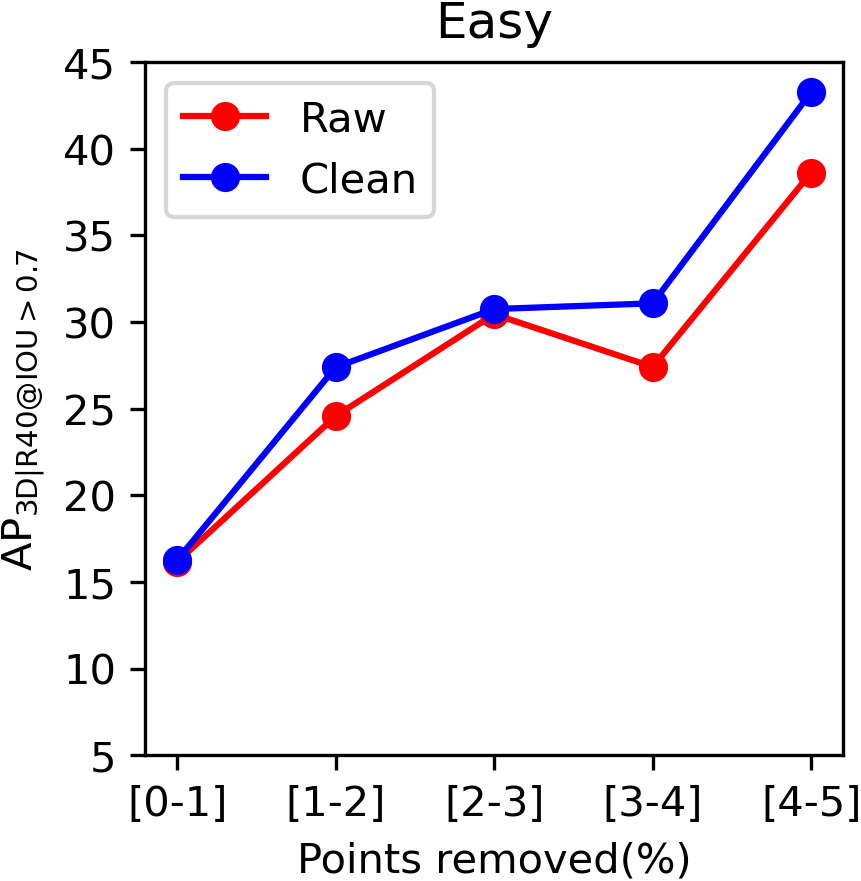}
    \end{subfigure}
    \begin{subfigure}[!h]{0.32\textwidth}
        \centering
        \includegraphics[width=\linewidth]{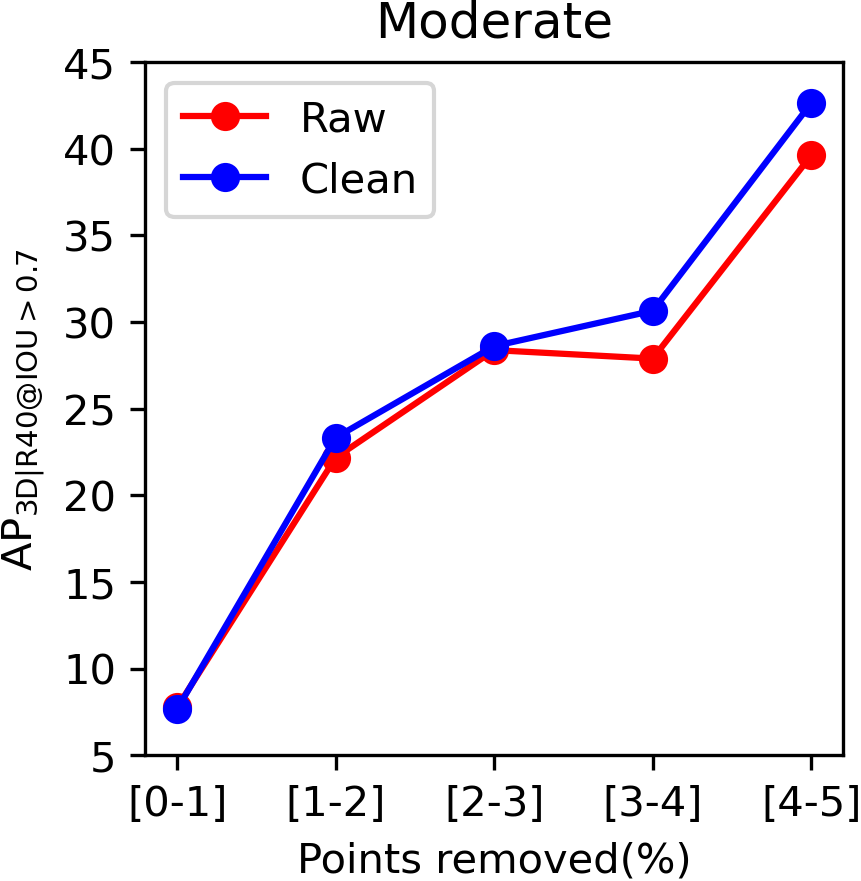}
    \end{subfigure}
    \begin{subfigure}[!h]{0.32\textwidth}
        \centering
        \includegraphics[width=\linewidth]{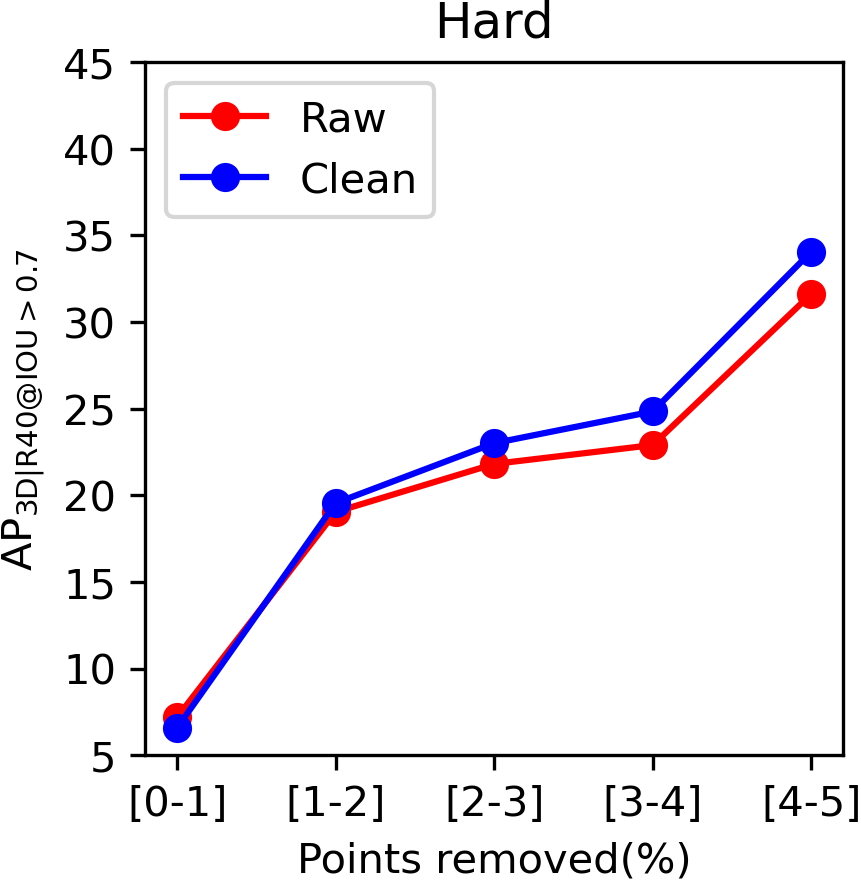}
    \end{subfigure}
    \caption{\textbf{\monoThreeD Analysis.} We demonstrate correlation between the \% of points removed and performance of \monodtr\cite{huang2022monodtr}. We observe that the increase in \apThreeDForty is inversely proportional to the \% of artifacts detected and removed showing that \methodName is beneficial to the performance of \monoThreeD.}
    \label{fig:mono_obj_analysis}
\end{figure*}

We observe that $[40-60]\%$ of the total number of GT labels have $<1\%$ artifacts detected and removed.
Furthermore, we observe that $\sim 95\%$ of the entire GT labels have $[0-5]\%$ of artifacts detected and removed.
Thus we limit the analysis to the range of $[0-5]\%$ points removed.
We evaluate the predicted bounding boxes in each of these percentage bins as shown in \cref{fig:mono_obj_analysis}.
We observe that as the percentage of points removed increases, the \apThreeDForty also increases in the Easy, Moderate, and Hard categories.
However, from \cref{fig:label_dist} and \cref{fig:mono_obj_analysis} we find that the performance increase is offset by the decrease in the ratio of GT labels as the percentage of artifacts detected and removed increases.
The results reported in \cref{tab:kitti_val_detection} can be naively interpreted as the mean of \apThreeDForty in \cref{fig:mono_obj_analysis} weighted by the ratio of GT labels in \cref{fig:label_dist}.
We note from \cref{fig:mono_obj_analysis} that the improvement margin in the Easy category is greater than that of Moderate and Hard. Hence, the performance of the Easy category is less affected due to the decrease in GT labels as compared to the Moderate/Hard category.
Therefore, the results reported in \cref{tab:kitti_val_detection} is supported by \cref{fig:label_dist} and \cref{fig:mono_obj_analysis}.

\begin{table*}[!t]
    \caption{\textbf{\monoDE Foreground-Background Analysis.} We train \idisc~\cite{bhat2023zoedepth}(Swin-L) on \kitti Eigen with raw, clean and semi-dense \depthmaps, and test on \kitti Stereo. 
    \methodName mostly benefits foreground pixels over raw \depthmaps.
    [Key: \firstKey{Best}, \secondKey{Second Best}].
    }
    \label{tab:kitti_iDisc_stereo_eval}
    \centering
    \scalebox{\scaleFraction}{
    \setlength\tabcolsep{0.1cm}
    \begin{tabular}{c|c|c c c|c c c c}
        Eval Region & Depthmap & \textbf{$\delta_{0.5}$}$\uparrow$ & \textbf{$\delta_1$}$\uparrow$ & \textbf{$\delta_2$}$\uparrow$ & \rmse$\downarrow$ & \rmseLog$\downarrow$ & \absRel$\downarrow$ & \sqRel$\downarrow$\\
        \hline
        \multirow{3}*{All} & Raw & 0.911 & 0.958 & 0.980 & 2.698 & 0.121 & 0.068 & 0.556\\
         & Semi-Dense & \first{0.935} & \first{0.978} & \first{0.993} & \first{2.160} & \first{0.083} & \first{0.043} & \first{0.211}\\ 
         & \methodName & \second{0.930} & \second{0.975} & \second{0.991} & \second{2.247} & \second{0.089} & \second{0.049} & \second{0.280}\\\hline
        \multirow{3}*{Foreground} & Raw & 0.805 & 0.865 & 0.916 & 4.527  & 0.206 & 0.182 & 2.474\\
         & Semi-Dense & \first{0.910} & \first{0.963} & \first{0.983} & \first{2.370} & \first{0.105} & \first{0.070} & \first{0.460}\\
         & \methodName & \second{0.879} & \second{0.938} & \second{0.967} & \second{3.141} & \second{0.136} & \second{0.102} & \second{1.061}\\\hline
        \multirow{3}*{Background} & Raw & 0.932 & 0.977 & 0.994 & 2.015 & 0.080 & 0.045 & 0.174\\
         & Semi-Dense & \first{0.940} & \second{0.981} & \second{0.995} & \second{2.093} & \second{0.073} & \second{0.038} & \second{0.161}\\ 
         & \methodName & \second{0.939} & \first{0.982} & \first{0.996} & \first{1.977} & \first{0.069} & \first{0.039} & \first{0.133}\\
    \end{tabular}
    }
\end{table*}

\clearpage 
\section{Additional Results}

\begin{figure*}[!t]
    \begin{subfigure}[]{\textwidth}
        \centering
        \includegraphics[width=0.33\linewidth]{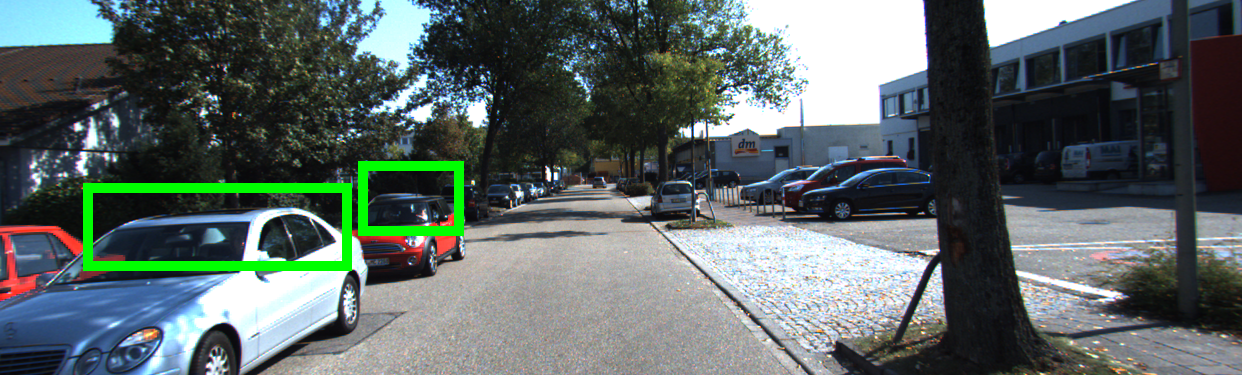}%
        \includegraphics[width=0.33\linewidth]{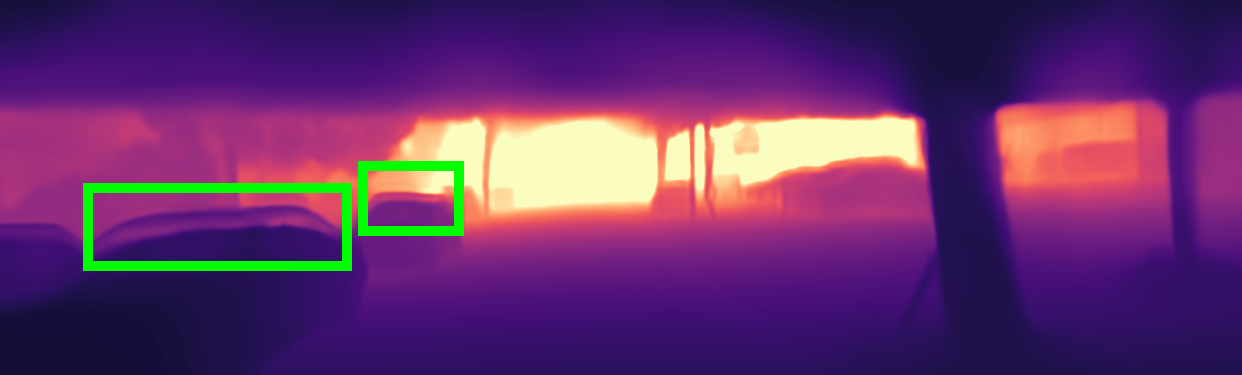}%
        \includegraphics[width=0.33\linewidth]{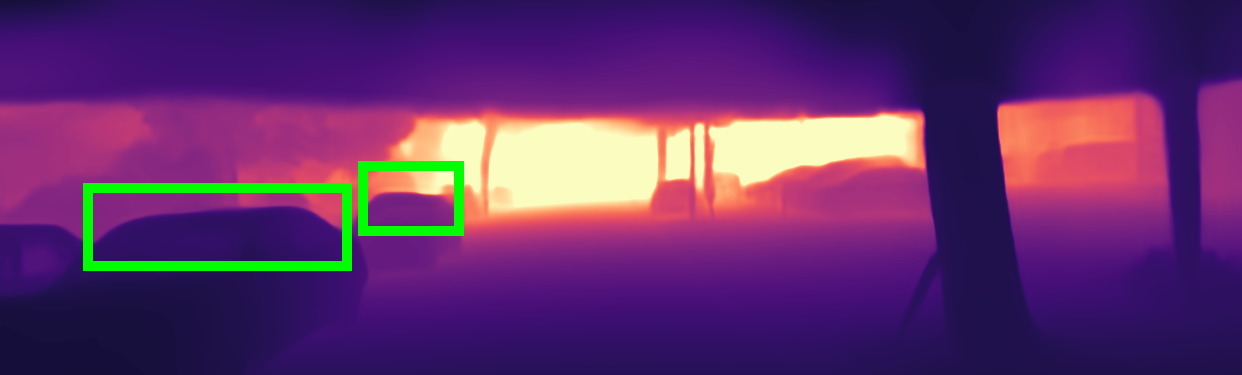}
    \end{subfigure}
    \begin{subfigure}[]{\textwidth}
        \centering
        \includegraphics[width=0.33\linewidth]{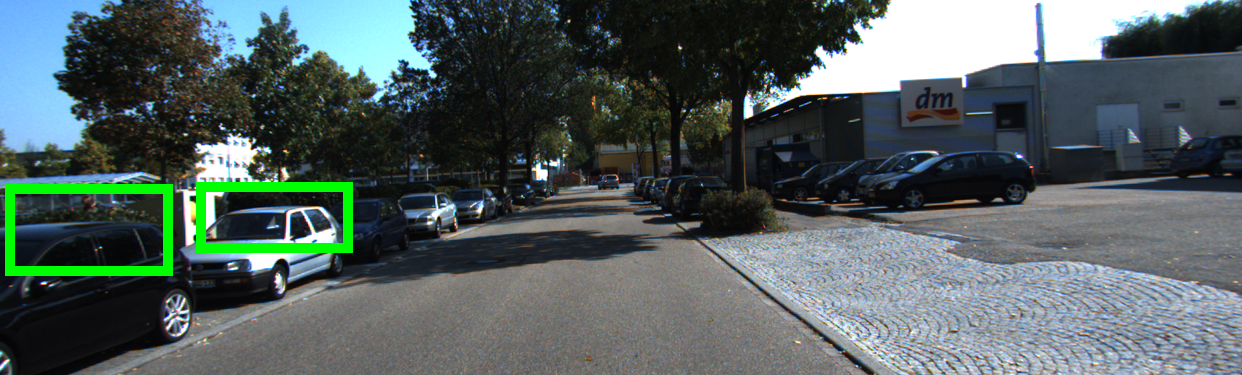}%
        \includegraphics[width=0.33\linewidth]{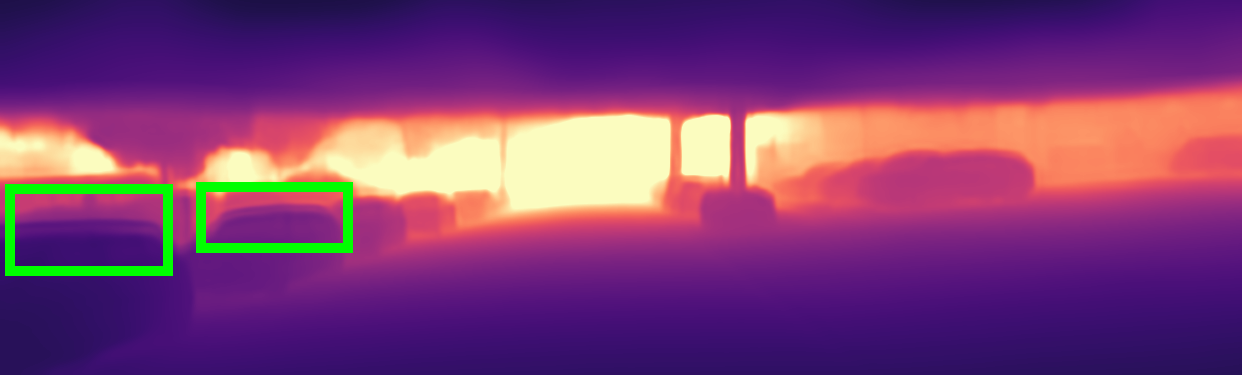}%
        \includegraphics[width=0.33\linewidth]{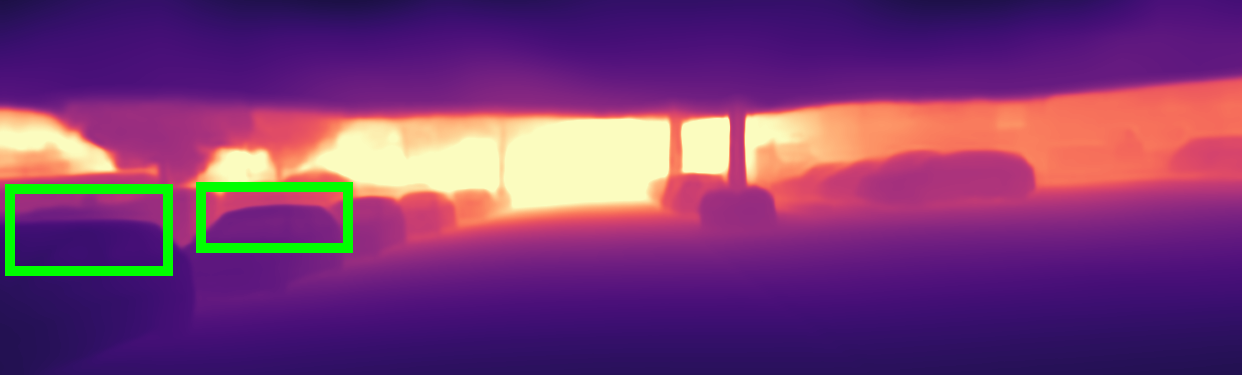}
    \end{subfigure}
    \begin{subfigure}[]{\textwidth}
        \centering
        \includegraphics[width=0.33\linewidth]{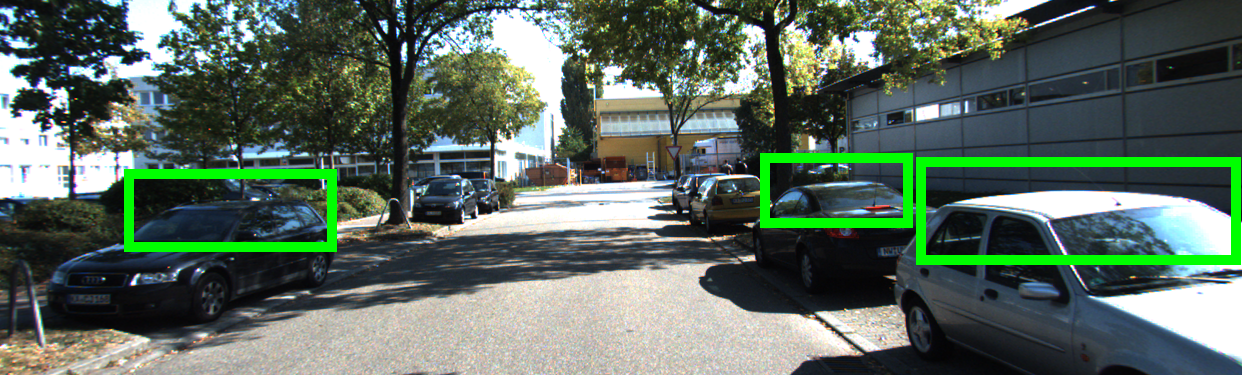}%
        \includegraphics[width=0.33\linewidth]{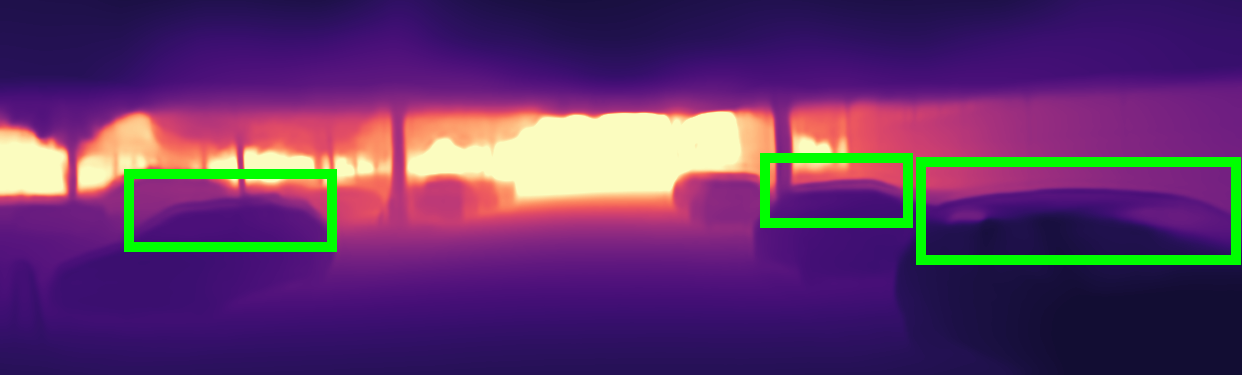}%
        \includegraphics[width=0.33\linewidth]{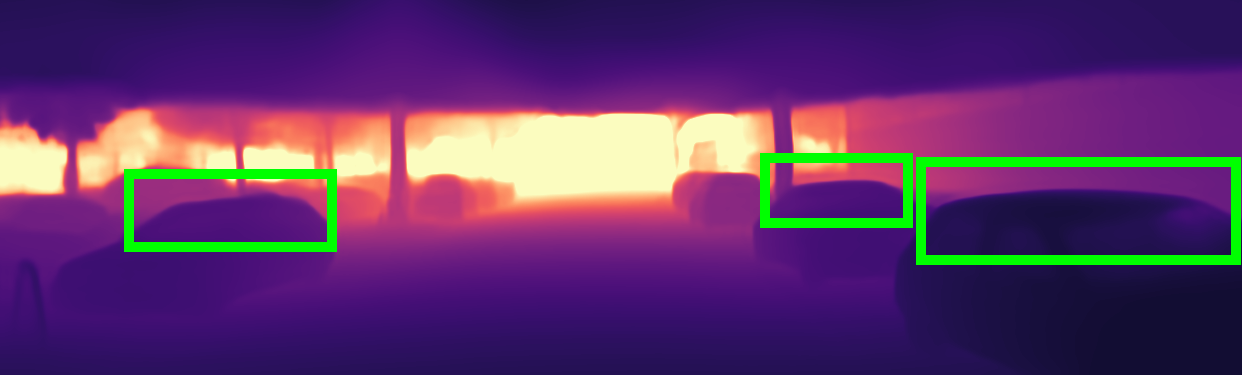}
    \end{subfigure}
    \begin{subfigure}[]{\textwidth}
        \centering
        \includegraphics[width=0.33\linewidth]{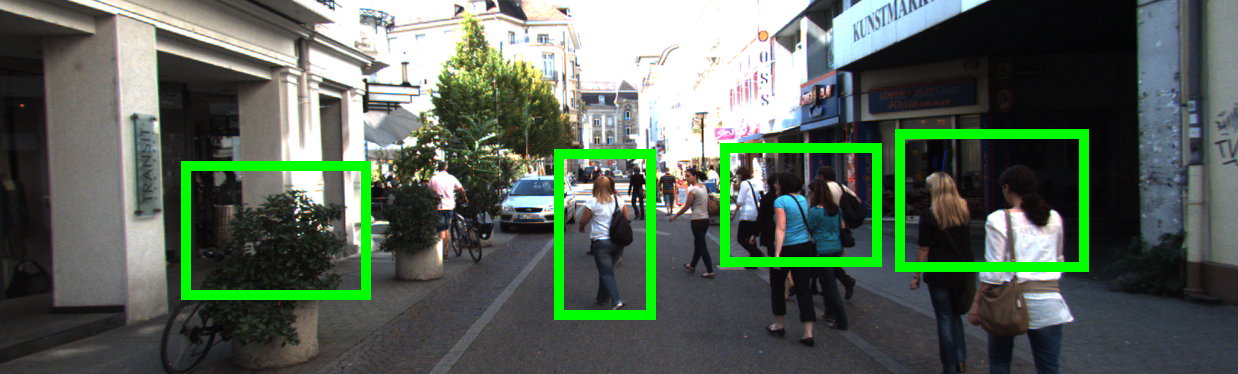}%
        \includegraphics[width=0.33\linewidth]{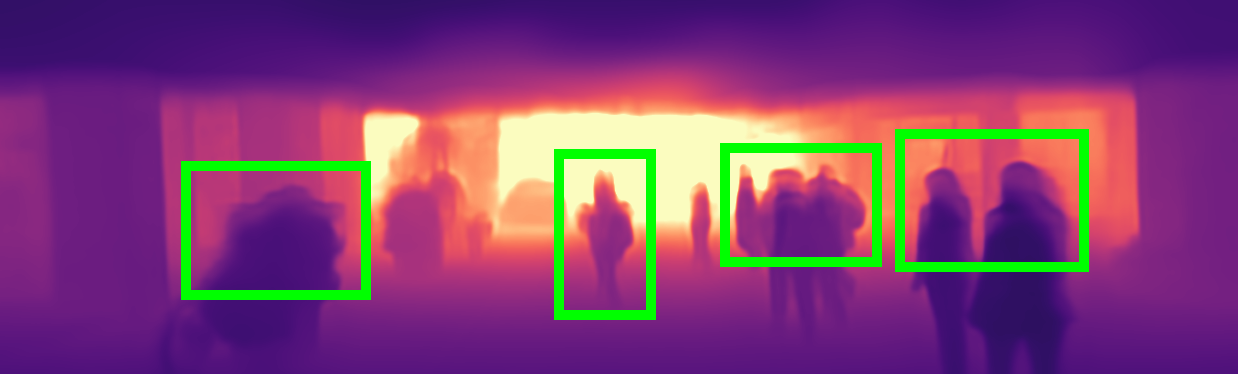}%
        \includegraphics[width=0.33\linewidth]{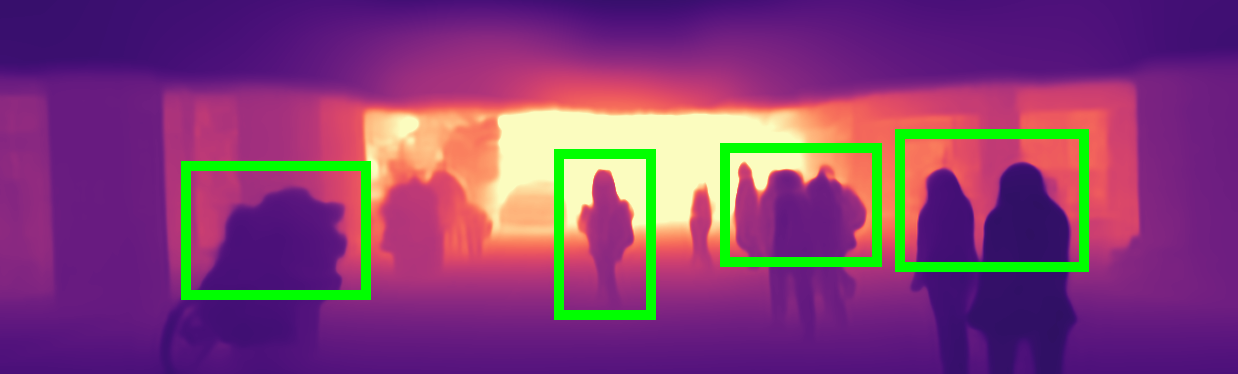}
    \end{subfigure}
    \begin{subfigure}[]{\textwidth}
        \centering
        \includegraphics[width=0.33\linewidth]{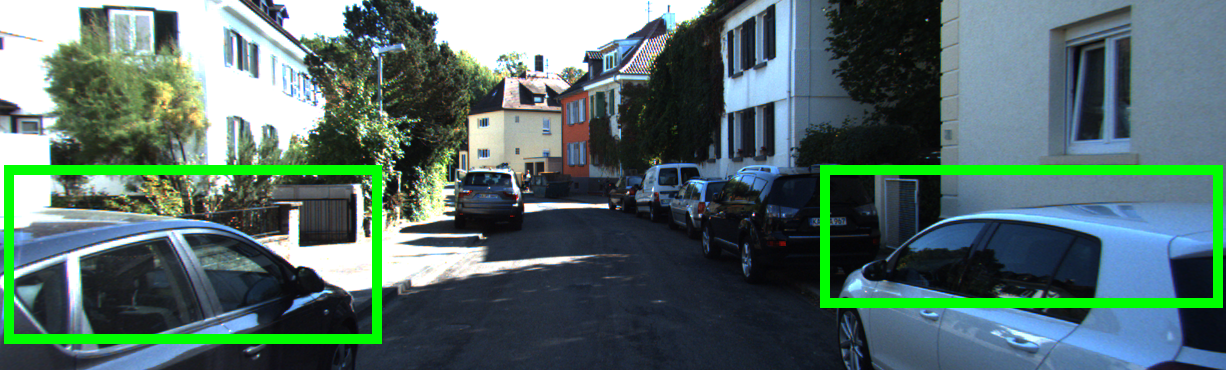}%
        \includegraphics[width=0.33\linewidth]{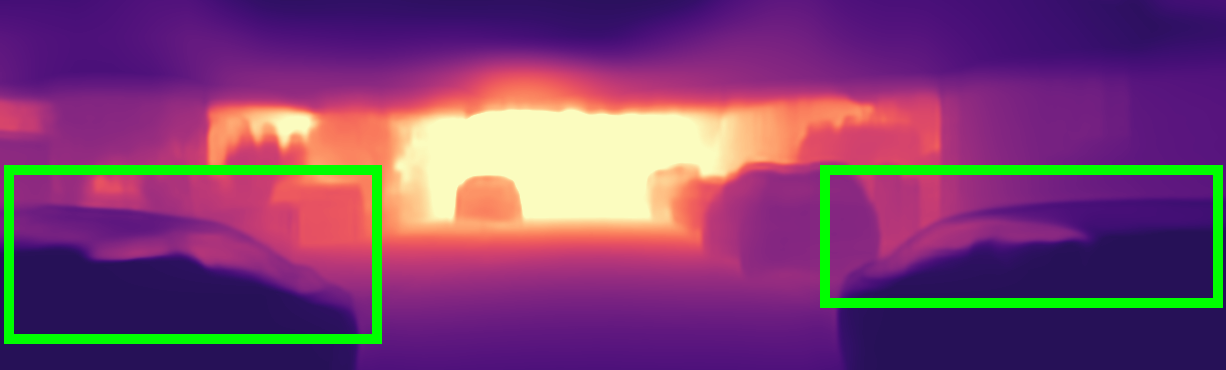}%
        \includegraphics[width=0.33\linewidth]{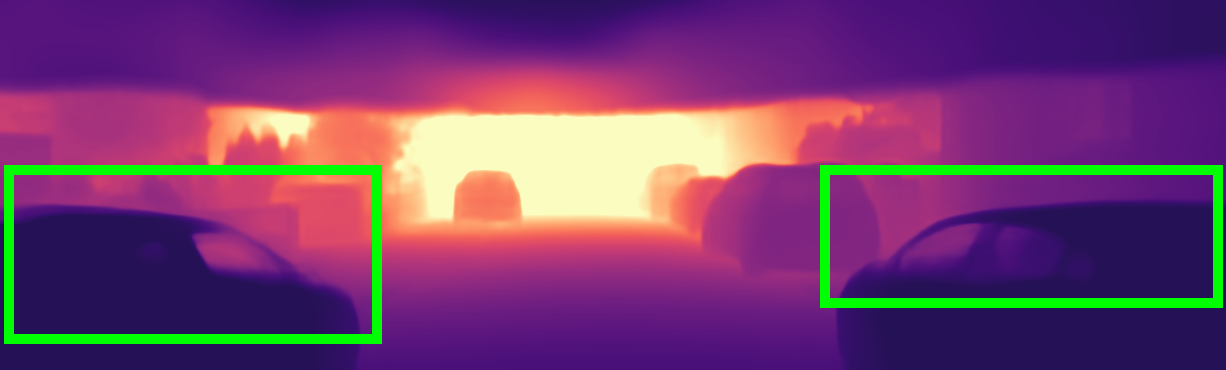}
    \end{subfigure}
    \caption{\textbf{Visualization of MonoDE on \kitti} [Key: 1\textsuperscript{st} column - RGB, 2\textsuperscript{nd} column - \zoeDepth model trained on raw \depthmaps, 3\textsuperscript{nd} column - \zoeDepth model trained on \methodName \depthmaps]}
    \label{fig:monoDE_kitti_vis}
\end{figure*}

\begin{figure}[!t]
    \begin{subfigure}[]{\textwidth}
        \centering
        \includegraphics[width=0.33\linewidth]{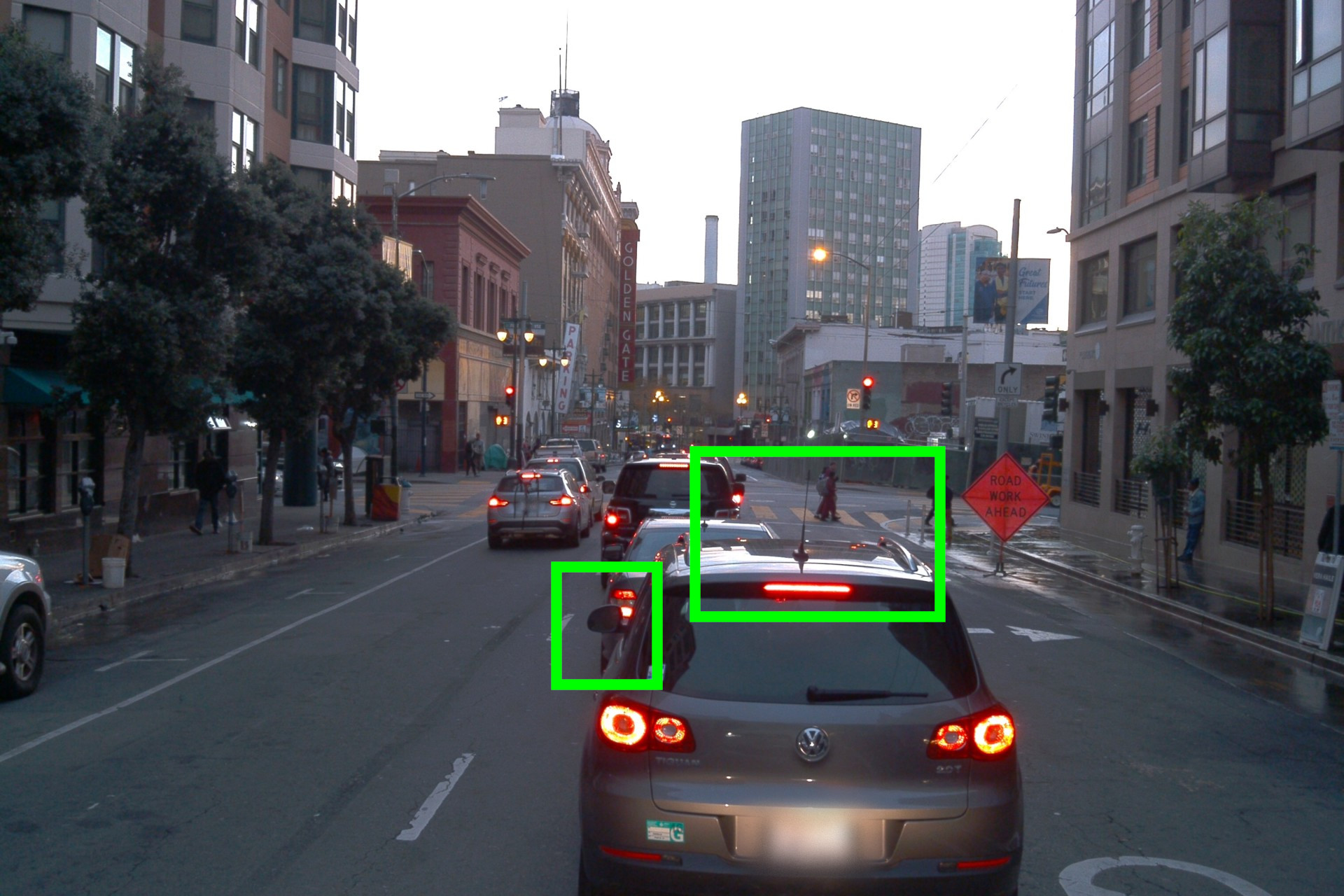}%
        \includegraphics[width=0.33\linewidth]{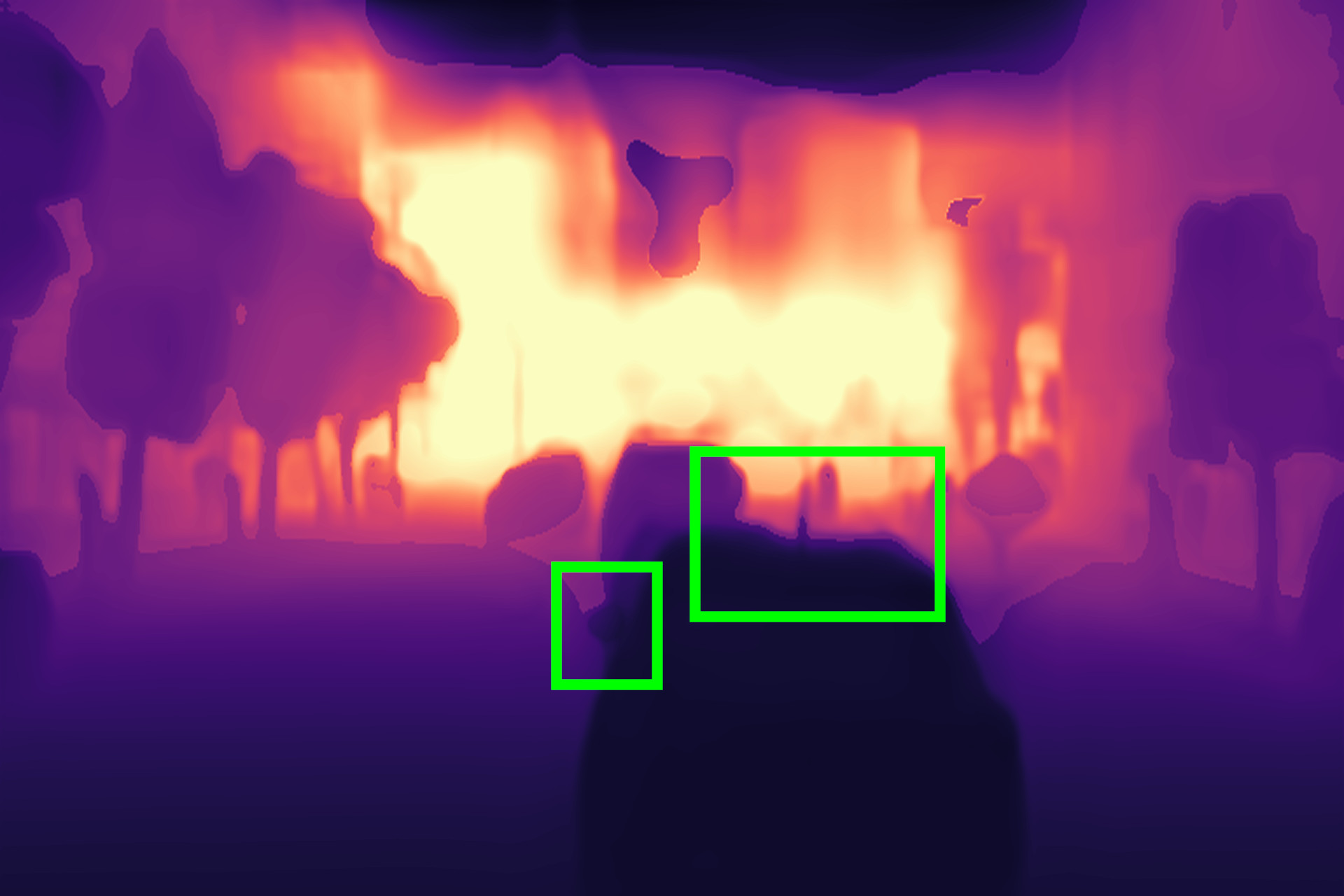}%
        \includegraphics[width=0.33\linewidth]{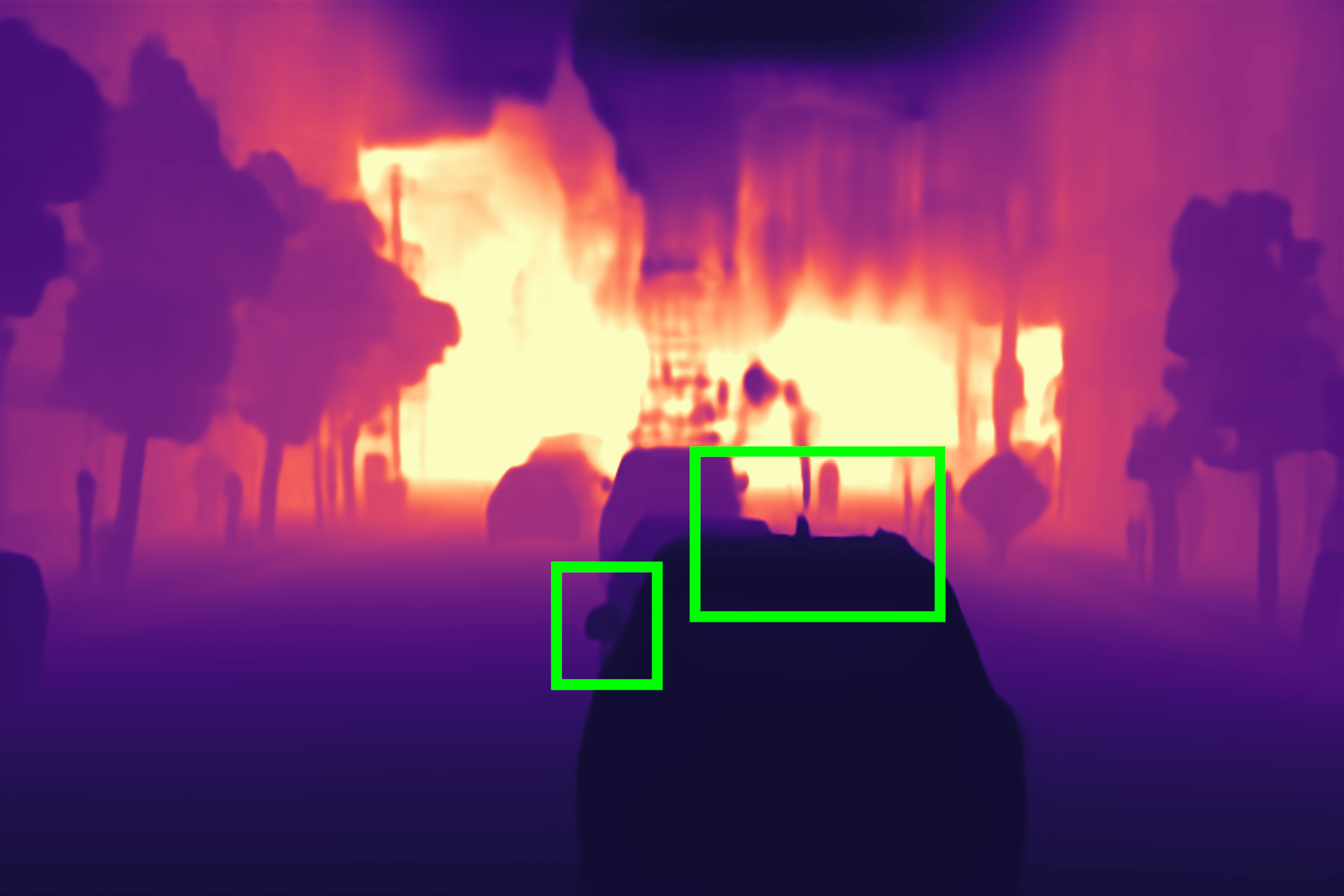}
    \end{subfigure}
    \begin{subfigure}[]{\textwidth}
        \centering
        \includegraphics[width=0.33\linewidth]{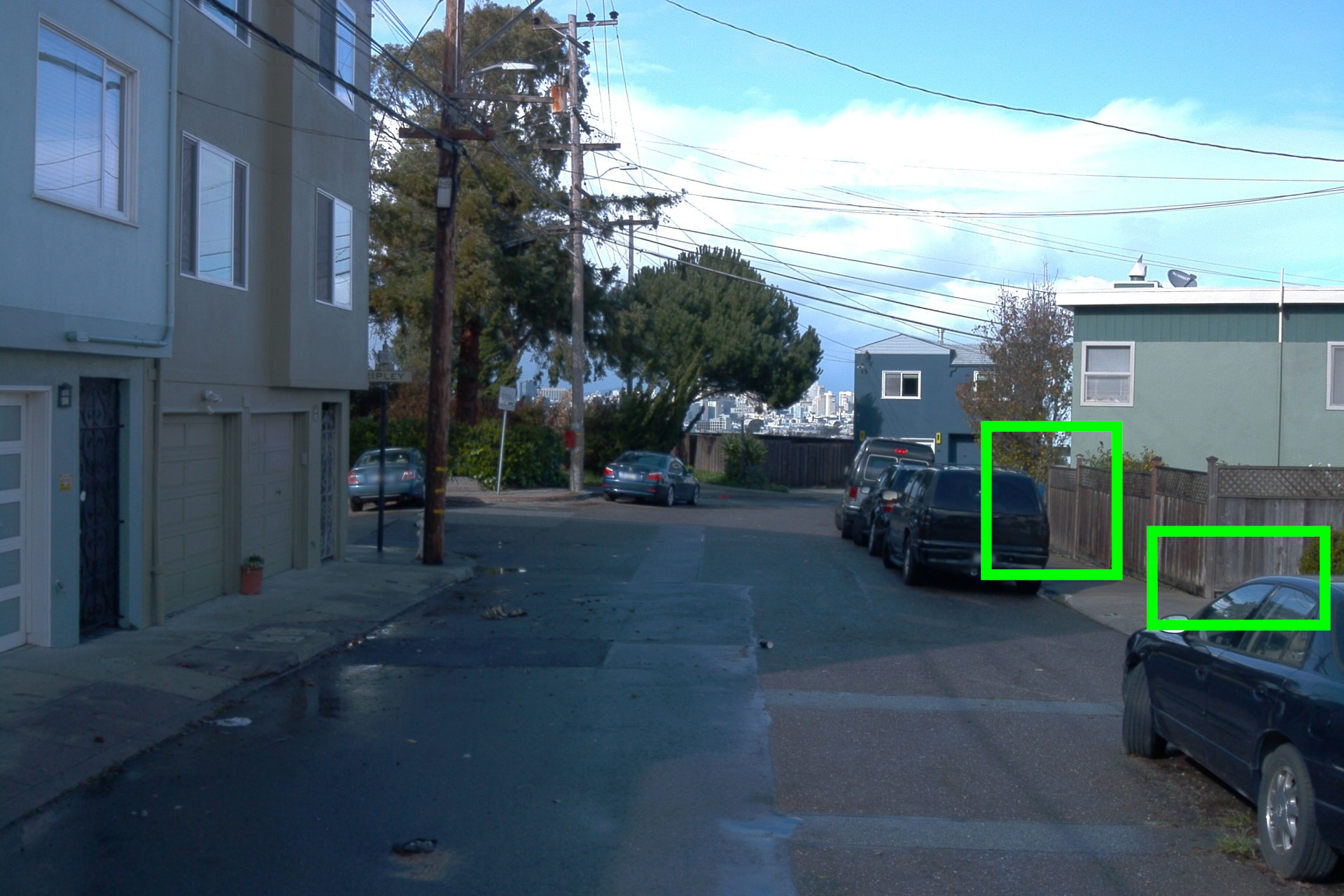}%
        \includegraphics[width=0.33\linewidth]{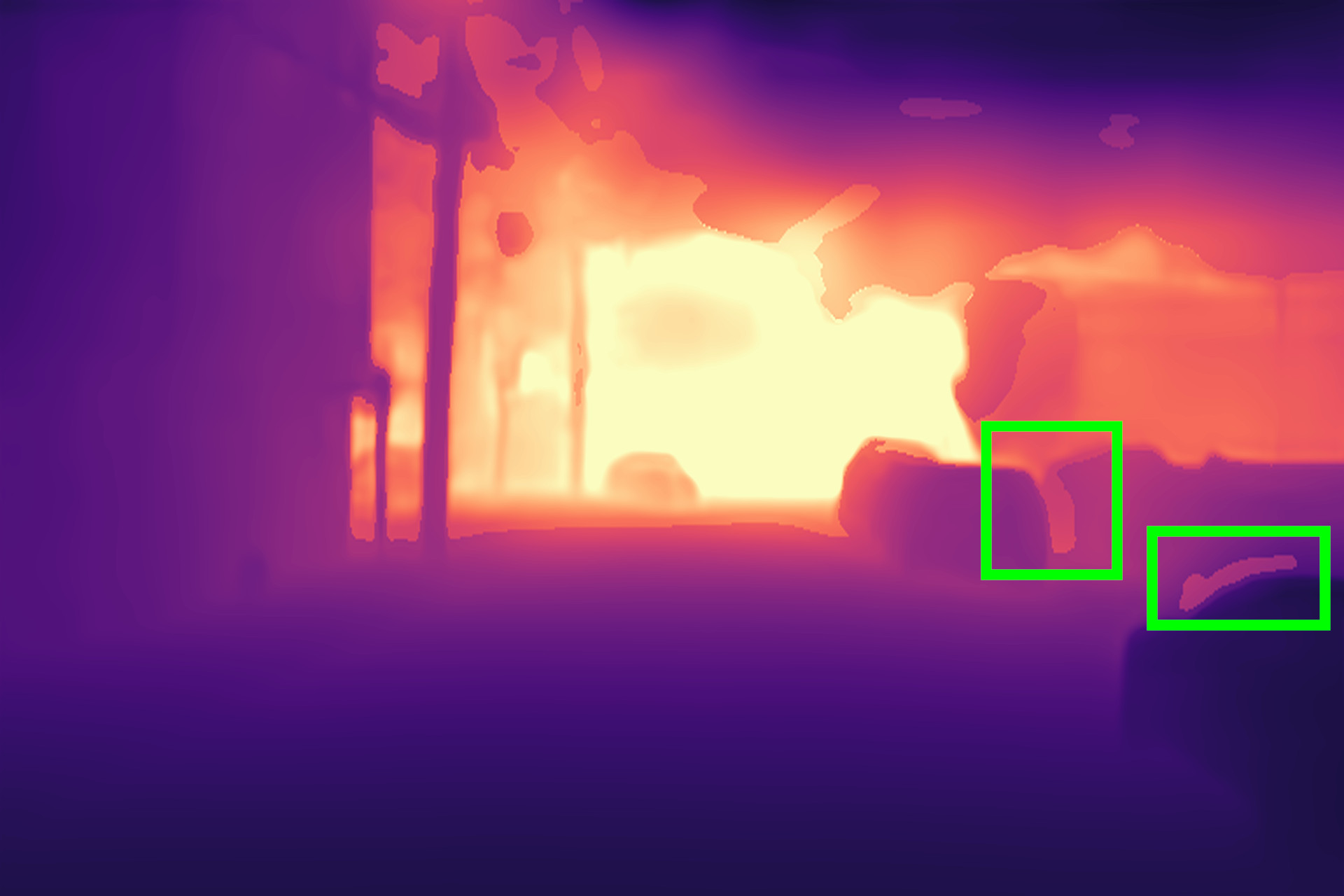}%
        \includegraphics[width=0.33\linewidth]{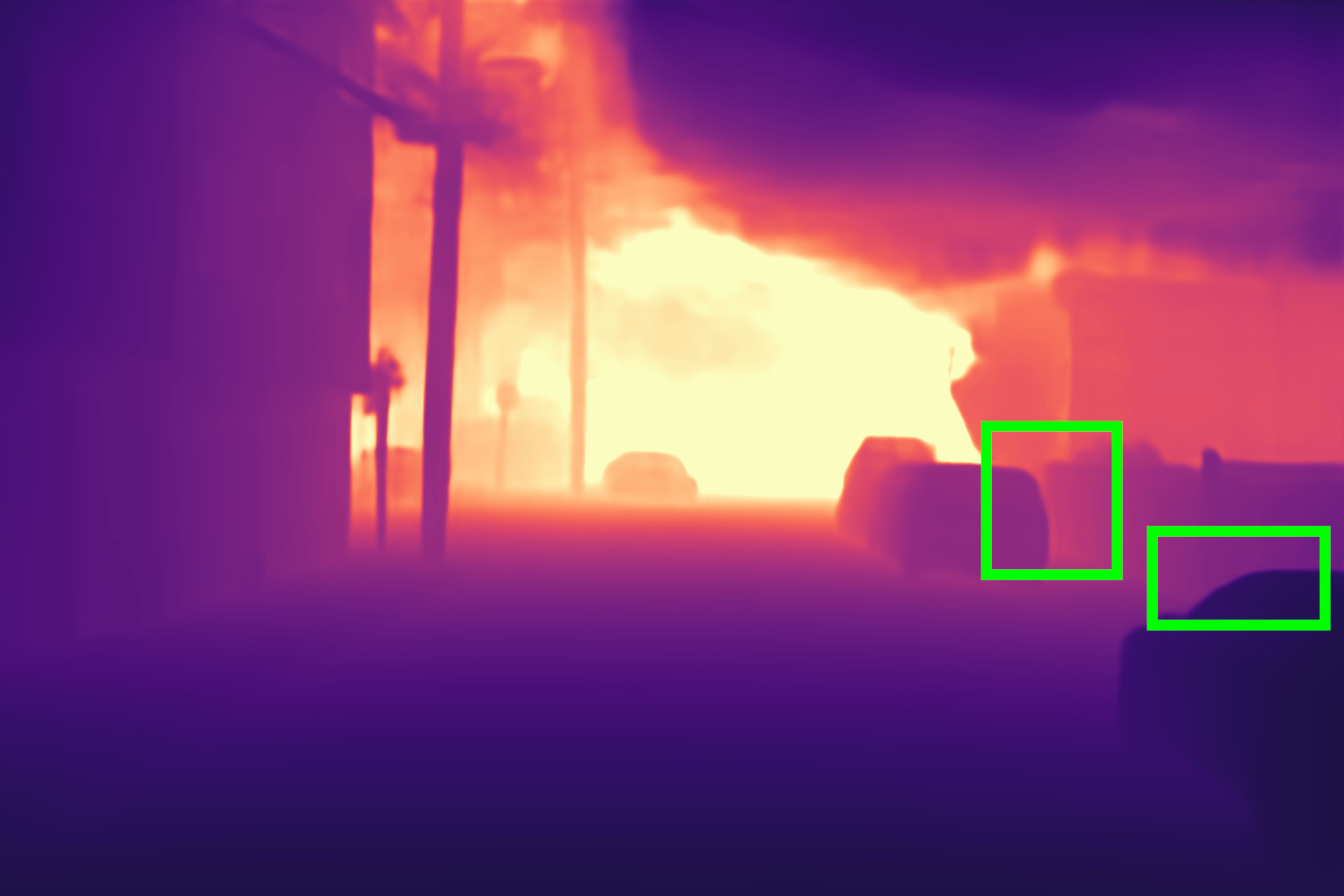}
    \end{subfigure}
    \begin{subfigure}[]{\textwidth}
        \centering
        \includegraphics[width=0.33\linewidth]{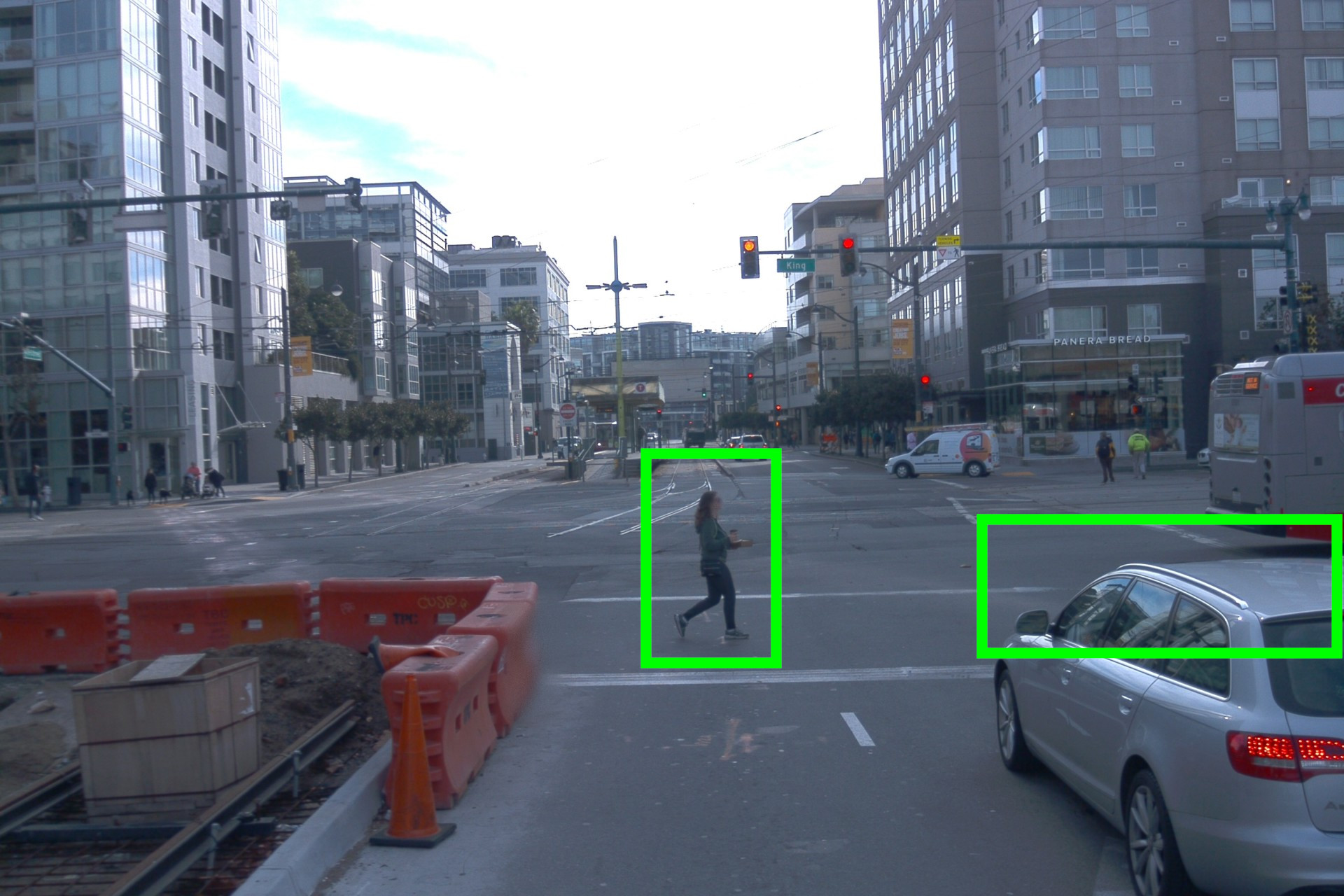}%
        \includegraphics[width=0.33\linewidth]{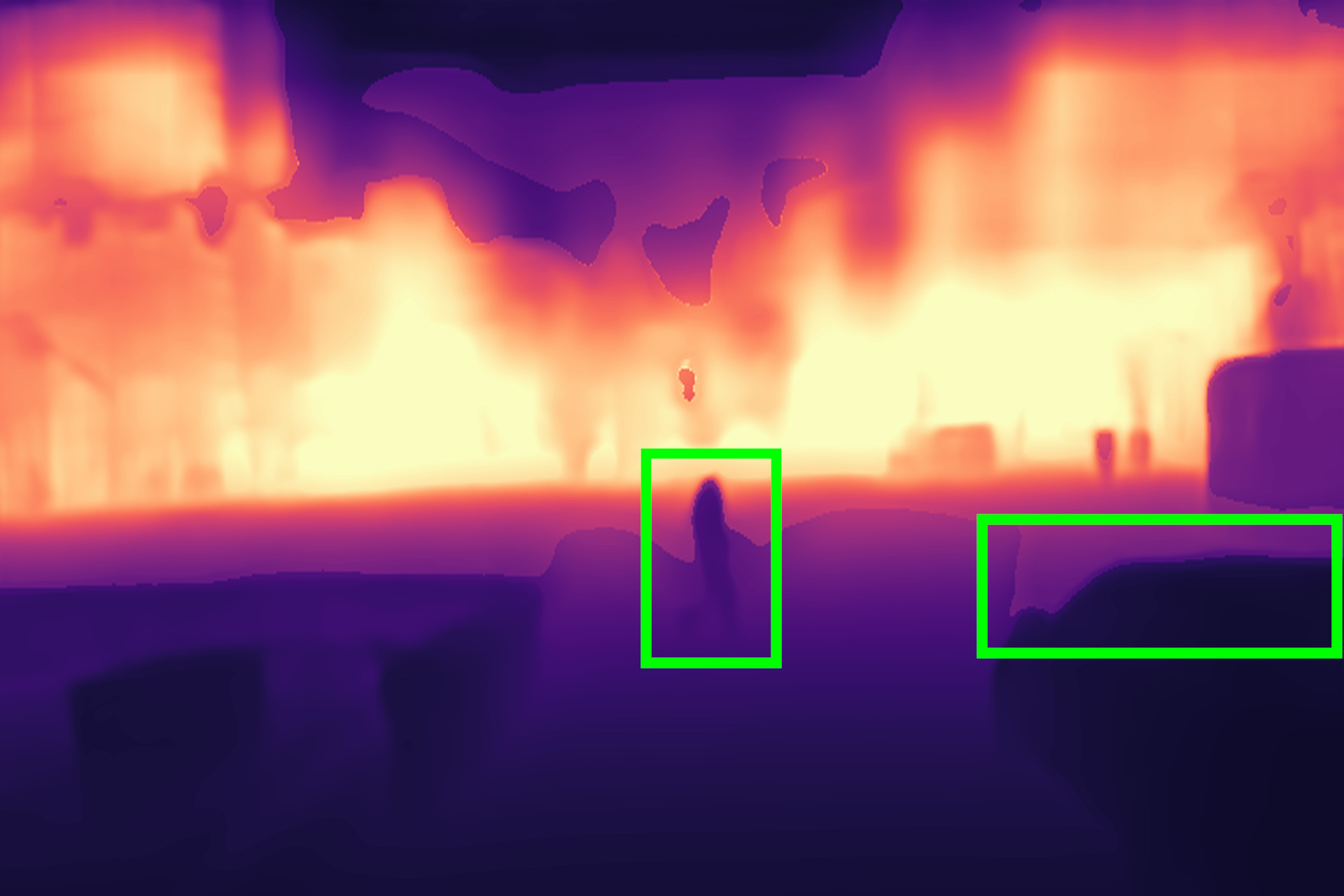}%
        \includegraphics[width=0.33\linewidth]{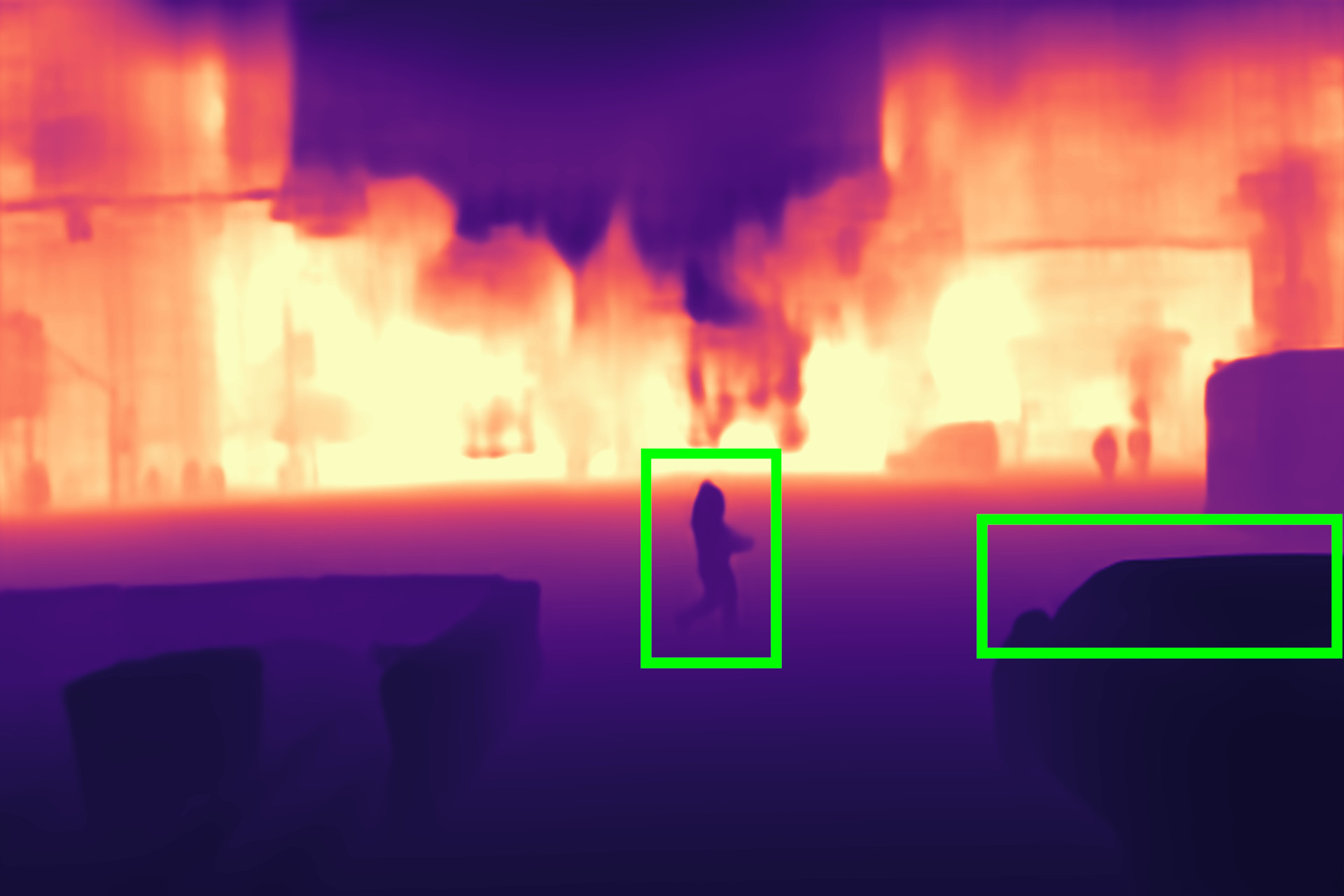}
    \end{subfigure}
    \begin{subfigure}[]{\textwidth}
        \centering
        \includegraphics[width=0.33\linewidth]{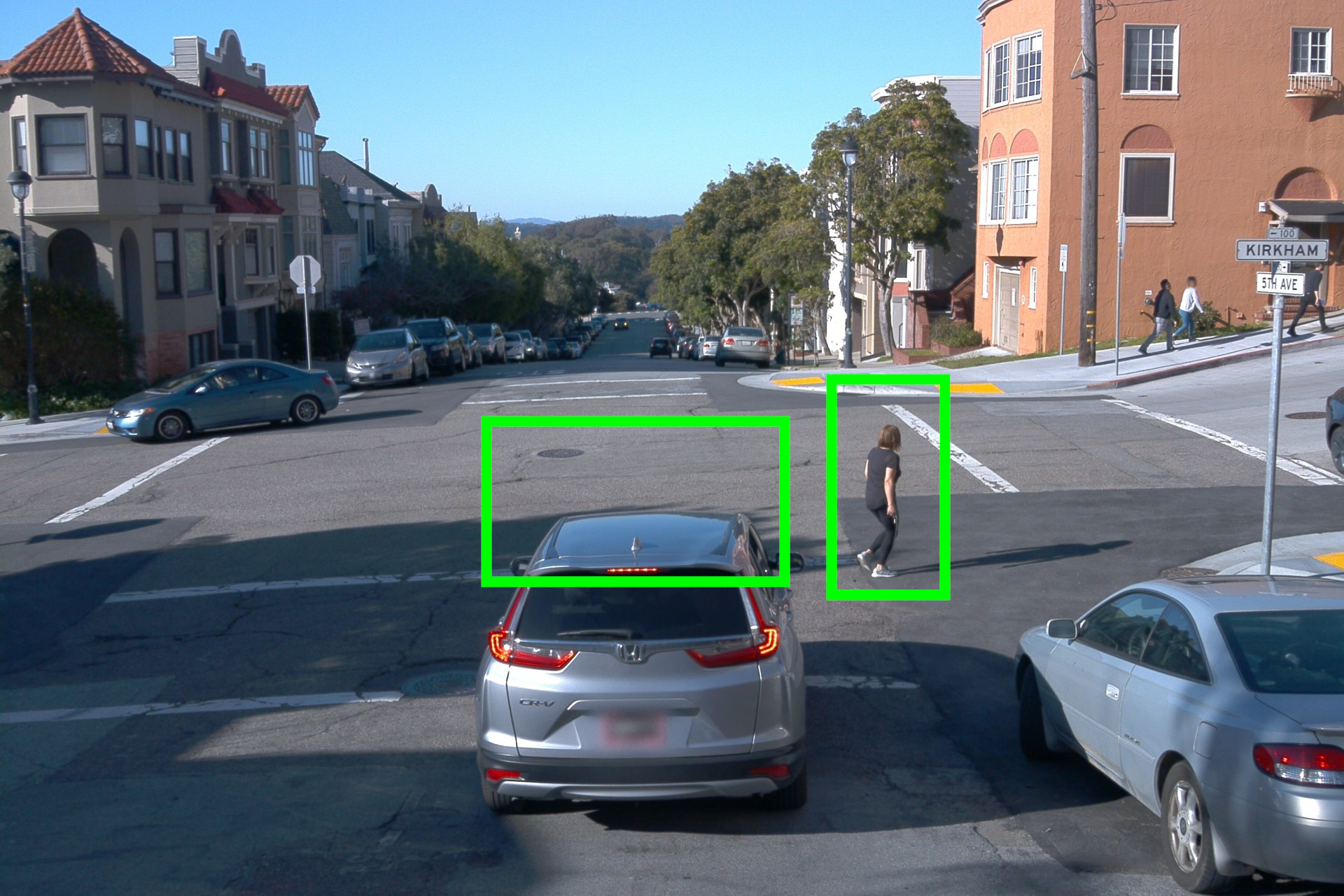}%
        \includegraphics[width=0.33\linewidth]{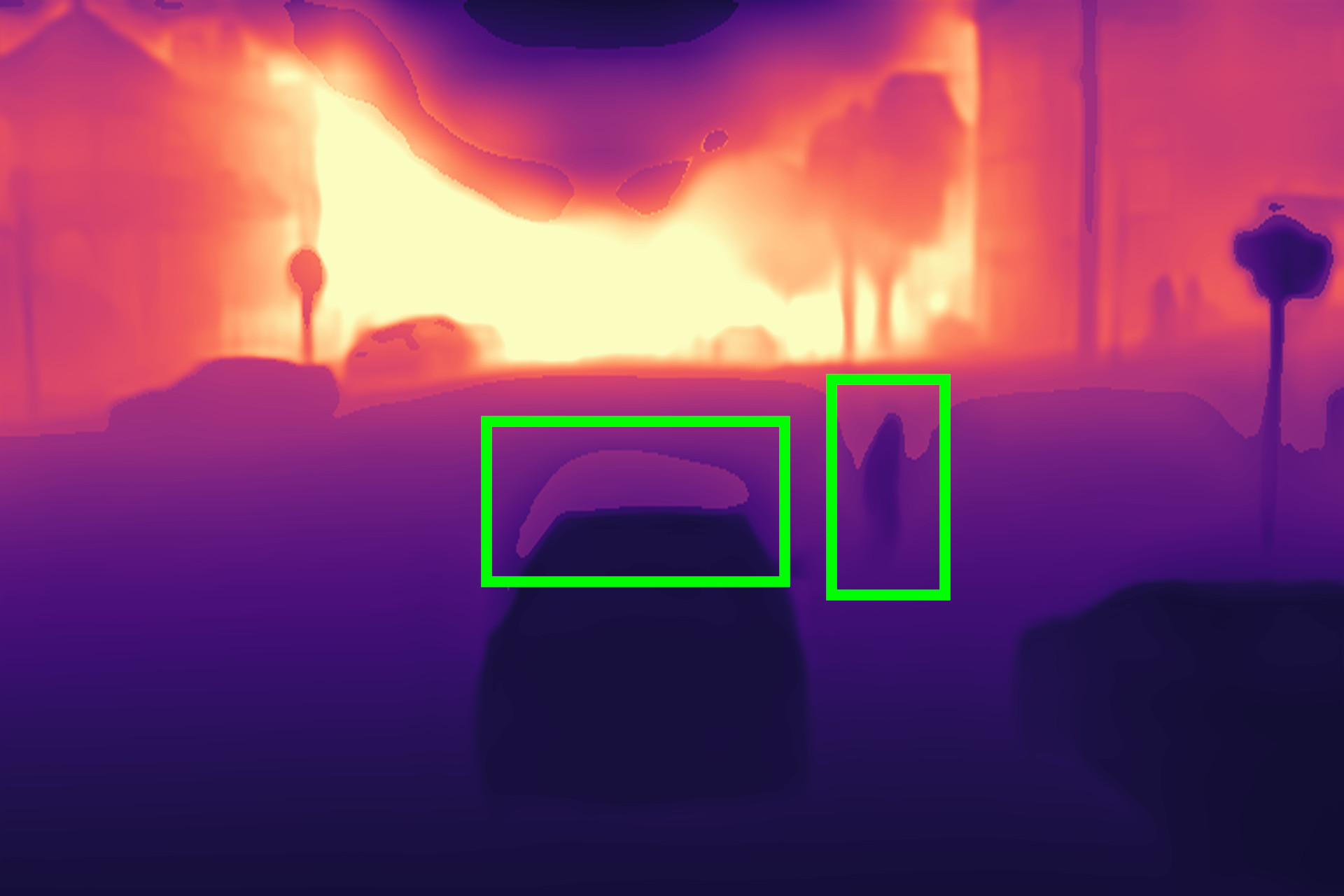}%
        \includegraphics[width=0.33\linewidth]{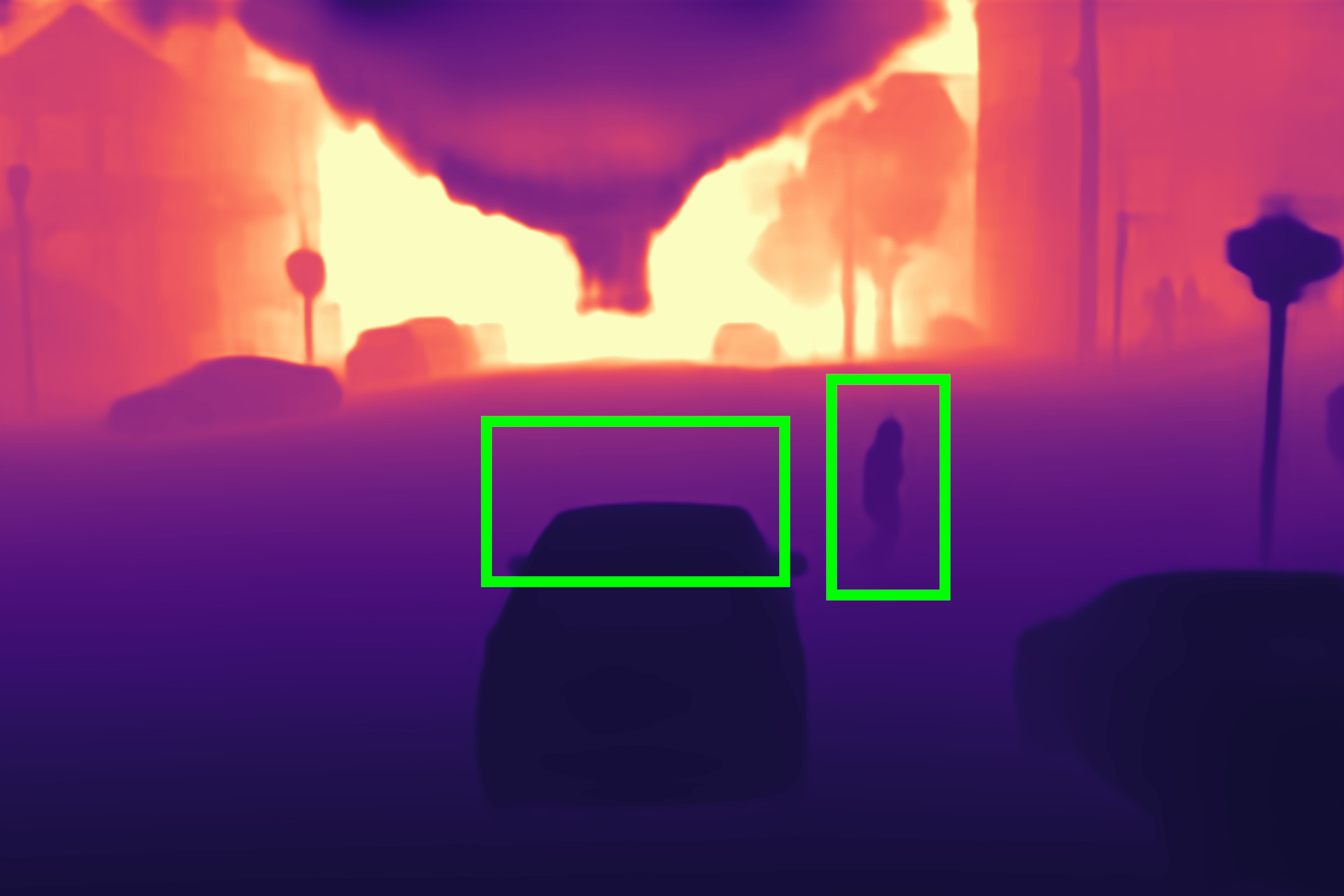}
    \end{subfigure}
    \begin{subfigure}[]{\textwidth}
        \centering
        \includegraphics[width=0.33\linewidth]{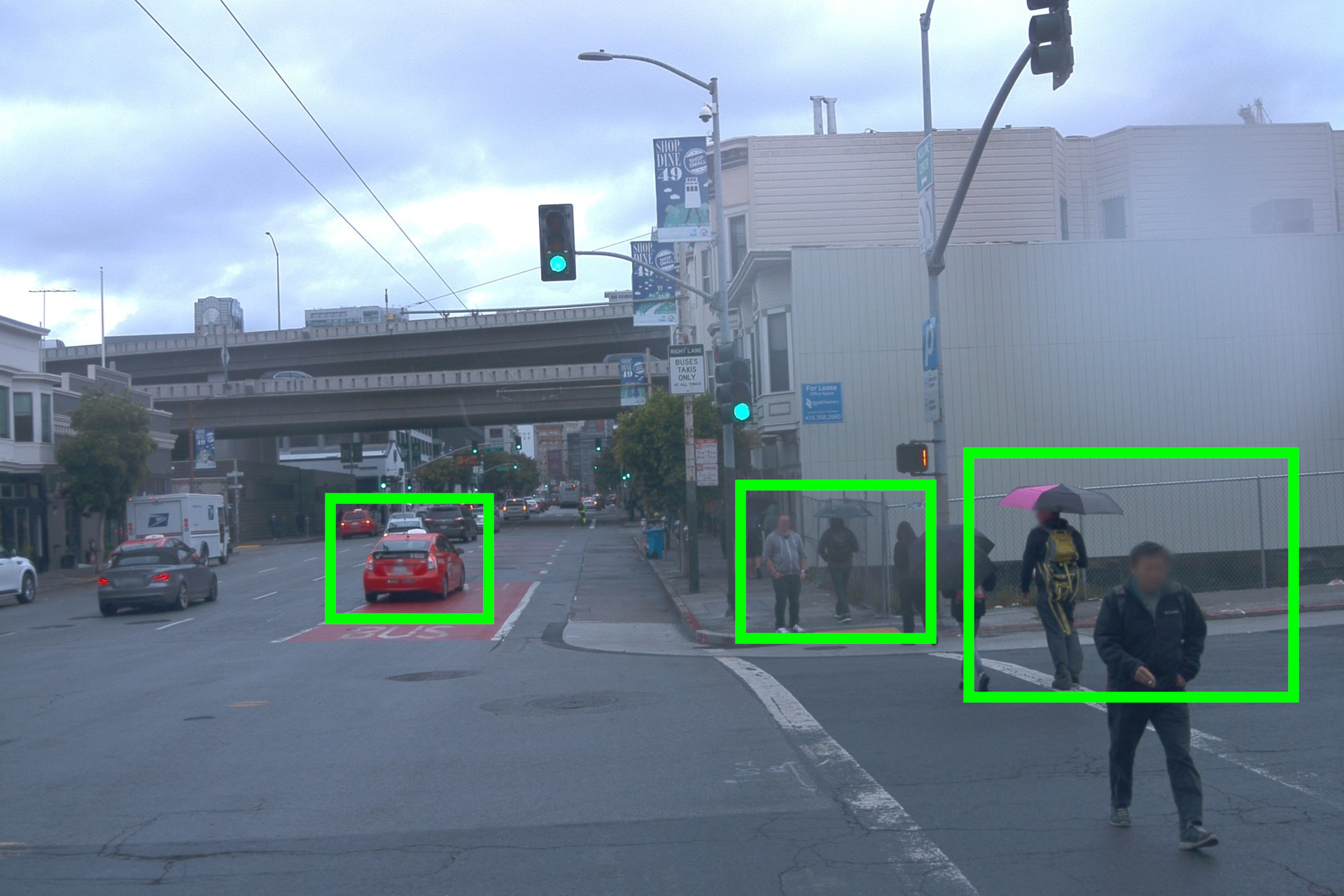}%
        \includegraphics[width=0.33\linewidth]{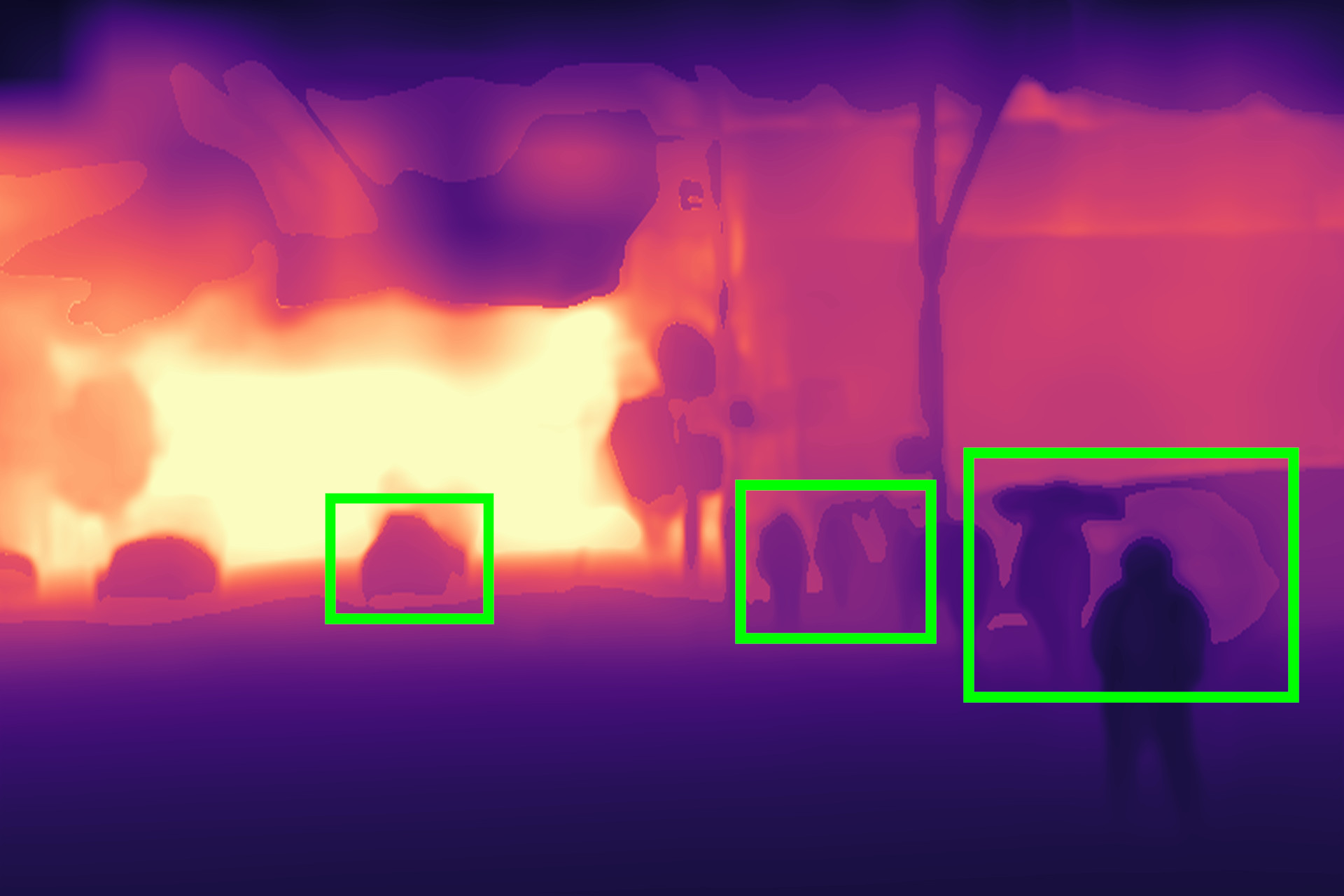}%
        \includegraphics[width=0.33\linewidth]{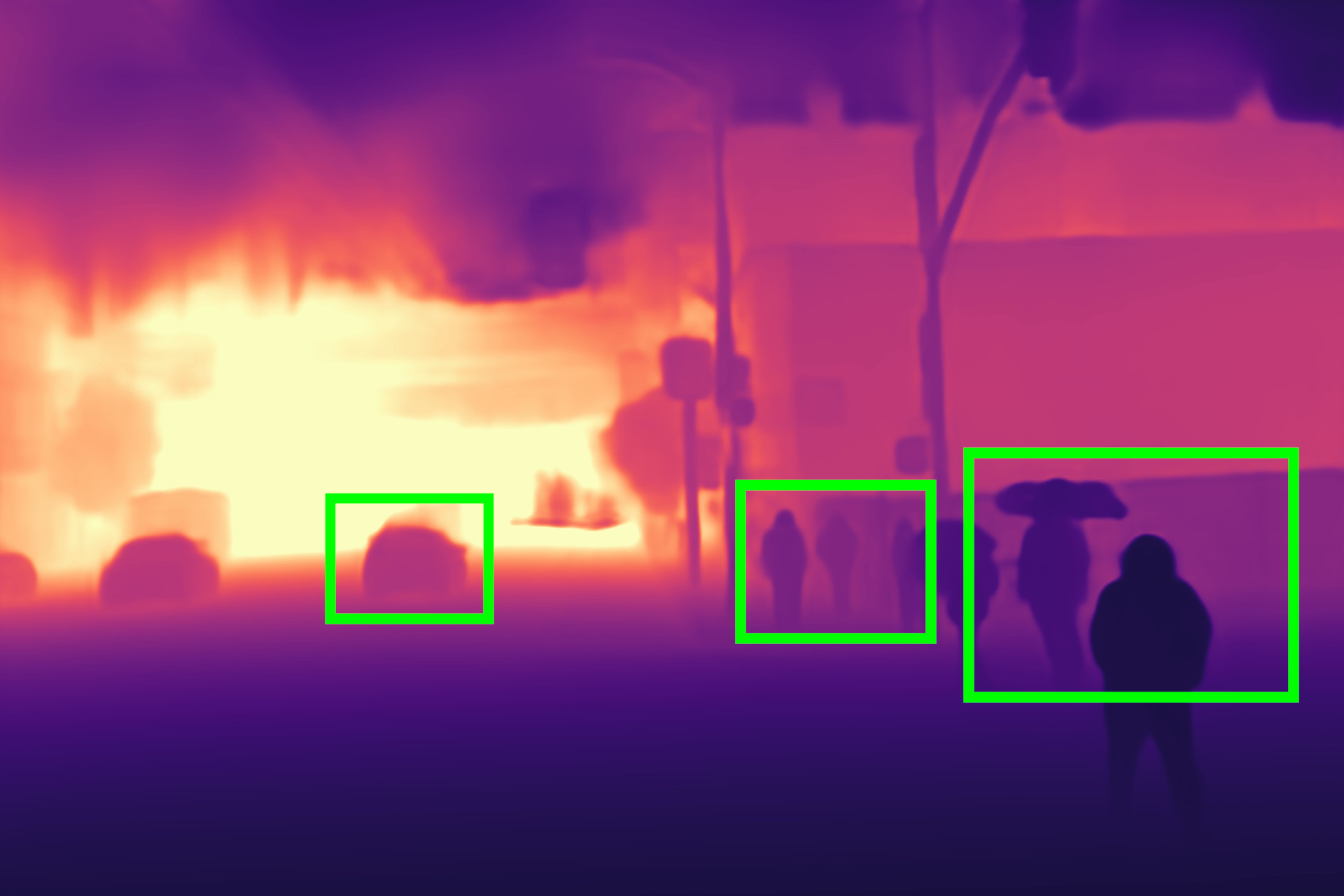}
    \end{subfigure}
    \caption{\textbf{\monoDE Visualization on \waymo.} [Key: 1\textsuperscript{st} column - RGB, 2\textsuperscript{nd} column - \zoeDepth model trained on raw \depthmaps, 3\textsuperscript{nd} column - \zoeDepth model trained on \methodName \depthmaps]}
    \label{fig:monoDE_waymo_vis}
\end{figure}

\subsection{\Depthmap Quality}
We provide short videos that show raw \depthmaps(top-left), clean \depthmaps(top-right), and binary masks (bottom-right) indicating the artifacts detected and removed by \methodName. We note from the videos that we do not relocate the incorrectly projected depth points but instead remove them from the \depthmaps. We also note that we only remove the artifacts while keeping the non-artifact points. The videos are named corresponding to the datasets on which the visualization was generated.

\subsection{\monoDE}
We provide additional visualizations of the predicted \depthmaps from \zoeDepth trained on raw and our \depthmaps as qualitative results supporting the benefit of \methodName. \cref{fig:monoDE_kitti_vis,fig:monoDE_waymo_vis} show visualizations from \kitti and \waymo datasets respectively.
We observe although there is a clear improvement in both \cref{fig:monoDE_kitti_vis,fig:monoDE_waymo_vis} in regions closer to the foreground/background boundary, there is a subtle difference in their respective improvements.
The majority of improvement in the predicted \depthmaps of \kitti dataset is in the region closer to the boundary and in the foreground which is in agreement with the results shown \cref{fig:monoDE_fore_analysis}.
However, the majority of improvement in the predicted \depthmaps of \waymo dataset is in the region closer to the boundary but in the background.

We provide results on \idisc\cite{piccinelli2023idisc} in \cref{tab:kitti_iDisc_stereo_eval} similar to the results in \cref{tab:kitti_zoe_stereo_eval}. We note that the improvement in the foreground is greater than that of the background.

\begin{figure}[!t]
    \centering
    \begin{subfigure}[]{0.32\textwidth}
        \includegraphics[width=\linewidth]{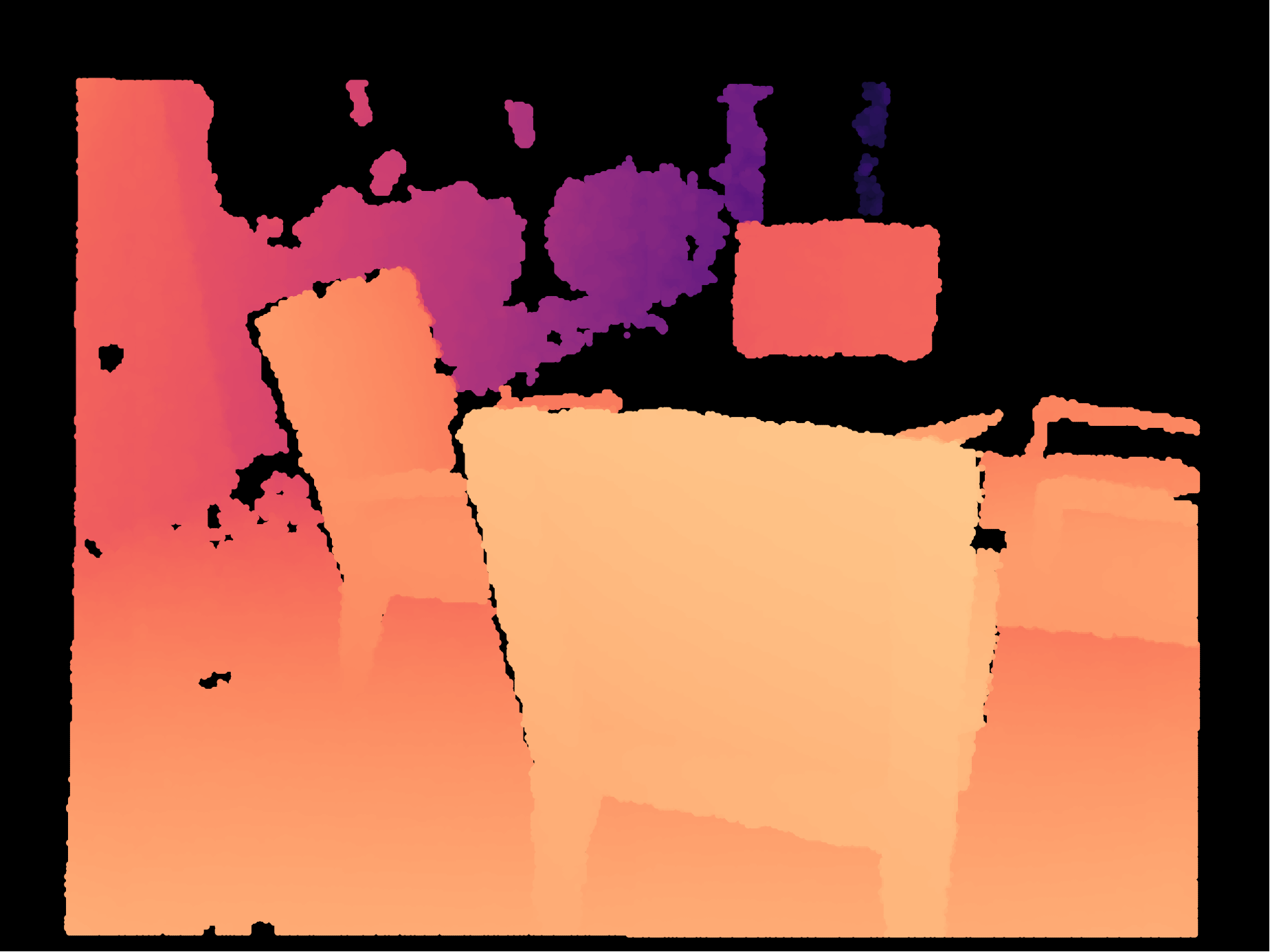}
    \end{subfigure}
    \begin{subfigure}[]{0.32\textwidth}
        \includegraphics[width=\linewidth]{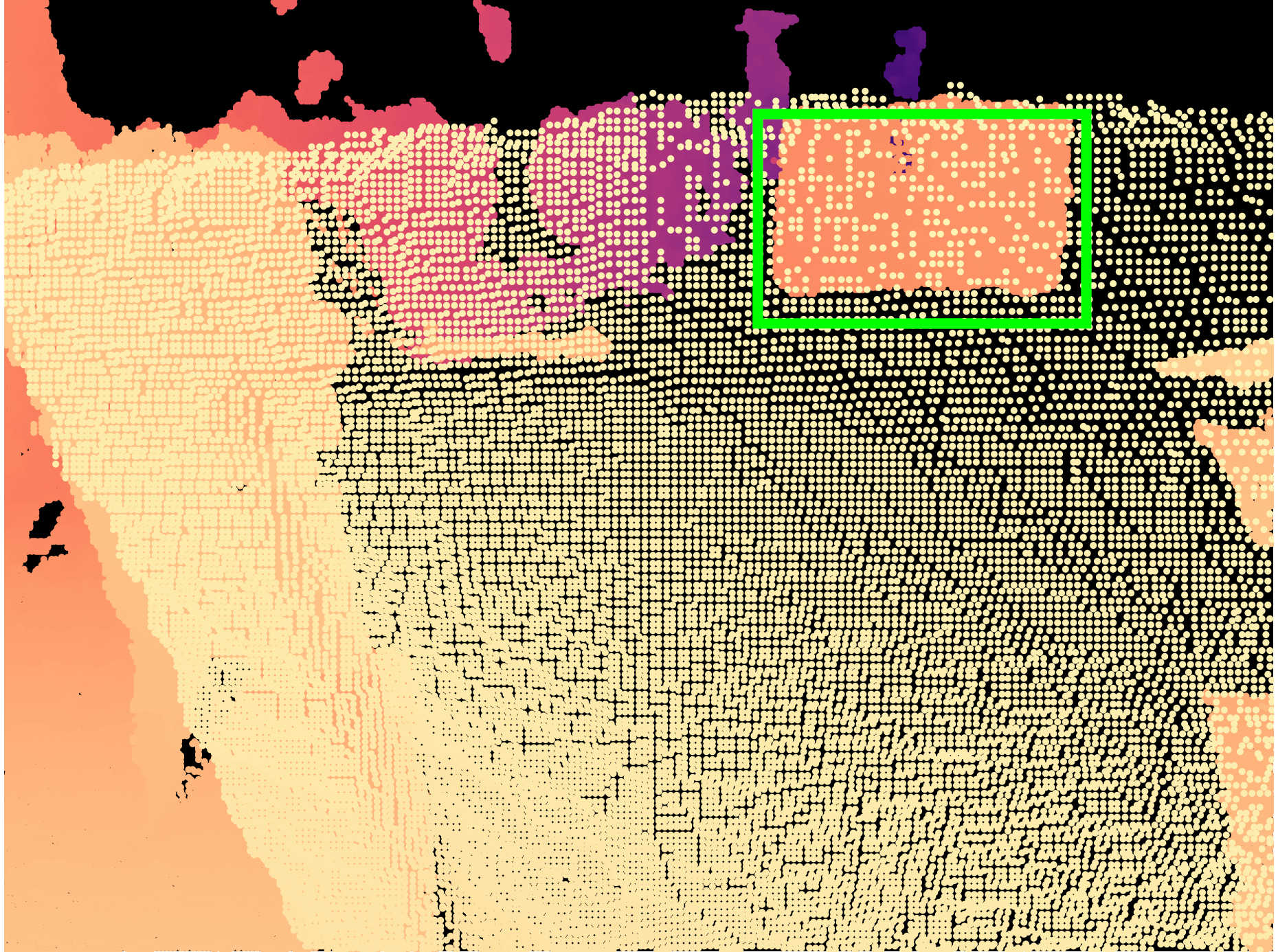}
    \end{subfigure}
    \begin{subfigure}[]{0.32\textwidth}
        \includegraphics[width=\linewidth]{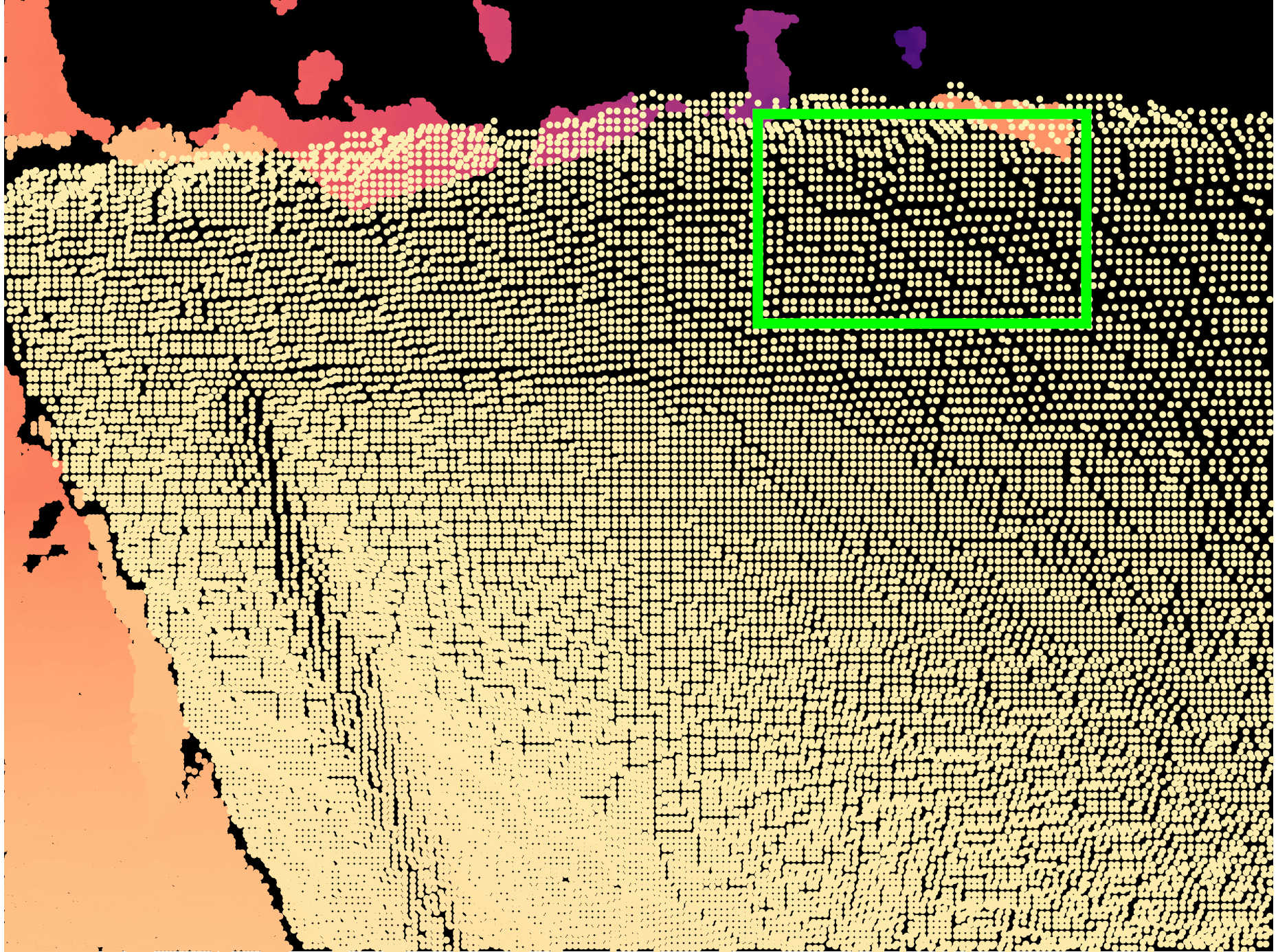}
    \end{subfigure}
    \begin{subfigure}[]{0.32\textwidth}
        \includegraphics[width=\linewidth]{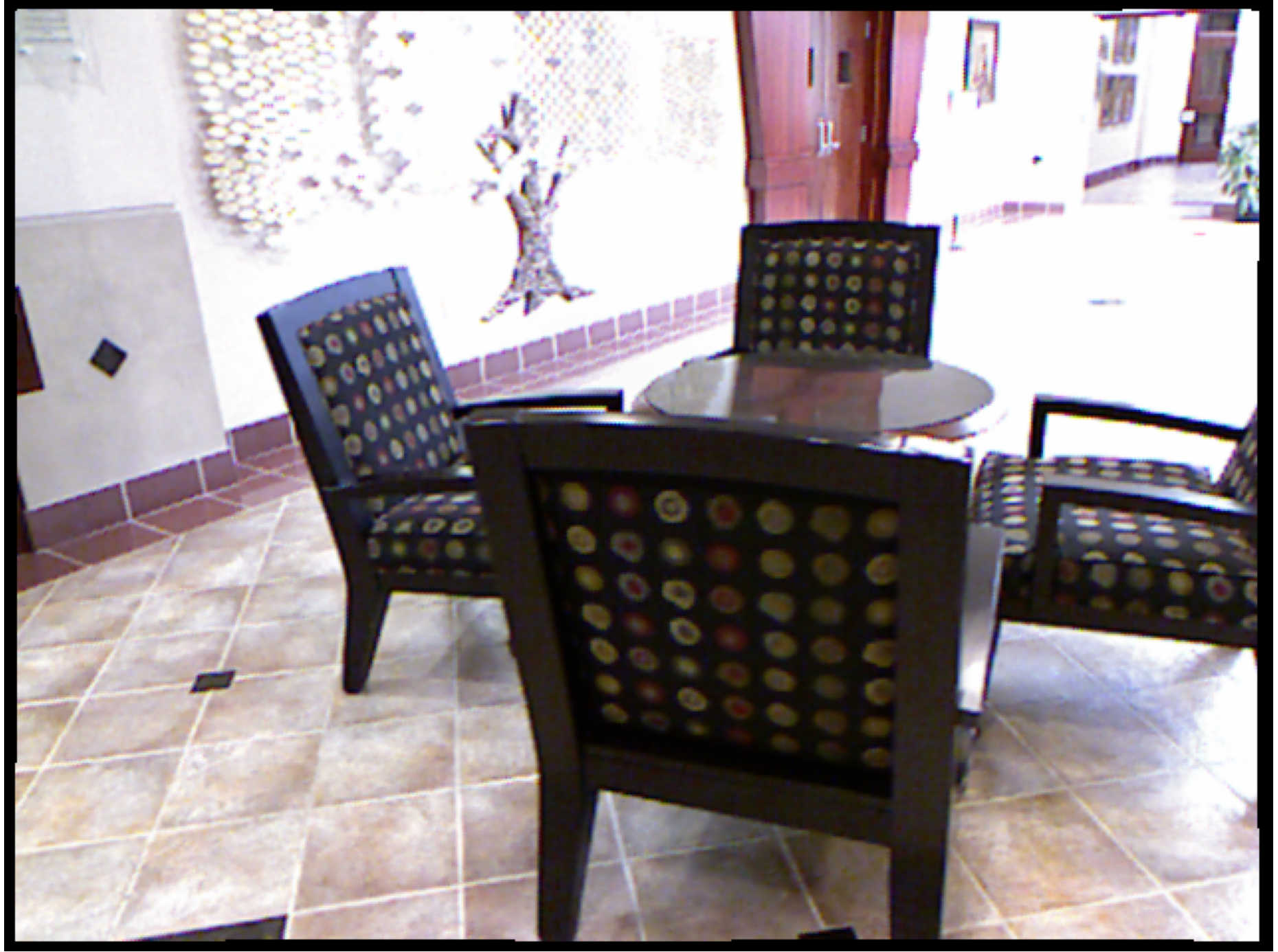}
        \caption{}
    \end{subfigure}
    \begin{subfigure}[]{0.32\textwidth}
        \includegraphics[width=\linewidth]{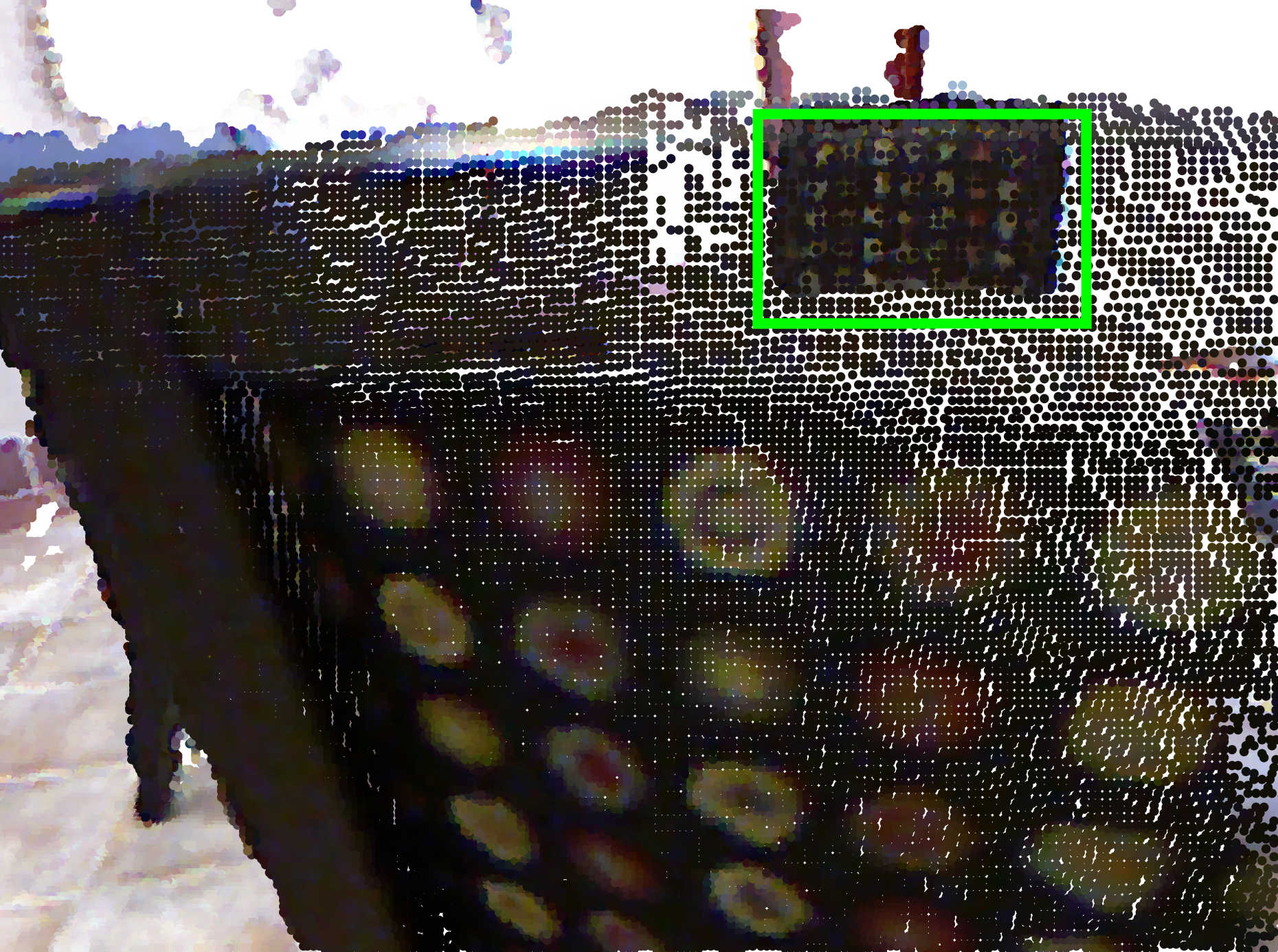}
        \caption{}
    \end{subfigure}
    \begin{subfigure}[]{0.32\textwidth}
        \includegraphics[width=\linewidth]{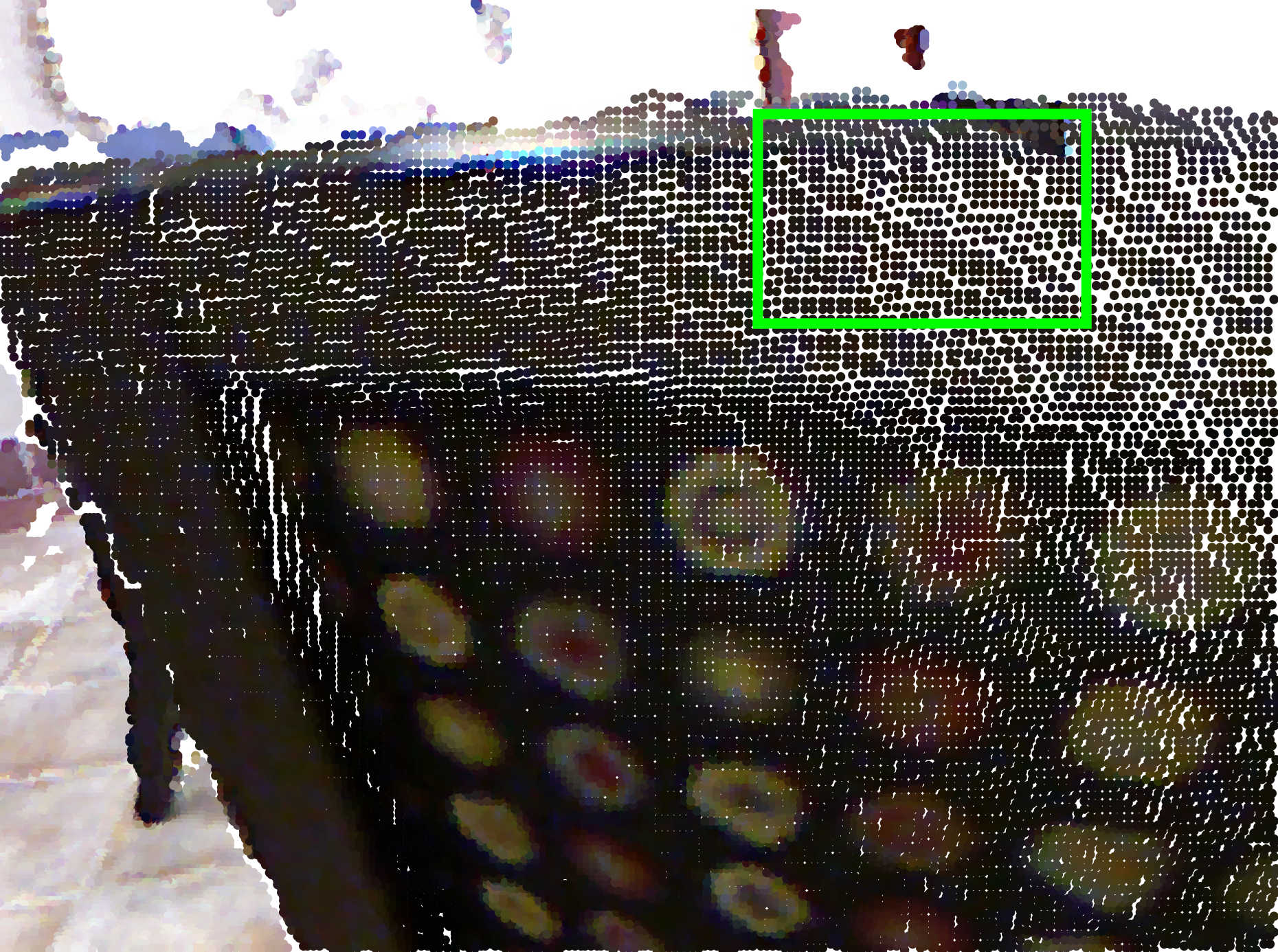}
        \caption{}
    \end{subfigure}
    \caption{\textbf{Indoor \Depthmap Visualization:} We show (a) GT RGB and \depthmap provided in the NYUv2 dataset and the corresponding novel view RGB and \depthmap renderings (b) before and (c) after using \methodName to remove the artifacts. }
    \label{fig:ar_vr_vis}
\end{figure}

\subsection{AR/VR Application}

As mentioned in \cref{sec:intro}, AR/VR applications also use \lidar to supplement the RGB camera with depth information. We use the NYUv2~\cite{silberman2012indoor} dataset to simulate \lidar-RGB pair and provide visualizations of artifacts and their removal by using \methodName.

The NYUv2~\cite{silberman2012indoor} dataset contains the GT RGB images and \depthmaps($\mathbf{D}_{GT}$) captured from the RGB-D sensor. 
We back-project the pixels into \threeD using provided camera parameters to simulate \lidar data. 
We then render RGB image and \depthmap($\mathbf{D}_{N}$) at a novel view by projecting the estimated \threeD points onto the novel view using given intrinsic parameters and arbitrary extrinsic parameters.
Thus, the GT \depthmap $\mathbf{D}_{GT}$ and novel-view \depthmap $\mathbf{D}_{N}$ constitute a virtual \depthmap $\mathbf{D}_{r}$ and RGB \depthmap $\mathbf{D}_{l}$ pair.
We apply \methodName on the pair of ($\mathbf{D}_{GT}$, $\mathbf{D}_{N}$) \depthmaps to remove artifacts in the novel view.
While performing rasterization for rendering, if multiple \threeD points are mapped to the same pixel, we only include RGB and depth values of the \threeD with the least depth.  
\cref{fig:ar_vr_vis} qualitatively shows that the RGB rendering after removing artifacts provides a better RGB image than with artifacts.

\section{Acknowledgements}
    This research was partially sponsored by the Army Research Office (ARO) grant W911NF-18-1-0330. 
    This document’s views and conclusions are those of the authors and do not represent the official policies, either expressed or implied, of the ARO or the U.S. government.
    We finally thank anonymous CVPR and ECCV reviewers for their exceptional feedback and constructive criticism that shaped this final manuscript.






\end{document}